\documentclass{article}
\pdfoutput=1
\usepackage[preprint]{data_paper}

\usepackage[utf8]{inputenc} 
\usepackage[T1]{fontenc}    
\usepackage{hyperref}       
\usepackage{url}            
\usepackage{booktabs}       
\usepackage{amsfonts}       
\usepackage{nicefrac}       
\usepackage{microtype}      
\usepackage{xcolor}         
\usepackage{graphicx}       
\usepackage{float}          
\usepackage{booktabs}
\usepackage{multirow}
\usepackage{bbding}
\usepackage{colortbl}
\usepackage{wrapfig}
\usepackage{threeparttable}
\usepackage{footmisc}
\usepackage{times}
\usepackage{latexsym}
\usepackage[T1]{fontenc}
\usepackage[utf8]{inputenc}
\usepackage{microtype}
\usepackage{inconsolata}
\usepackage{tikz} 
\usepackage{threeparttable}
\usepackage{array}
\usepackage{multirow}
\usepackage{makecell}
\usepackage{amssymb}
\usepackage{changes}
\usepackage{amsmath}
\usepackage{enumitem}
\usepackage{listings}
\usepackage{upquote} 
\usepackage{enumitem}
\usepackage{tabularx}
\usepackage{geometry}
\usepackage{longtable}
\usepackage{caption}
\usepackage{subfigure}
\usepackage{cleveref}
\usepackage{minitoc}

\crefname{figure}{Figure}{Figures}
\creflabelformat{figure}{#2#1#3}

\definecolor{bb}{rgb}{0.12,0.565,1}
\definecolor{gg}{rgb}{0.2,0.8,0.2}
\definecolor{yy}{rgb}{1,0.85,0.2}
\definecolor{rr}{rgb}{1,0.239,0.35}
\definecolor{lightgreen}{rgb}{0.68, 1.0, 0.68}

\newcommand{\ours}[0]{BioKGBench}
\newcommand{\ouragent}[0]{BKGAgent}
\newcommand{\ourAgent}[0]{BKGAgent}
\newcommand{\mypara}[1]{\vspace{1mm}\noindent\textbf{#1}~~}

\title{\ours{}: 
A Knowledge Graph Checking Benchmark of AI Agent for Biomedical Science
}

\author{%
  Xinna Lin$^{1,2}$\enskip\enskip
  Siqi Ma$^{1}$\enskip\enskip
  Junjie Shan$^{1}$\enskip\enskip
  Xiaojing Zhang$^{1}$\enskip\enskip
  Shell Xu Hu$^{3}$\enskip\enskip \\
  {\bfseries Tiannan Guo$^{1}$}\enskip\enskip 
  {\bfseries Stan Z. Li$^{1}$}\enskip\enskip
  {\bfseries Kaicheng Yu$^{1}$}\thanks{Corresponding Author}\\
  \textsuperscript{1}Westlake University \enskip
  \textsuperscript{2}Zhejiang University \enskip
  \textsuperscript{3}Samsung AI Center, Cambridge \enskip \\
  \texttt{\{linxinna, kyu\}@westlake.edu.cn}
}

\begin{document}

\doparttoc
\faketableofcontents

\maketitle

\begin{abstract}

Pursuing artificial intelligence for biomedical science, a.k.a. AI Scientist, draws increasing attention, where one common approach is to build a copilot agent driven by Large Language Models~(LLMs). 
However, to evaluate such systems, people either rely on direct Question-Answering~(QA) to the LLM itself, or in a biomedical experimental manner. How to precisely benchmark biomedical agents from an AI Scientist perspective remains largely unexplored. To this end, we draw inspiration from one most important abilities of scientists, understanding the literature, and introduce \ours{}. In contrast to traditional evaluation benchmark that only focuses on factual QA, where the LLMs are known to have hallucination issues, we first disentangle ``Understanding Literature'' into two atomic abilities, i) ``Understanding'' the unstructured text from research papers by performing scientific claim verification, and ii) Ability to interact with structured Knowledge-Graph Question-Answering~(KGQA) as a form of ``Literature'' grounding. We then formulate a novel agent task, dubbed KGCheck, using KGQA and domain-based Retrieval-Augmented Generation (RAG) to identify the factual errors of existing large-scale knowledge graph databases.   We collect over two thousand data for two atomic tasks and 225 high-quality annotated data for the agent task. Surprisingly, we discover that state-of-the-art agents, both daily scenarios and biomedical ones, have either failed or inferior performance on our benchmark. We then introduce a simple yet effective baseline, dubbed \ouragent{}. On the widely used popular knowledge graph, we discover over 90 factual errors which provide scenarios for agents to make discoveries and demonstrate the effectiveness of our approach. The code and data are available at \href{https://github.com/westlake-autolab/BioKGBench}{https://github.com/westlake-autolab/\ours{}}.

\end{abstract}

\section{Introduction}

Large Language Models (LLMs) are so powerful that they facilitate nearly every aspect of daily life and work right now, even research \cite{Zhao2023ASO,baek2024researchagent,He2023ASO,Zhou2023ASO}. Observing their marvelous successes in text generation \cite{yu2022survey,celikyilmaz2020evaluation}, text summarization \cite{el2021automatic,gambhir2017recent}, and other tasks \cite{jin2024position,Tang2023GraphGPTGI}, along with their consistent failures such as hallucinations \cite{ji2023towards,yao2023llm}, one can conclude that LLMs are powerful in certain tasks involving large-scale unstructured data like daily text or images, but relatively powerless when dealing with data-hungry scenarios. As such, researchers then construct AI agents \cite{wu2023brief,tian2023evil} assisting LLMs with external tools to extend the capabilities of LLMs. These attempts are fruitful in many fields, including autonomous computer \cite{steiner2008autonomous}, shopping web-agent \cite{lee2004ijade}, code developing \cite{dalle2004simcode}, society simulation \cite{drogoul2018multi,lan2023llm}, etc. A natural subsequent attempt is to develop AI agents to simulate scientists, aiding or even taking over the process of scientific discovery~\cite{baek2024researchagent}.

\begin{figure}[t]
    \vspace{-0.7cm}
    \centering
    \includegraphics[width=\textwidth]{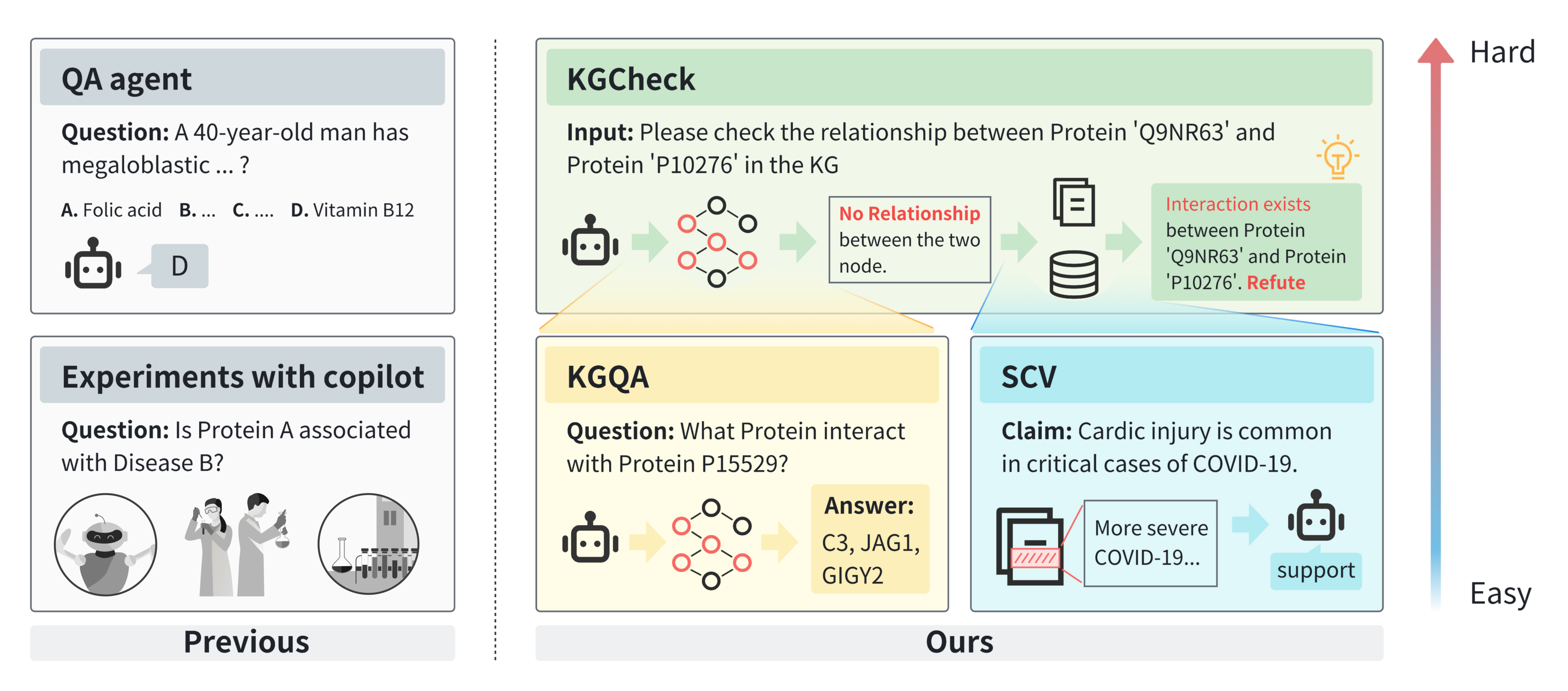}
    \vspace{-6mm}
    \caption{
    \textbf{(Left)} Previous benchmarks for domain-specific AI Agents either focus on the low-level tasks like question answering or are embedded in a complicated pipeline as a scientist copilot. \textbf{(Right)} We close the gap by constructing a knowledge graph checking task that consists of two atomic sub-tasks: Knowledge Graph Question Answering~(KGQA) and Scientific Claim Verification~(SCV), to provide a better evaluation of AI Agents in biomedical science domain.
   }
    \label{fig:teaser}
    \vspace{-1cm}
\end{figure}

As in Figure \ref{fig:teaser}, existing attempts can be grouped into two categories: i) to build an AI agent for a specific task, such as Question Answering (QA) in a specific domain~\cite{zhang2018medical}; ii) to encompass multiple AI agents to formulate a multi-agent system as the copilot of scientists, automating certain scientific activities, such as experiment result analysis~\cite{bi2023oceangpt,wang2023scientific}.

Literature review is the most critical ability that a scientist should possess \cite{snyder2019literature,thomas2020review}. It does not only involve reading and memorizing, but also requires scientists to understand and critically analyze. Researchers and scientists widely spend a significant amount of time in reading recent works. To save human efforts in scientific discovery, it is necessary for AI scientists to be able to accurately understand and analyze the existing research. Many researchers have dedicated to literature understanding in AI agents \cite{cai2024sciassess,li2024chatcite}, while a systematic evaluation system is missing and even underexplored. The current finest evaluation system \cite{cai2024sciassess} consists of multiple-choice questions extracted from literature, which cannot fully reveal the underlying reasoning regime of an agent's success or failure, leaving no clue for future advancement nor indicating whether the agent understands the reasoning rationale or merely memorizes data patterns.

On the other hand, another crucial research direction is to help AI agents capture the underlying logic of literature through domain-specific Knowledge Graphs~(KGs) \cite{abu2021domain,kejriwal2019domain}. 
KGs store massive knowledge triples in a graph-structured format \cite{hogan2021knowledge, alqaaidi2024knowledge}, complementing LLMs with external knowledge while providing frameworks for interpretation and reasoning \cite{meyer2023llm}. However, manually constructing such KGs is both intellectually and physically intensive. These domain-specific KGs require annotators with profound domain-specific knowledge, leading to high costs to create or maintain the knowledge graphs. As such, we observe that the existing and well-known biomedical KGs \cite{santos2022knowledge,chandak2023building} are not fully reliable due to outdated information. We attribute such discrepancy to the static nature of KGs, which lack mechanisms for dynamic updates to align with the evolution of external knowledge sources. 

In this paper, we propose a novel agent evaluation benchmark \ours{} to address both challenges simultaneously. As in Figure~\ref{fig:teaser}~(right), the ultimate goal of our benchmark is to verify the correctness of nodes and triples in the knowledge graph based on various information, including papers and well-maintained databases. We dub this task Knowledge Graph Checking~(KGCheck). Agents need to first query the information recorded on the KGs as directed, then cross-reference this information with external literature or databases to combat hallucinations. This task evaluates the agents’ capacities to both process and understand structured data (like KGs) and unstructured data (like literature). It is worth mentioning that the process of verifying knowledge within KGs closely mirrors the methodology of human scientific research, including database queries and extensive literature reviews. This similarity not only underscores the task's relevance to real-world scientific inquiry but also provides intriguing insights. Furthermore, we decompose this task into two more atomic subtasks: Knowledge Graph Question Answering~(KGQA) and Scientific Claim Verification~(SCV), enabling a more detailed evaluation of the agents' capabilities in processing and understanding of structured and unstructured data, respectively.

We extensively analyze existing AI agents on our benchmark and find that none of the existing agents can accomplish our tasks without moderate adaptation. Therefore, we introduce our agent \ourAgent{}, the first agent framework to interact with external knowledge graphs as well as research papers. Experiments demonstrate fascinating results that our agent is capable of discovering real conflicts in the existing large-scale datasets. Within 225 professional-annotated data in Clinical Knowledge Graph (CKG) \cite{chandak2023building}, our agent \ourAgent{} successfully identified some conflicting or missing pairs. This evidence further supports the academic value of our agent by providing researchers with a tool to update their own knowledge bases, offering substantial potential in both academic and commercial markets.
\section{Related Work}
\textbf{Science Agent.} The swift progression of large language models (LLMs) has catalyzed the widespread deployment of intelligent agents across diverse fields, notably within the science domain. Notable examples include ChemCrow~\cite{bran2023chemcrow} and Coscientist~\cite{boiko2023autonomous} in the field of chemistry, DoInstruct~\cite{bi2023oceangpt} in ocean science, and GeneGPT \cite{jin2024genegpt}, Almanac \cite{zakka2024almanac}, MedAgents~\cite{tang2023medagents} in biomedical domain, etc. Among them, biomedical agents, in particular, have garnered significant attention due to its critical importance. Biomedical agents~\cite{gao2024empowering} impact areas ranging from hybrid cell simulation~\cite{xiao2024cellagent}, the design of cellular circuits~\cite{chandrasekaran2024three} to the development of new therapies~\cite{zhenzhu2024gpt} and so on. We posit that biomedical agents will emerge as a focal point of research. However, the current benchmark in this field remains inadequate. For instance, MedAgents is evaluated in MedQA~\cite{zhang2018medical}, MedMCQA~\cite{pal2022medmcqa}, PubMedQA~\cite{jin2019pubmedqa}, relying heavily on inherent knowledge of LLMs, which leads to hallucinations easily. Our proposed \ours{} is a dynamic benchmark that evaluates the capabilities of agents in utilizing external tools and knowledge retrieval, thereby addressing this gap.  

\textbf{Agent Benchmark.} As agents are progressively applied across various domains, the urgency to construct corresponding benchmarks is escalating. Currently, the majority of benchmarks for evaluating agents adopt the approach of evaluating LLM-as-Agent~\cite{liu2023agentbench}, linking LLMs to external frameworks to assess their performance on specific tasks. For instance, AgentBench~\cite{liu2023agentbench} is a general benchmark for evaluating an agent's reasoning and decision-making capabilities, SWE-bench~\cite{jimenez2023swe} assesses an agent's proficiency in software engineering, and AgentClinic~\cite{schmidgall2024agentclinic} examines an agent's performance in a simulated clinical environment. However, a benchmark in AI Scientist perspective remains largely unexplored. Our benchmark originates from this perspective, taking the processing and understanding of large-scale data scenarios as the entry point, representing an initial attempt in this direction.

\textbf{Agent Integrating LLMs and KGs.} The collaborative use of LLM and KG has become one of the leading methodologies in contemporary agent design, aimed at alleviating uncertainties stemming from the intrinsic mechanisms of LLMs \cite{pan2024unifying, Chen2023DifferentiableNR, Yang2023GiveUT}. This paradigm not only capitalizes on the generalization ability of LLMs but also employs KGs as an external, trustworthy, and structured data source, thereby achieving reasoning proficiency that strikingly emulates human intellect\cite{pan2024unifying}. For instance, StructGPT \cite{jiang2023structgpt} boosts an LLM's performance on general questions by tapping into the information from a supplied KG. Similarly, KG-Agent \cite{jiang2024kg} leverages knowledge from KGs, synthesizing instruction data for fine-tuning an open-sourced LLM, thereby achieving competitive performance on general question-answering tasks. However, to our knowledge, while this paradigm has been widely applied to the general question-answering area, its potential remains untapped in the biomedical field. \ouragent{}, hence, is poised to fill this gap.
\section{\ours{}}
\label{method}
\setlength{\columnsep}{9pt}
\begin{wraptable}{r}{0.5\linewidth}
    \vspace{-7mm}
    \centering
    \caption{Statistics of our \ours{}. }
    \setlength{\tabcolsep}{2.6pt}
    \begin{tabular}{llllll}
        \Xhline{1pt}
        \multirow{3}*{Task}            & \multirow{3}*{Metric}    & \multirow{3}*{Scope}       & \multicolumn{3}{c}{Data}   \\ \cmidrule(r){4-6}
                        &           &                   & Dev   & Test  & All        \\ \midrule
        KGQA            & F1        & KG                & 60    & 638   & 698        \\
        SCV             & Acc.      & Text        & 120   & 1,265  & 1,385        \\
        KGCheck         & Acc.      & KG + T   & 20    & 205   & 225        \\ 
        \Xhline{1pt}
    \end{tabular}
    \label{tab:dataset_stats}
    \vspace{-3mm}
\end{wraptable}

Here, we present our benchmark in detail. As aforementioned, one key ability of ``AI Scientists'' is to understand domain knowledge. However, current LLM-driven agent systems inevitably suffer from hallucinations as a consequence of the statistical nature of LLMs along with the lack of scientific training data compared to data from daily scenarios. We notice that a recent trend in research is to use AI agents to leverage external tools to address these limitations~\cite{bran2023chemcrow,bi2023oceangpt}. Drawing inspiration from this, we design two atomic abilities to evaluate AI scientists, i) Knowledge Graph Question Answering~(KGQA) aiming to address the hallucination issue by grounding the knowledge with structured knowledge graphs; and ii) Scientific Claim Verification~(SCV) based on retrieved text from peer-reviewed research papers. In addition, we propose an encompassing task combining these two atomic abilities, to perform Knowledge Graph Checking~(KGCheck) as shown in Figure~\ref{fig:teaser}. The motivation behind this stems from our interviews with experts from biomedical domains. Their answers to the question ``What is the most expected AI agent you would like to use in your daily research?'' often included an AI agent that helps in extensive literature review and claim verification. We report the statistics over the scopes of knowledge search, including knowledge graphs and academic literature, in Table~\ref{tab:dataset_stats}.

\begin{wrapfigure}{r}{0.5\linewidth}
    \vspace{-7mm}
    \centering
    \includegraphics[width=\linewidth]{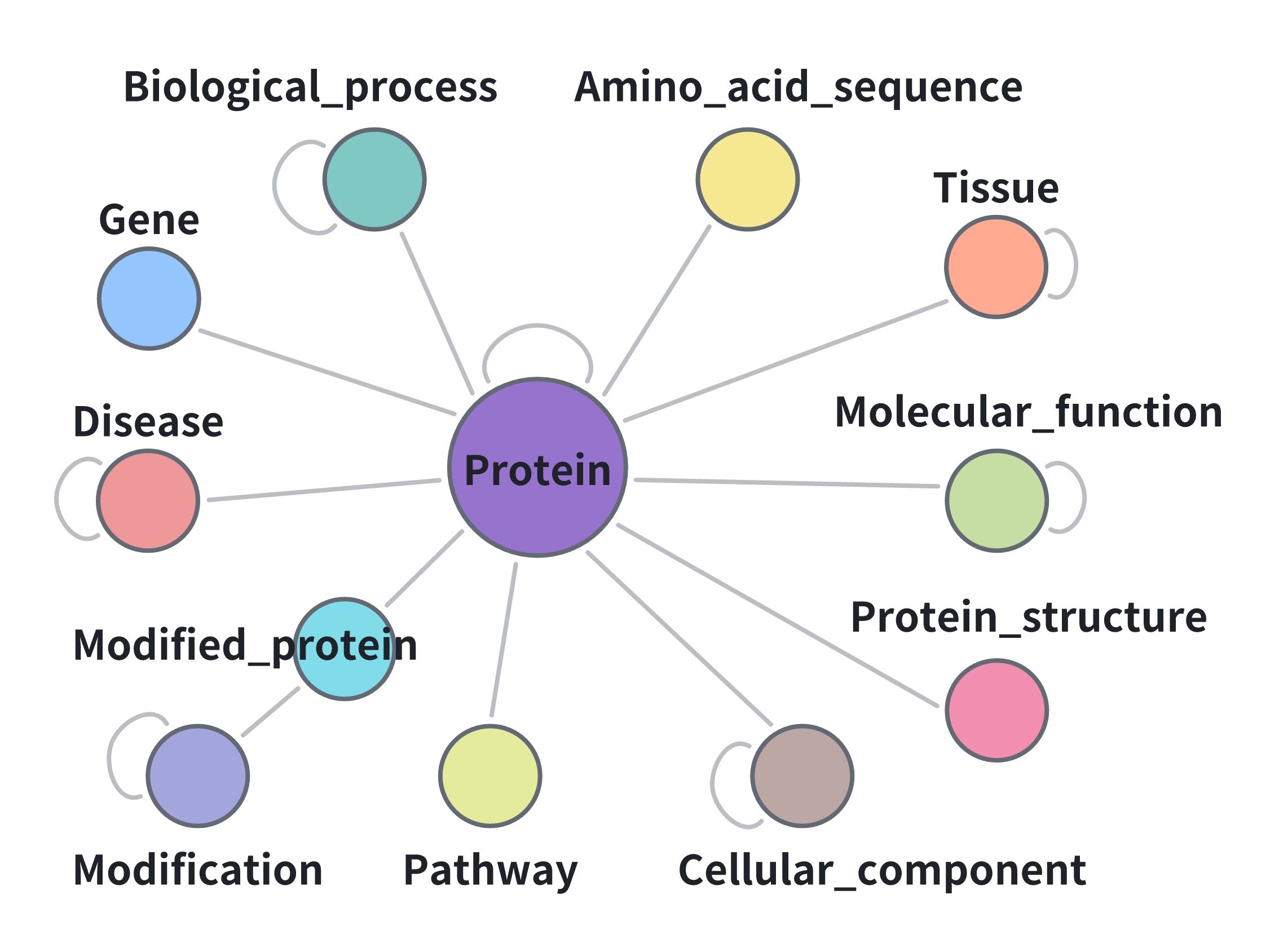}
    \vspace{-8mm}
    \caption{The sub-graph of the Clinical Knowledge Graph (CKG) retains 12 types of nodes and 18 kinds of relationships.}
    \label{fig:subkg}
    \vspace{-5mm}
\end{wrapfigure}

\subsection{Atomic Ability}
\subsubsection{Knowledge Graph Question Answering} 
This atomic task in the benchmark is to evaluate the agents’ ability to interact with structured Knowledge Graph Question Answering as a grounding of academic literature. Without loss of generality, we choose Clinical Knowledge Graph~(CKG)~\cite{li2020real} as the source of our data, which is one of the most popular large-scale knowledge graph databases in the biomedical domain. 
CKG is a knowledge graph database with data imported from diverse biomedical databases, aimed at streamlining automated knowledge discovery through the graph's extensive information. 

\begin{table}[b!]
\centering
\vspace{-5mm}
\caption{Statics of three different reasoning types of KGQA dataset.}
\begin{tabular}{m{2.5cm} m{2.5cm} m{3cm} m{2.5cm} c}
\Xhline{1pt}
 \textbf{Reasoning Type} & \centering \textbf{Graph} & \centering \textbf{Example Question} & \centering \textbf{Question Types} & \textbf{\%} \\

\hline
\textbf{One-hop}  &
\centering
\begin{tikzpicture}
    \draw (0,0) circle (4pt);
    \draw[->] (0.15, 0.08) -- (0.7, 0.4);
    \draw[blue, thick] (0.87, 0.5) circle (4pt);
    \draw[->] (0.15, -0.08) -- (0.7, -0.4);
    \draw[blue, thick] (0.87, -0.5) circle (4pt);
\end{tikzpicture} 
& What proteins does the protein O94842 act on? & \centering 8 & 56.0\\
\hline
\textbf{Multi-hop}  &
\centering
\begin{tikzpicture}
    \draw (0,0) circle (4pt);
    \draw[->] (0.15, 0) -- (0.8, 0);
    \draw (1, 0) circle (4pt);
    \draw[->] (1.15, 0.08) -- (1.7, 0.4);
    \draw[blue, thick] (1.87, 0.5) circle (4pt);
    \draw[->] (1.15, -0.08) -- (1.7, -0.4);
    \draw[blue, thick] (1.87, -0.5) circle (4pt);
\end{tikzpicture} 
& What diseases are associated with the protein encoded by the gene KCNS1? & \centering 4 & 28.7\\
\hline
\textbf{Conjunction}  &
\centering
\begin{tikzpicture}
    \draw (0,0) circle (4pt);
    \draw[->] (0.15, 0.08) -- (0.7, 0.4);
    \draw[blue, thick] (0.87, 0.5) circle (4pt);
    \draw[->] (0.15, 0) -- (0.8, 0);
    \draw (1, 0) circle (4pt);
    \draw[->] (0.15, -0.08) -- (0.7, -0.4);
    \draw[blue, thick] (0.87, -0.5) circle (4pt);
    \draw (1.7, 0) circle (4pt);
    \draw[->] (1.58, 0.08) -- (1.01, 0.41);
    \draw[->] (1.58, -0.08) -- (1.01, -0.41);
\end{tikzpicture} 
& Which pathway are the proteins P02778 and P25106 both annotated in? & \centering 4 & 15.3\\
\Xhline{1pt}
\end{tabular}
\vspace{-.2cm}
\label{tab:kgqa}
\end{table}
As the original database is unnecessarily large, we focus on a sub-graph to mitigate the challenge while preserving all relevant information. Starting from the origin of CKG---protein, we select the sub-graph to contain exactly 12 categories of biological entities, as indicated in Figure \ref{fig:subkg}. Thus, the sub-graph consists of 484,955 entities (nodes) across 12 categories (Biologically defined) and 18,959,943 relationships (edges) of 18 types, with each type consisting of relationships between a unique pair of entity categories.

After the sub-graph is ready, we construct the question set for the Question Answering (QA) database in two steps. We first handcraft question templates by selecting biomedical fields and pinpointing entities and relations in the CKG. Natural language questions were constructed in various formats, ensuring their accuracy through peer reviews and expert consultations. We then expand our dataset with autogenerated questions by matching CKG data to constructed QA templates, resulting in the generation of 698 questions across three reasoning types and 16 question categories (refer to Table~\ref{tab:kgqa}). 

In this task, we outfit LLMs with a set of atomic KG-querying tools and ask them to answer biomedical questions by querying the provided KG. The responses will be compared with the gold answers and evaluated using the F1 score, where the gold answer to the input question is typically characterized by a set of KG entities. It is noteworthy that our KGQA is built upon a biomedical KG rather than a common sense KG, the former being characterized by its highly dense information storage. This compactness greatly increases the complexity for agents to employ tools and perform reasoning on the structured graph to complete QA tasks. This task enables the development of assessing the robustness and tool learning ability of agents built upon various LLMs, and hopefully it would aid in guiding agents to leverage the extensive biomedical knowledge within the KG, thereby propelling scientific discovery.

\subsubsection{Scientific Claim Verification} 
\begin{table}[]
\centering
\caption{Examples of reconstructed dataset for SCV, where data from PubMedQA is converted from QA to declarative claims. "NEI" stands for "Not Enough Information".}
\resizebox{\textwidth}{!}{%
\begin{tabular}{lcc}
\Xhline{1pt}
\textbf{Example Claim} & \textbf{Label} & \textbf{\%}\\
\midrule
A deficiency of folate increases blood levels of homocysteine.    & Support   & 65.2 \\
Therapeutic anticoagulation in the trauma patient is safe.              & Refute    & 33.1 \\
Sternal fracture in growing children is a rare and often overlooked fracture.                           & NEI       & 1.7\\ 
\Xhline{1pt}
\end{tabular}%
}
\label{tab:scv}
\vspace{-5mm}
\end{table}

This task is designed to evaluate LLMs' understanding of unstructured text from research papers in a retrieval-augmented generation manner. Following the definition in \cite{wadden2020fact}, the task is to identify evidence related to the claim from the research literature and give a verdict of ``Support'', ``Refute'', or ``NEI'' (Not Enough Information) based on it. We reconstruct two high-quality biomedical datasets, PubMedQA \cite{jin2019pubmedqa} and SciFact \cite{wadden2020fact}, into one dataset for SCV, yielding a corpus constituted of abstracts derived from 5,664 scholarly articles, alongside a dataset comprising 1,385 biomedical claims, as shown in Table \ref{tab:scv}.

\subsection{Agent Task}
Building upon the atomic abilities, we propose a novel and comprehensive task, KGCheck. This task necessitates the initial application of the tool-query functionality to extract information from the KG. Subsequently, it employs the RAG approach or database access to procure evidence pertaining to the queried information, facilitating a determination of either ``Support'' or ``Refute''. This methodology enables agents to scrutinize the knowledge encapsulated within a large-scale KG, a venture of particular importance considering the prevalence of inaccuracies within numerous datasets, including prominent ones such as ImageNet \cite{deng2009imagenet}.

\begin{table}[!h]
\vspace{-3mm}
\caption{Four different checking types of KGCheck.}
\resizebox{\textwidth}{!}{%
\begin{tabular}{>{\centering\arraybackslash}m{2cm} >{\centering\arraybackslash}m{2cm} >{\centering\arraybackslash}m{3cm} >{\centering\arraybackslash}m{2cm} >{\centering\arraybackslash}m{3.5cm}}
\Xhline{1pt}
\multicolumn{2}{c}{\textbf{Check Type}} & \textbf{Graph} & \textbf{\%} & \textbf{Support: Refute} \\ \hline
\multirow{3}{*}{\centering Node} & Existence & 
\begin{tikzpicture}[baseline=(1.base), node distance={15mm}, main/.style = {draw, dashed, circle}]
\node[main] (1) {?};
\end{tikzpicture}
& 20.0 & \multirow{3}{*}{\centering 71.0:29.0} \\ \cline{2-4}
 & Attribute & 
\begin{tikzpicture}[baseline=(1.base), node distance={15mm}, main/.style = {draw, circle}]
\node[main] (1) {?};
\end{tikzpicture}
& 24.4 & \\ \hline
\multirow{3}{*}{\centering Triple} & Existing & 
\begin{tikzpicture}[node distance={15mm}, main/.style = {draw, circle}] 
\node[main] (1) {a};
\node[main] (2) [right of=1] {b}; 
\draw[->] (1) -- node[midway, above] {?} (2);
\end{tikzpicture}
& 25.8 & \multirow{3}{*}{\centering 46.4:53.6} \\ \cline{2-4}
 & Potential & 
\begin{tikzpicture}[node distance={15mm}, main/.style = {draw, circle}] 
\node[main] (1) {a};
\node[main] (2) [right of=1] {b}; 
\draw[->, dashed] (1) -- node[midway, above] {?} (2);
\end{tikzpicture}
& 29.8 & \\ \Xhline{1pt}
\end{tabular}%
}
\vspace{-2mm}
\label{tab:kgcheck}
\end{table}

For this task, we collect 225 high-quality annotated data, as illustrated in Table~\ref{tab:kgcheck}. Given the massive data encapsulated within KGs via triples, we delineate the inspection process into two distinct categories: single-node and triple-based. The single-node inspection is divided into node existence and attribute value assessments, while the triple inspection encompasses scenarios with and without edges between two nodes:
\begin{itemize}[leftmargin=*]
    \item \textbf{Existence}: We note that databases may excise entries during updates due to inaccuracies or redundancies, whereas KGs remain static post-construction, similar to LLMs in some respects. If nodes corresponding to obsolete entities persist in the KG, the label is ``Refute''; if they are congruent with real-time updated external databases, the label is ``Support''.
    \item \textbf{Attribute}: Our KG is characterized by high information density, with each node and edge encapsulating numerous attribute values, which we scrutinize for accuracy and completeness.
    \item \textbf{Existing Relationship}: We check whether existing edges contradict information from external, real-time updated databases and literature. If external knowledge corroborates the relationship, the label is ``Support''; conversely, it is ``Refute''.
    \item \textbf{Potential Relationship}: If a relationship is confirmed by databases or literature but is not represented in the KG, the label is ``Refute''; otherwise, it is ``Support''.
\end{itemize}
Despite utilizing the latest databases (as of May 2024), we identified errors within the KG, evidenced by 96 ``Refute'' annotations. These data are valuable and provide scenarios for agents to comprehend knowledge from heterogeneous sources and make \textbf{discoveries}.

\subsection{BKGAgent: A Simple Baseline}

\begin{figure}[!h]
    \centering
    \vspace{-0.3cm}
    \includegraphics[width=\textwidth]{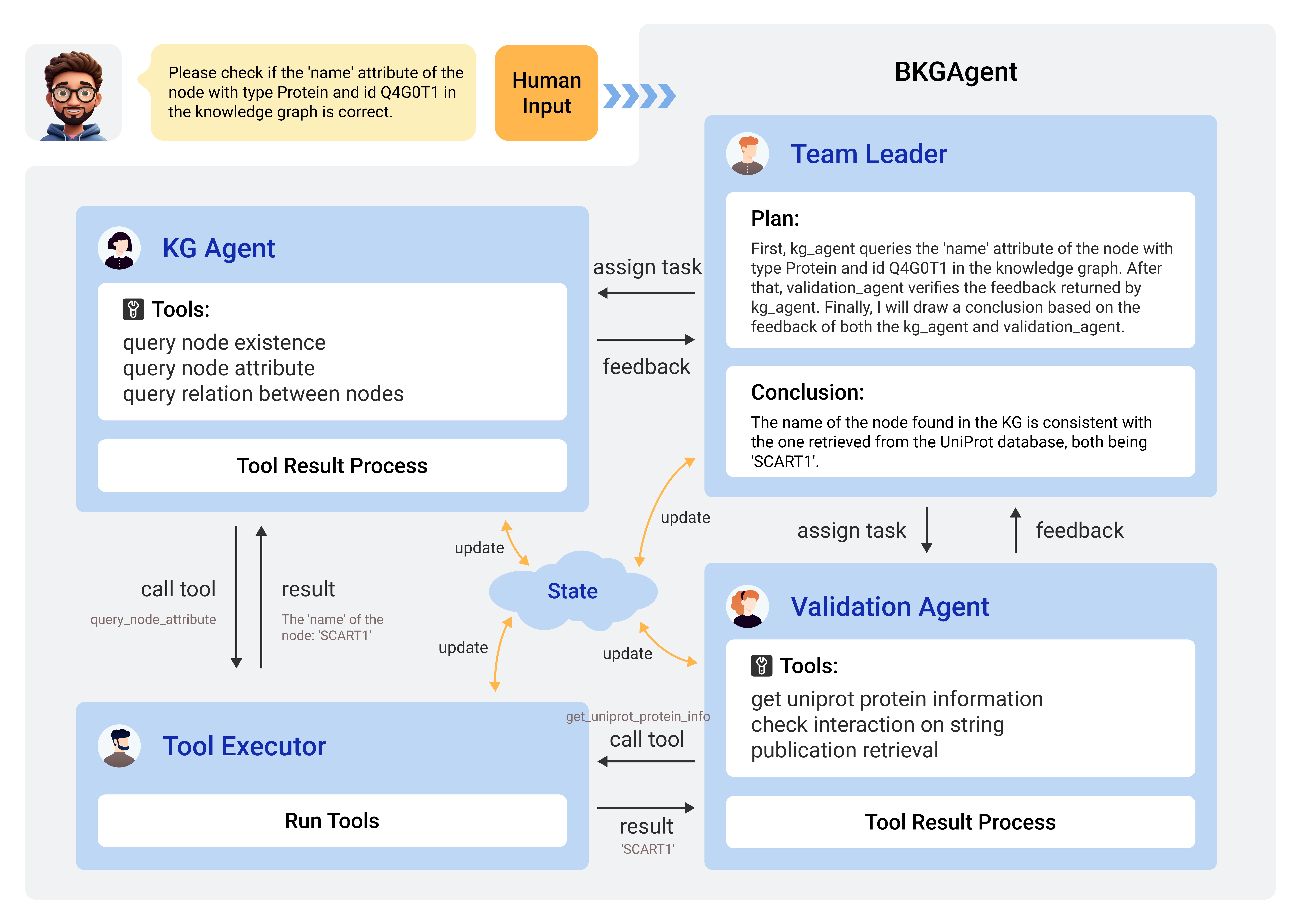}
    \vspace{-6mm}
    \caption{Framework of our BKGAgent.}
    \label{fig:multiagent}
    \vspace{-0.2cm}
\end{figure}

We propose a \textbf{b}iomedical \textbf{k}nowledge-\textbf{g}raph \textbf{a}gent~(BKGAgent), as shown in Figure \ref{fig:multiagent}. It's a multi-agent framework based on \emph{langgraph}~\cite{Chase_LangGraph_2023}, capable of retrieving information from knowledge graph and cross-validating its correctness with multiple information sources. Our framework is comprised of three agents: the team leader for the progress control, the KG agent for information retrieval from KG, and the validation agent for checking the correctness of the information from KG. This setup simulates the workflow of a human research team, where a leader supervises the assistants' work and makes the final decision based on their feedback. Additionally, the tool executor is solely responsible for executing functions, and is not based on LLMs.

When a user assigns a task, the leader initially breaks down the task and announces the plan. Then the KG agent is activated to retrieve task-related information from the KG. This involves specifying the tool and its arguments to the tool executor, interpreting the tool result, and communicating it back to the leader. After that, the validation agent is called to verify the information with a workflow similar to that of the KG agent. Finally, the leader will draw a conclusion and return it to the user.
\section{Experiments}

\subsection{Main Results and Analysis: Atomic Abilities}

\begin{table}[t]
\centering
\caption{Test set (standard) results of two easy tasks: KGQA, SCV. \textbf{Bold}/\underline{underline} and \textcolor{rr}{red}/\textcolor{bb}{blue} indicate the best and second in the subgroup and overall.}

\footnotesize
\renewcommand\tabcolsep{5pt}
\label{tab:main_results}
\resizebox{\columnwidth}{!}{
\begin{tabular}{@{}clccccccc@{}}
\toprule
\multirow{2}{*}{\begin{tabular}[c]{@{}c@{}}LLM\\ Type\end{tabular}}     & \multirow{2}{*}{Models}         & \multicolumn{4}{c}{KGQA}                                       & \multicolumn{3}{c}{SCV}                       \\ \cline{3-9} 
                                                                        &                                 & F1            & EM            & Executability &              & Accuracy      & Right Quotes  & Error         \\ \hline
\multirow{2}{*}{API}                                                    & GPT-4~\cite{Achiam2023GPT4TR}                          & \textcolor{rr}{\textbf{81.8}}& \textcolor{rr}{\textbf{79.2}}& \textbf{88.4}         &                                         & \underline{83.9} & \textcolor{rr}{\textbf{87.7}} & \textbf{0.4}        \\
                                                                        & GLM-4~\cite{GLM}                           & \underline{72.4}         & \underline{70.4}         & \underline{82.7}         &              & \textcolor{rr}{\textbf{86.9}} & \underline{86.5}   & \underline{0.6}          \\ \hline
\multirow{3}{*}{\begin{tabular}[c]{@{}c@{}}OSS\\ (Large)\end{tabular}}  & Qwen1.5-72B-Chat~\cite{qwen}                & \underline{74.7}         & \underline{72.2}         & \textcolor{bb}{\underline{96.1}}&                       & \underline{85.7}     & \underline{83.3}           & \textcolor{rr}{\textbf{0.1}}     \\
                                                                        & Llama-3-70B-Instruct~\cite{llama3modelcard}            & \textcolor{bb}{\textbf{80.7}}   & \textcolor{bb}{\textbf{77.8}}   & \textcolor{rr}{\textbf{97.0}}         &                                   & \textcolor{bb}{\textbf{85.9}}     & \textcolor{bb}{\textbf{86.6}}  & \textcolor{bb}{\underline{0.2}} \\
                                                                        & DeepSeek-LLM-67B-Chat~\cite{Bi2024DeepSeekLS}           & 69.6         & 66.8         & 86.3         &                                                  & 76.6     & 82.6           & 0.4 \\ \hline
\multirow{3}{*}{\begin{tabular}[c]{@{}c@{}}OSS\\ (Medium)\end{tabular}} & Qwen1.5-32B-Chat~\cite{qwen}                & \underline{64.6}         & \textbf{62.1}         & \textbf{83.0}         &                    & \textbf{79.7}     & \textbf{83.0}           & \underline{0.4}    \\
                                                                        & Qwen1.5-14B-Chat~\cite{qwen}                & \textbf{66.0}         & \underline{61.6}         & 78.7         &                             & \underline{66.1}     & \underline{67.4}           & \textcolor{bb}{\textbf{0.2}}    \\
                                                                        & Baichuan2-13B-Chat~\cite{Yang2023Baichuan2O}              & 43.7         & 42.0         & \underline{82.2}         &                                      & 26.3     & 35.8           & 33.6  \\ \hline
\multirow{2}{*}{\begin{tabular}[c]{@{}c@{}}OSS\\ (Small)\end{tabular}}  & Llama-3-8B-Instruct~\cite{llama3modelcard}             & \textbf{54.7}         & \textbf{51.3}         & \textbf{84.8}         &                       & \textbf{78.5}     & \textbf{83.3}          & \textbf{0.5}  \\
                                                                        & Qwen1.5-7B-chat~\cite{qwen}                 & \underline{44.5}         & \underline{40.3}         & \underline{77.9}         &              & 72.5    & 39.1          & \underline{2.2}          \\ \hline
\multirow{3}{*}{\begin{tabular}[c]{@{}c@{}}OSS\\ (MoE)\end{tabular}}    & Mixtral-8x7B-Instruct-v0.1~\cite{Jiang2024MixtralOE}      & \textbf{70.1}         & \textbf{67.9}         & \textbf{84.7}         &                       & \textbf{77.8}   & \textbf{82.5}           & \underline{2.3}  \\
                                                                        & Starling-LM-alpha-8x7B-MoE-GPTQ~\cite{starling2023} & 12.4         & 10.9         & 30.7         &                                                  & \underline{55.0}           & 56.2           & \textcolor{rr}{\textbf{0.1}}               \\
                                                                        & Qwen1.5-MoE-A2.7B-Chat~\cite{qwen}          & \underline{28.7}         & \underline{26.7}         & \underline{71.9}         &              & \underline{55.0}     & \underline{57.8}          & 3.0  \\ \hline
\end{tabular}
}
\vspace{-5mm}
\end{table}

The detailed experimental results of atomic abilities evaluation on LLMs are shown in Table~\ref{tab:main_results}, and we summarize our key findings as follows:
\begin{itemize}[leftmargin=*]
\item \textbf{Disparity between open-source and commercial API models.} Commercial API models like GPT-4 and GLM-4 generally outperform open-source models in several key metrics. GPT-4, for example, consistently achieves higher scores in both KGQA and SCV tasks, highlighting the advantage of proprietary training techniques and larger computational resources.
\item \textbf{Strong performance of open-source large models.} Some large OSS models, such as Llama-3-70B-Instruct, perform competitively, sometimes surpassing API models in specific metrics. Llama-3-70B-Instruct, in particular, excels in KGQA executability, suggesting that optimized training can enable open-source models to rival or exceed commercial counterparts.
\item \textbf{Model parameters do not always correlate with better performance.} In the OSS (Medium) and OSS (Small) categories, smaller models like Llama-3-8B-Instruct sometimes outperform larger models like Qwen1.5-32B-Chat in SCV tasks, indicating that model architecture, training data quality, and fine-tuning strategies significantly impact performance. Notably, Qwen1.5-14B-Chat outperforms Qwen1.5-32B-Chat in KGQA, suggesting the latter's pre-training may be insufficient.
\item \textbf{Domain-specific models lack transferability.} DeepSeek-LLM-67B-Chat excells in mathematical problems~\cite{Bi2024DeepSeekLS}, but underperforms in biomedical-related tasks, highlighting its lack of cross-domain transferability. This suggests that specialization in one area may compromise generalizability.
\item \textbf{Inconsistent performance of MoE models.} While Mixtral-8x7B-Instruct-v0.1 performs well in both KGQA and SCV tasks, other MoE models like Starling-LM-alpha-8x7B-MoE-GPTQ and Qwen1.5-MoE-A2.7B-Chat show significantly lower scores. This inconsistency suggests that the effectiveness of MoE models heavily depends on the implementation and integration of the expert models. Additionally, Mixtral-8x7B-Instruct-v0.1, though strong in main metrics, struggles with controlling response format, indicating that individual expert models still require improvement.
\item \textbf{Biomedical knowledge embedded in model parameters.} The new metric ``right quote'' for SCV assesses the alignment of retrieved quotes with ground truth evidence. Some models, such as GLM-4, Qwen1.5-72B-Chat, and Qwen1.5-7B-Chat, exhibit higher accuracy metrics than ``right quote'' metrics. This suggests these models can accurately assess input claims even without sufficient literature evidence, indicating they possess specialized biomedical knowledge.
\end{itemize} 

\begin{wrapfigure}{r}{0.5\linewidth}
\vspace{-1.2em}
\centering
\includegraphics[width=1\linewidth]{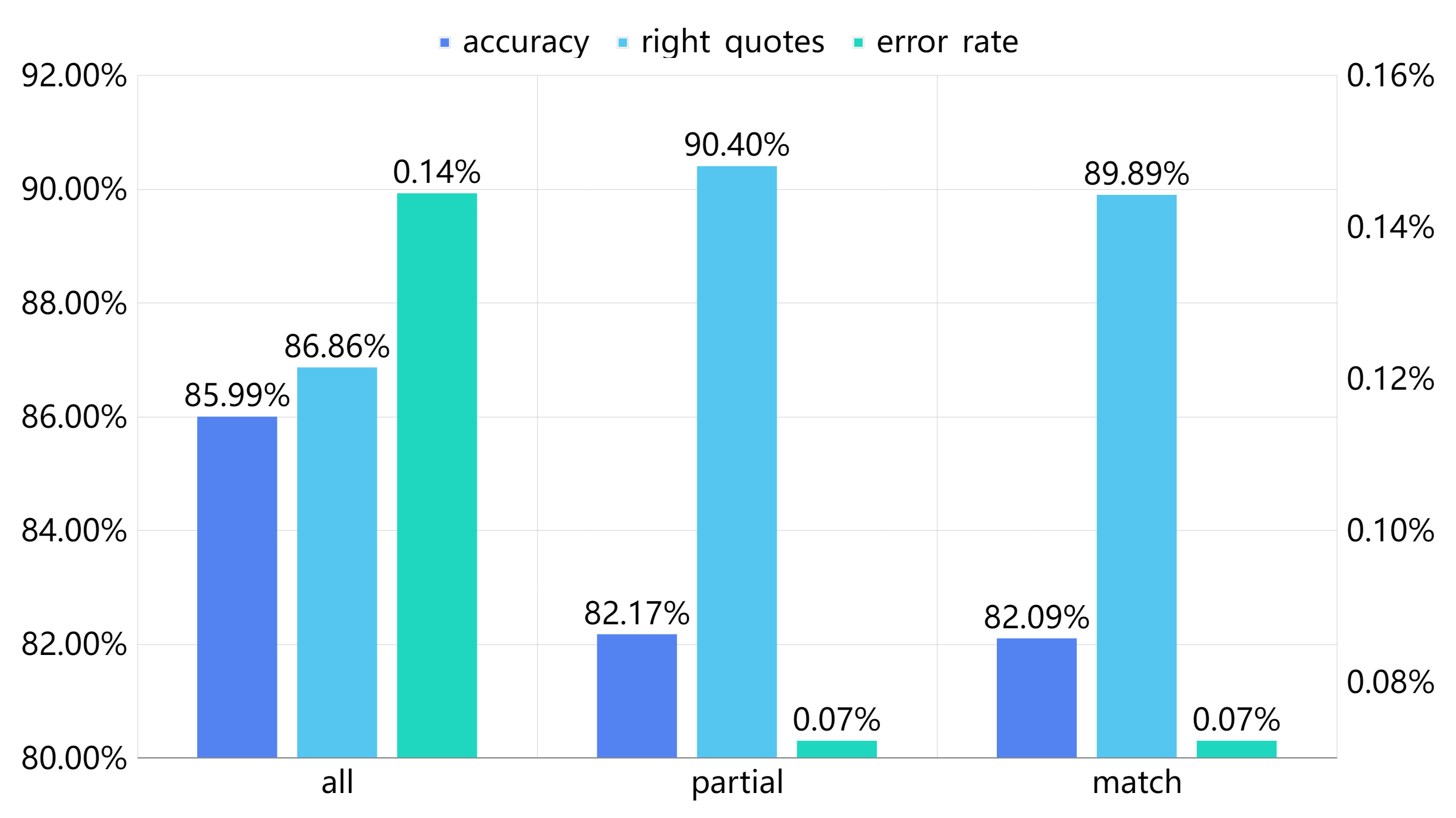}
  \vspace{-1.9em}
	\caption{Llama-3-70B-Instruct's performance in RAG across different scopes of literature.}
	\label{fig:rag_scope}
\vspace{-0.7em}
\end{wrapfigure}

We also conduct an ablation experiment on three scopes of RAG, as shown in Figure~\ref{fig:rag_scope}, where `all' refers to the abstract of 5,664 articles, `partial' denotes the 1,888 abstracts containing ground truth evidence of claims, and `match' corresponds to the abstracts of the ground truth evidence for the claims. Interestingly, we observe an unexpected phenomenon where the model's performance in the `match' setting only increases in terms of the right quotes metric, while the accuracy metric actually decreases. In the `all' setting, we initially thought that irrelevant literature would introduce interference, but the accuracy metric, on the contrary, increases. This suggests that there is a potential connection among the extensive literature, where large models exhibit a form of \textbf{``analogical reasoning''}. This provides us with insights for conducting extensive literature research in simulating human scientific research.

\subsection{Main Results and Analysis: \ouragent{}}

\begin{table}[]
\caption{Process and outcome metrics of \ouragent{} on KGCheck task, where 
\label{tab:bmcagent}
\textbf{bold} numbers denote better scores among the two models while \textcolor{gg}{green} numbers denote 100\% acuracy.}
\resizebox{\textwidth}{!}{%
\begin{tabular}{lcccccc}
\Xhline{1pt}
\multirow{2}{*}{Model} & \multicolumn{2}{c}{KG Query Task} & \multicolumn{2}{c}{Validation Task} & \multicolumn{2}{c}{Final Result} \\ \cline{2-3} \cline{4-5} \cline{6-7}
                       & Tool Selection   & Executability  & Tool Selection    & Execurability   & Exact Match    & Executability   \\ \hline
\multicolumn{1}{c}{} & \multicolumn{6}{c}{\cellcolor[HTML]{D6F3F2}Web Database KGCheck}          \\
GPT-4~\cite{Achiam2023GPT4TR}        & 65.0      & 65.0    & 88.1    & 88.8    & \textbf{64.5}    & 96.9    \\
Llama-3-70B-Instruct~\cite{llama3modelcard}      & \textbf{96.9}      & \textbf{96.9}    & \textbf{97.5}    & \textbf{97.5}    & 38.1    & \textcolor{gg}{\textbf{100.0}}    \\
\multicolumn{1}{c}{} & \multicolumn{6}{c}{\cellcolor[HTML]{D6F3F2}Publication Database KGCheck} \\
GPT-4~\cite{Achiam2023GPT4TR}        & 67.7      & 67.7    & 69.2    & 69.2    & \textbf{61.5}    & \textbf{95.4}    \\
Llama-3-70B-Instruct~\cite{llama3modelcard}      & \textcolor{gg}{\textbf{100.0}}      & \textcolor{gg}{\textbf{100.0}}    & \textbf{95.4}    & \textbf{95.4}    & 41.5    & 63.1    \\ \Xhline{1pt}
\end{tabular}%
}
\end{table}
The experiment results including both process and final answer are shown in Table~\ref{tab:bmcagent}, where the executability metric of the assistant agent means whether it is correctly activated. The exact match score of the final result agrees with what we observed in the atomic abilities experiments: GPT-4 surpasses Llama-3-70B-Instruct in the final result, which shows the best performance in the atomic abilities experiments. We discovered some interesting phenomena based on the analysis of \ouragent{} chat history.

\begin{figure}[]
    \centering
    \includegraphics[width = \textwidth]{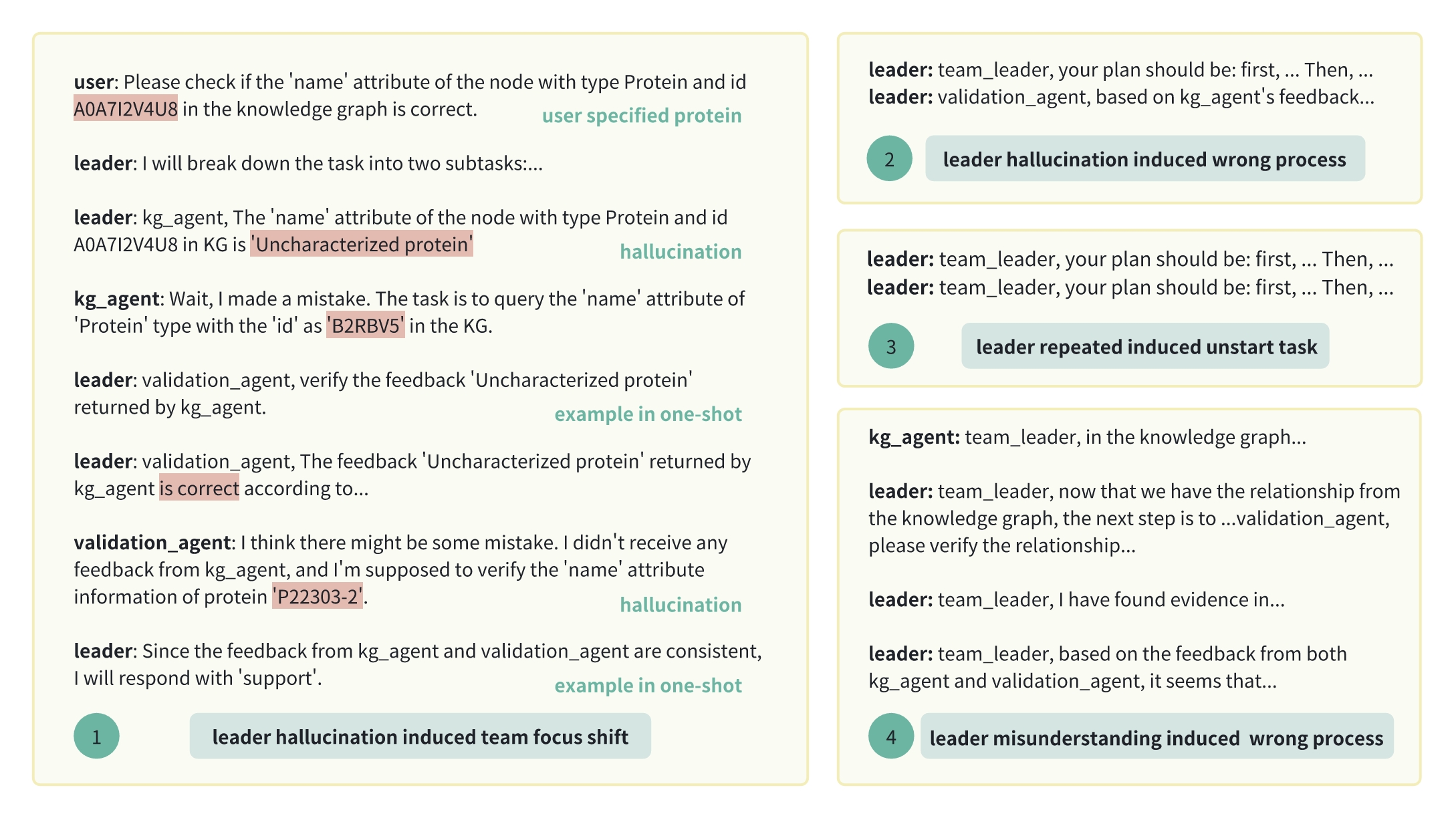}
    \vspace{-0.5cm}
    \caption{
    \textbf{Error analysis.} Here, we show a failure case due to a leader's various mistakes: the hallucination of the leader misleading the later task or using the wrong process, the leader producing unnecessary repeated tasks and misunderstanding leads to the wrong process. 
    }
    \vspace{-0.7cm}
    \label{fig:errorcase}
\end{figure}

\begin{itemize}[leftmargin=*]
\item \textbf{Leader agent performance significantly influences the team behavior.} While the behavior of the assistant agents like KG agent can be modified by the leader's instruction, the leader itself lacks action-related feedback from others, meaning that a bad decision made by the leader may lead to a catastrophe. We found four common error cases induced by the leader, as shown in Figure~\ref{fig:errorcase}. Among these cases, the leader either fails to give effective instructions to team members, becomes trapped in repeated self-talks, or attempts to perform the tasks that are meant for the assistants.
\item \textbf{Assistant agents can always select the right tool.} As shown in the results, assistant agents's executability is almost the same as its right tool selection rate, which means once the agent is correctly woken up, it has a good chance of selecting the right tool.
\item \textbf{Good process results do not necessarily mean a good final result.} Surprisingly, the agent based on Llama surpasses the one based on GPT in terms of assistant agents’ tool selection. However, the final result is far from satisfactory. We watch for the chat history of the Llama-based agent and find out that while the right answer can be found in the history of the agent in many cases, the team starts chatter instead of sending the result to the user thus leading to a final wrong answer.
\item \textbf{Balanced abilities of \ouragent{}.} We classify our tasks into two categories by the way the agent checks the information from KG, namely the web database (UniProt~\cite{uniprot2023uniprot} and STRING~\cite{szklarczyk2023string}) check task and the publication database check task. The performance of them is separately evaluated. Considering the exact match score of the final result, both the GPT-based agent and Llama-based agent show similar performance in both types of tasks, indicating \ouragent{} does not show any preference for specific tasks.
\end{itemize}

\subsection{Capability Analysis}
Biomedical agents are not widely employed, as most tasks rely solely on models (refer to Appendix~\ref{appendix:other_rw}). Additionally, most agents are primarily focused on QA tasks. For example, MedAgents\cite{tang2024medagents} employs a Multi-disciplinary Collaboration (MC) framework and was tested on nine QA datasets. General agents such as the multi-agent framework AutoGen and the single-agent framework AutoGPT are capable of performing web searches and database retrieval enhancements through tool calls. However, since accessing knowledge graphs and biomedical information is beyond daily usage, and the information directly gained from the web being unreliable, it is challenging to achieve the same level of biomedical information retrieval and verification effectiveness using existing general agent frameworks.

As presented in Table~\ref{tab:agentcomparison}, our framework encompasses the basic abilities, especially for retrieval in the biomedical domain. The range of our information for KG result validation includes the UniProt database, STRING database, and publications selected from PubMed, which surpasses the limited information from a simple web search.

\begin{wraptable}{r}{0.5\textwidth}
\vspace{-7mm}
\centering
  \caption{Comparison of capabilities for \ouragent{} and other frameworks. 
  }
  \vspace{1mm}
  \label{tab:agentcomparison}
  \centering
  \setlength{\tabcolsep}{1pt}
  \begin{tabular}{llcccc}
    \Xhline{1pt}
   Scenario &Framework     & MA     & KGq & IR & IF\\
    \midrule
    \multirow{8}{*}{Common}  &AutoGPT\footnotemark[1]  &\XSolidBrush &\XSolidBrush &\XSolidBrush &\XSolidBrush  \\
     &AutoGen~\cite{wu2023autogen}  &\CheckmarkBold &\XSolidBrush &\XSolidBrush &\CheckmarkBold  \\
     &HuggingGPT~\cite{shen2023hugginggpt}  &\CheckmarkBold &\XSolidBrush &\XSolidBrush &\CheckmarkBold  \\
     &OpenAgents~\cite{xie2023openagents}  &\CheckmarkBold &\XSolidBrush &\XSolidBrush &\CheckmarkBold  \\
     &AgentVerse~\cite{chen2023agentverse}  &\CheckmarkBold &\XSolidBrush &\XSolidBrush &\CheckmarkBold  \\
     &Xagent~\cite{xagent2023}  &\CheckmarkBold &\XSolidBrush &\XSolidBrush &\CheckmarkBold  \\
     &BabyAGI~\cite{Yoheinakajima}  &\CheckmarkBold &\XSolidBrush &\XSolidBrush &\CheckmarkBold  \\
     \cline{1-6}
     \midrule
    \multirow{3}{*}{Domain} &MedAgents~\cite{Tang2023MedAgentsLL} &\CheckmarkBold &\XSolidBrush &\XSolidBrush &\CheckmarkBold \\
     &gpt-researcher\footnotemark[2] &\CheckmarkBold &\XSolidBrush &\XSolidBrush &\CheckmarkBold \\
     &\textcolor{gg}{\textbf{\ouragent{}(ours)}} &\textcolor{gg}{\CheckmarkBold}  &\textcolor{gg}{\CheckmarkBold}  &\textcolor{gg}{\CheckmarkBold} &\textcolor{gg}{\CheckmarkBold} \\
     \Xhline{1pt}
     \multicolumn{5}{l}{\small MA=multi-agent; KGq=KG-query;} \\
     \multicolumn{5}{l}{\small IR=information retrieval; IF=intervention free} \\
     \vspace{-5mm}
     \end{tabular}%
\vspace{-3mm}
\end{wraptable}

\section{Conclusion}
We present \ours{}, an interactive benchmark that encompasses the KGCheck task with two atomic capabilities for assessment: KGQA and SCV. KGCheck offers agents a valuable scenario for detecting knowledge hallucination within large-scale data, akin to the experience of researchers making discoveries amidst voluminous literature in the real world. We conduct evaluations of these two atomic capabilities across 13 LLMs and select the top-performing open-source model, Llama-3-70B-Instruct, and API-based model, GPT-4, to construct \ouragent{}—--a multi-agent system serving as the baseline. Comparisons with existing general and biomedical agents revealed their poor performance due to the absence of certain process capabilities, thereby demonstrating the challenging nature of our benchmark. We expect \ours{} to serve as a valuable endeavor towards paving the path for biomedical agents to become AI scientists.

\footnotetext[1]{\url{https://news.agpt.co/}}
\footnotetext[2]{\url{https://gptr.dev/}}

\mypara{Limitations and Future Work.}
In KGCheck, we guide agents to identify knowledge-based errors within the KG by providing them with specific instructions. This process involves atomic-level inspections from single nodes to triples, which agents could potentially implement autonomously. Future work will explore how agents can autonomously conduct real-time error detection in large datasets by leveraging logic rules and prior knowledge. Additionally, our case studies reveal that despite agents' current underperformance in KGCheck tasks (as shown in Figure~\ref{fig:errorcase}), their occasional correct responses are often due to chance, attributed to the binary nature of answers which inherently allows a 50\% success rate from random guesses. More granular metrics instead of exact match score should be introduced for a more accurate evaluation, which will be part of future work.

\bibliographystyle{unsrt}
\bibliography{references}

\newpage
\part{Appendix}
\parttoc
\newpage

\appendix

\newpage

\section{Dataset and Code}
\label{ddaccess}
\subsection{Dataset and Code Access}
Following NeurIPS Dataset and Benchmark Track guidelines, our data and code are available at \url{https://github.com/westlake-autolab/BioKGBench} under MIT license. More details are also documented at this address. The Croissant metadata record is available at \url{https://huggingface.co/api/datasets/AutoLab-Westlake/BioKGBench-Dataset/croissant}.

\subsection{Responsibility Statement}
We declare that we assume full responsibility for any violations of rights. We confirm that the data used in this paper is properly licensed and ensure that its use complies with all relevant laws, regulations, and licensing agreements.

\section{Datasheet}
Here, we provide a detailed description of our benchmark dataset, following the guidelines of the ``Datasheet for Datasets''~\cite{gebru2021datasheets}.
\subsection{Motivation}
Our benchmark dataset was created to address the lack of benchmarks for evaluating biomedical agents from the perspective of an ``AI scientists''. In~\cite{gao2024empowering}, it is stated that ``AI scientists'' can be realized as AI agents supported by humans, LLMs, ML models, and other tools like experimental platforms that cooperate to solve complex tasks. However, the current evaluation methods for biomedical agents remain unexplored, limited to simple question-answering tasks, which not only fail to avoid the hallucination problem inherent in solely relying on LLMs but also do not assess agents' abilities to utilize external tools and knowledge bases. Our proposed benchmark fills this gap by designing tasks ranging from easy to hard, based on two atomic capabilities: tool-query and memory-RAG. These tasks evaluate the agents' ability to leverage external support, including external knowledge and tools, when handling large-scale and multi-modal data. Moreover, Our data collected for KGCheck, the most challenging task, provides scenarios for agents to comprehend knowledge from heterogeneous sources and make discoveries.

The conception and construction of this dataset were jointly completed by the biomedical experts and AI researchers listed in the author list.

\subsection{Composition}

We provide the necessary data for constructing the knowledge graph, literature for RAG, as well as the development and test data for KGQA, SCV, and KGCheck, where knowledge graph and literature are external knowledge sources provided for agent.

The knowledge graph is derived from a subset of Clinical Knowledge Graph (CKG)~\cite{santos2022knowledge}. We specifically retain twelve key node types to ensure there is no loss of generality: Protein, Biological process (BP), Molecular function (MF), Cellular component (CC), Amino acid sequence, Tissue, Protein structure, Pathway, Modified protein, Modification, Disease, and Gene. The statistics of the triples in our knowledge graph are presented in Table~\ref{tab:kg_stats}. Detailed information stored in our knowledge graph is shown in Table~\ref{tab:kgcontent}.

\begin{table}[!ht]
    \centering
\aboverulesep=0ex
\belowrulesep=0ex
    \caption{The data statistics of our knowledge graph drawn from CKG. 
    \label{tab:kg_stats}}
    \resizebox{\textwidth}{!}{
    \begin{tabular}{llll}
    \toprule
        Head Node & Tail Node & Relation & Number \\ 
        \midrule
        \multirow{1}{*}{Protein} & Protein\_structure & HAS\_STRUCTURE & 271,512 \\ 
        ~ & Amino\_acid\_sequence & HAS\_SEQUENCE & 20,598 \\ 
        ~ & Cellular\_component & ASSOCIATED\_WITH & 3,796,383 \\ 
        ~ & Tissue & ASSOCIATED\_WITH & 7,117,321 \\ 
        ~ & Disease & ASSOCIATED\_WITH & 5,882,437 \\ 
        ~ & Molecular\_function & ASSOCIATED\_WITH & 85,013 \\ 
        ~ & Biological\_process & ASSOCIATED\_WITH & 153,219 \\ 
        ~ & Protein & ACTS\_ON & 985,376 \\ 
        ~ & Pathway & ANNOTATED\_IN\_PATHWAY & 357,739 \\ 
        ~ & Protein & CURATED\_INTERACTS\_WITH & 3,448 \\ 
        ~ & Modified\_protein & HAS\_MODIFIED\_SITE & 4,498 \\ 
        \midrule
        Disease & Disease & HAS\_PARENT & 16,058 \\ 
        \midrule
        Modified\_protein & Protein & IS\_SUBSTRATE\_OF & 6,633 \\ 
        ~ & Modification & HAS\_MODIFICATION & 4,559 \\ 
        \midrule
        Gene & Protein & TRANSLATED\_INTO & 179,854 \\ 
        \midrule
        Biological\_process & Biological\_process & HAS\_PARENT & 49,081 \\ 
        \midrule
        Molecular\_function & Molecular\_function & HAS\_PARENT & 13,659 \\ 
        \bottomrule
    \end{tabular}
    }
\end{table}

\begin{table}[!h]
\centering
\caption{Details of the information stored in the nodes of our knowledge graph.}

\renewcommand{\arraystretch}{1.2} 
\small 
\begin{tabularx}{\textwidth}{>{\raggedright}p{2.3cm} >{\raggedright}p{2.5cm} >{\raggedright\arraybackslash}X}
\toprule
\textbf{Entity Type} & \textbf{Content} & \textbf{Example}\\
\midrule
Protein & name, id, accession, synonyms & \{`name': `PLEKHG6', `id': `Q3KR16', `accession': `PKHG6\_HUMAN', `synonyms': [`PKHG6\_HUMAN', `PLEKHG6', `9606.ENSP00000380185', `ENSG00000008323'], `taxid': `9606'\} \\
Disease & name, description, id(DOID), type, synonyms & \{`synonyms': [`sulfamethoxazole allergy', `SMX allergy', `SMZ allergy', `sulphamethoxazole allergy'], `name': `sulfamethoxazole allergy', `description': `A drug allergy that has\_allergic\_trigger sulfamethoxazole. [url:\url{https://www.ncbi.nlm.nih.gov/pubmed/7602118}]', `id': `DOID:0040016', `type': `-26'\} \\
Protein structure & link, id, source & \{`link': \url{http://www.rcsb.org/structure/6XWD}, `id': `6XWD', `source': `Uniprot'\} \\
Amino acid sequence   &  sequence, header, source, id, size & \{`sequence': `LRGAAGRLGGGLLVL', `size': `15', `header': `tr|A0A0A0MTA2|A0A0A0MTA2\_HUMAN', `source': `UniProt', `id': `A0A0A0MTA2'\} \\
Cellular component    &  name, description, id, type, synonyms  & \{`name': `Golgi membrane', `description': `The lipid bilayer surrounding any of the compartments of the Golgi apparatus. [GOC:mah]', `id': `GO:0000139', `type': `-21', `synonyms': [`Golgi membrane']\} \\
Molecular function    & name, description, id(GO), type, synonyms & \{`name': `polymeric immunoglobulin binding', `description': `Interacting selectively and non-covalently with a J-chain-containing polymeric immunoglobulin of the IgA or IgM isotypes. [GOC:add, ISBN:0781735149]', `id': `GO:0001790', `type': `-21', `synonyms': [`polymeric immunoglobulin binding']\}  \\
Biological process    & name, description, id(GO), type, synonyms & \{`synonyms': [`mitochondrion inheritance'], `name': `mitochondrion inheritance', `description': `The distribution of mitochondria, including the mitochondrial genome, into daughter cells after mitosis or meiosis, mediated by interactions between mitochondria and the cytoskeleton. [GOC:mcc, PMID:10873824, PMID:11389764]', `id': `GO:0000001', `type': `-21'\}  \\
Pathway               & name, description, linkout, id, source & \{`name': `Antigen processing: Ubiquitination \& Proteasome degradation', `description': `Antigen processing: Ubiquitination \& Proteasome degradation', `linkout': \url{https://reactome.org/PathwayBrowser/##/R-HSA-983168}, `id': `R-HSA-983168', `source': `Reactome'\}  \\
Tissue                & name, description, id, type, synonyms & \{`name': `stratum basale', `description': `The deepest layer, as of the epidermis or the endometrium. In the epidermis it is a single layer of cells. In the endometrium it provides the regenerative tissue after menstrual loss of the functional layer. [Dorlands\_Medical\_Dictionary:MerckMedicus]', `id': `BTO:0004680', `type': `-25', `synonyms': [`stratum basale', `basal layer']\}  \\
Modified protein      & sequence\_window, protein, position, source, id, residue & \{`sequence\_window': `MEPAPARsPRPQQDP', `protein': `P29590', `position': `8', `source': `SIGNOR', `id': `P29590\_S8-p', `residue': `S'\}  \\
Modification          & synonyms, name, description, id, type & \{`synonyms': [`Unimod', `Source: ``none'''], `name': `Unimod', `description': `Entry from Unimod. [PubMed:18688235]', `id': `MOD:00003', `type': `-41'\}  \\
Gene                & taxid, synonyms, name, id, family & \{`taxid': `9606', `synonyms': [`54843', `ENSG00000137501', `OTTHUMG00000166977', `uc010rti.4', `AJ303364'], `name': `synaptotagmin like 2', `id': `SYTL2', `family': ```Protein phosphatase 1 regulatory subunits|Synaptotagmin like tandem C2 proteins'''\}  \\
\bottomrule
\end{tabularx}
\label{tab:kgcontent}
\end{table}

\begin{figure}[htbp]
    \centering
    \subfigure[Distribution of literature IF (Impact Factor)]{
        \includegraphics[width=0.45\textwidth]{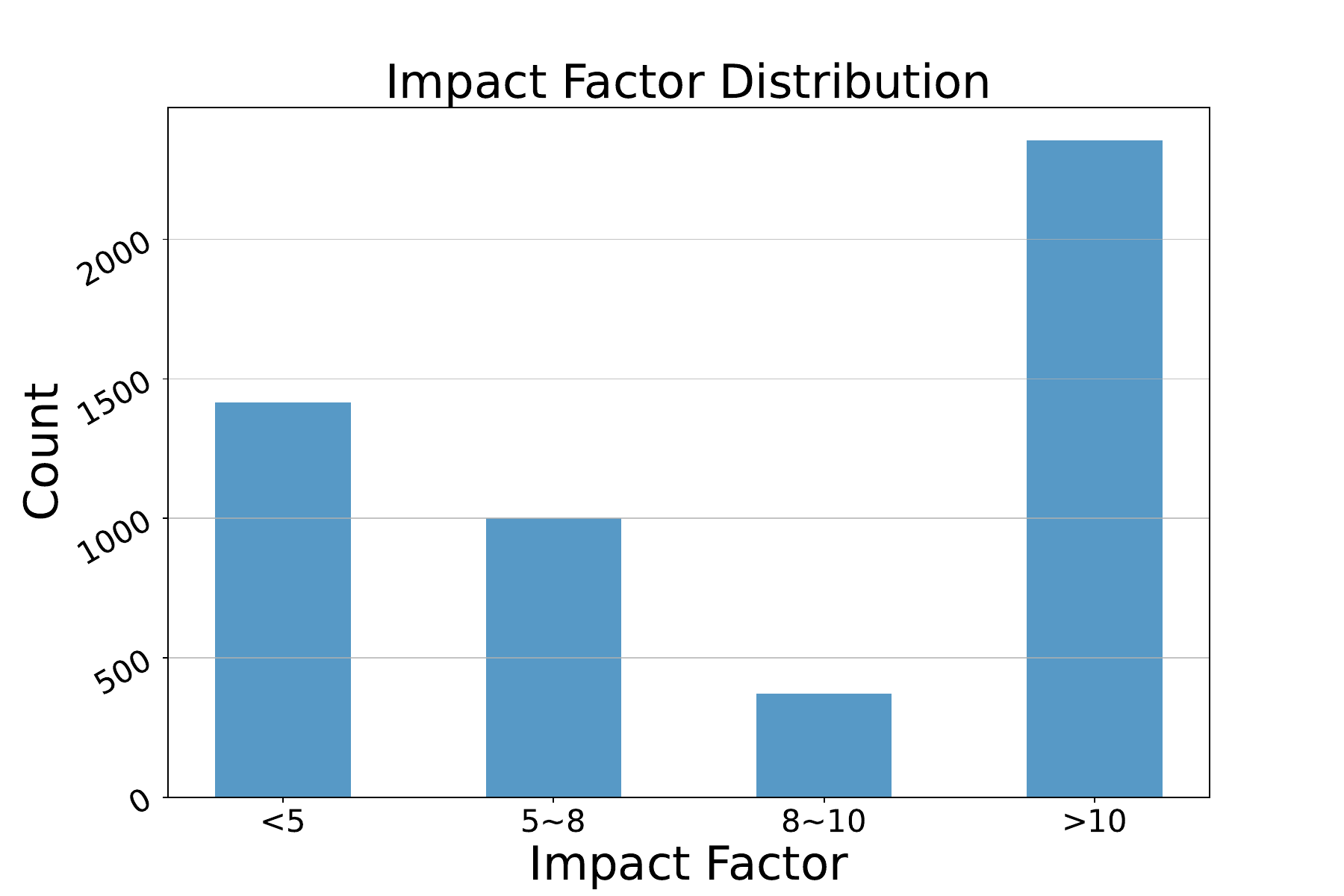}
        \label{fig:abs_if}
    }
    \hfill
    \subfigure[Distribution of literature publication dates]{
        \includegraphics[width=0.45\textwidth]{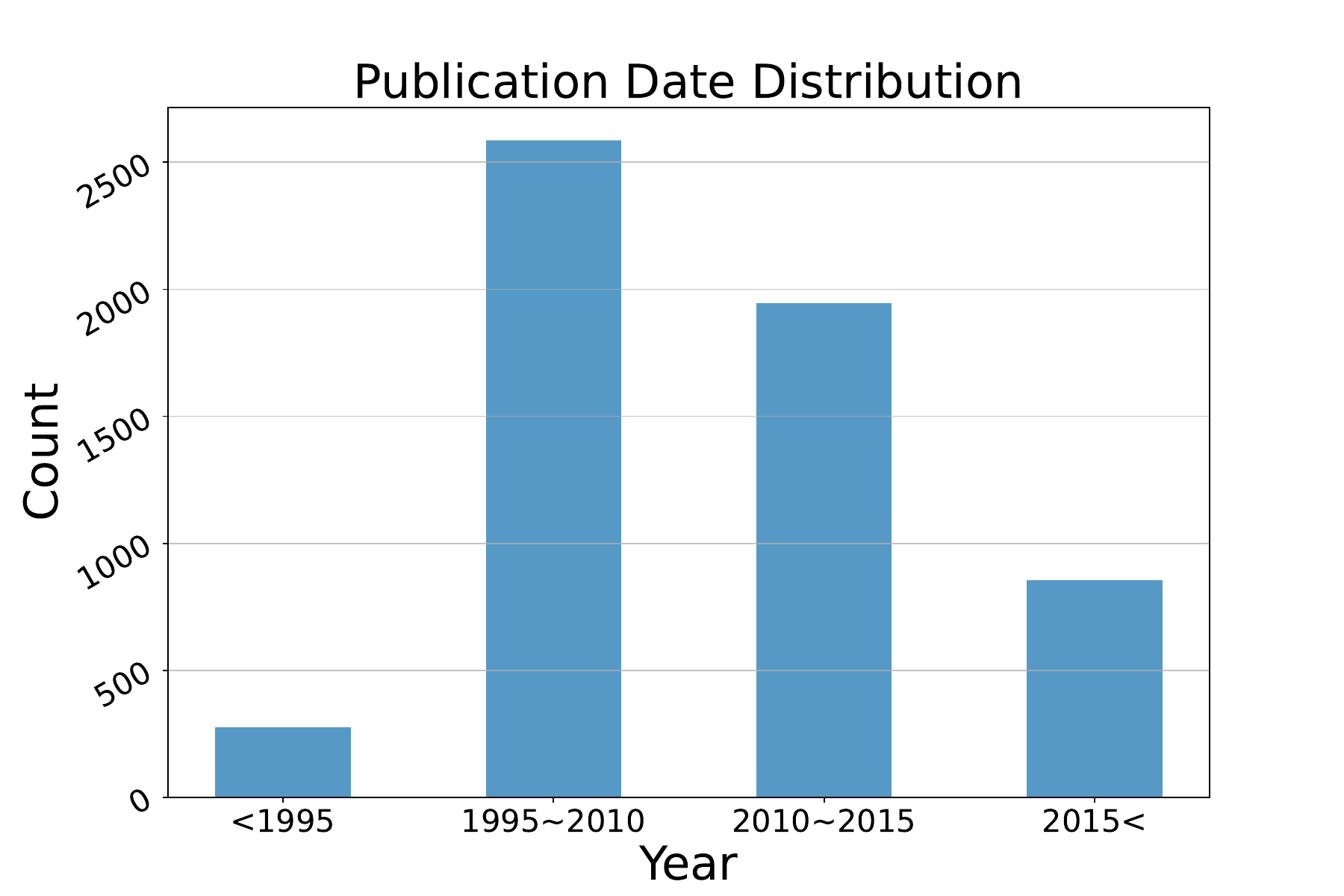}
        \label{fig:abs_pubdate}
    }
    \caption{Specific information of the 5,664 abstracts}
    \label{fig:abs_detail}
\end{figure}

\begin{figure}[htbp]
    \centering
    \subfigure[Distribution of literature IF (Impact Factor)]{
        \includegraphics[width=0.45\textwidth]{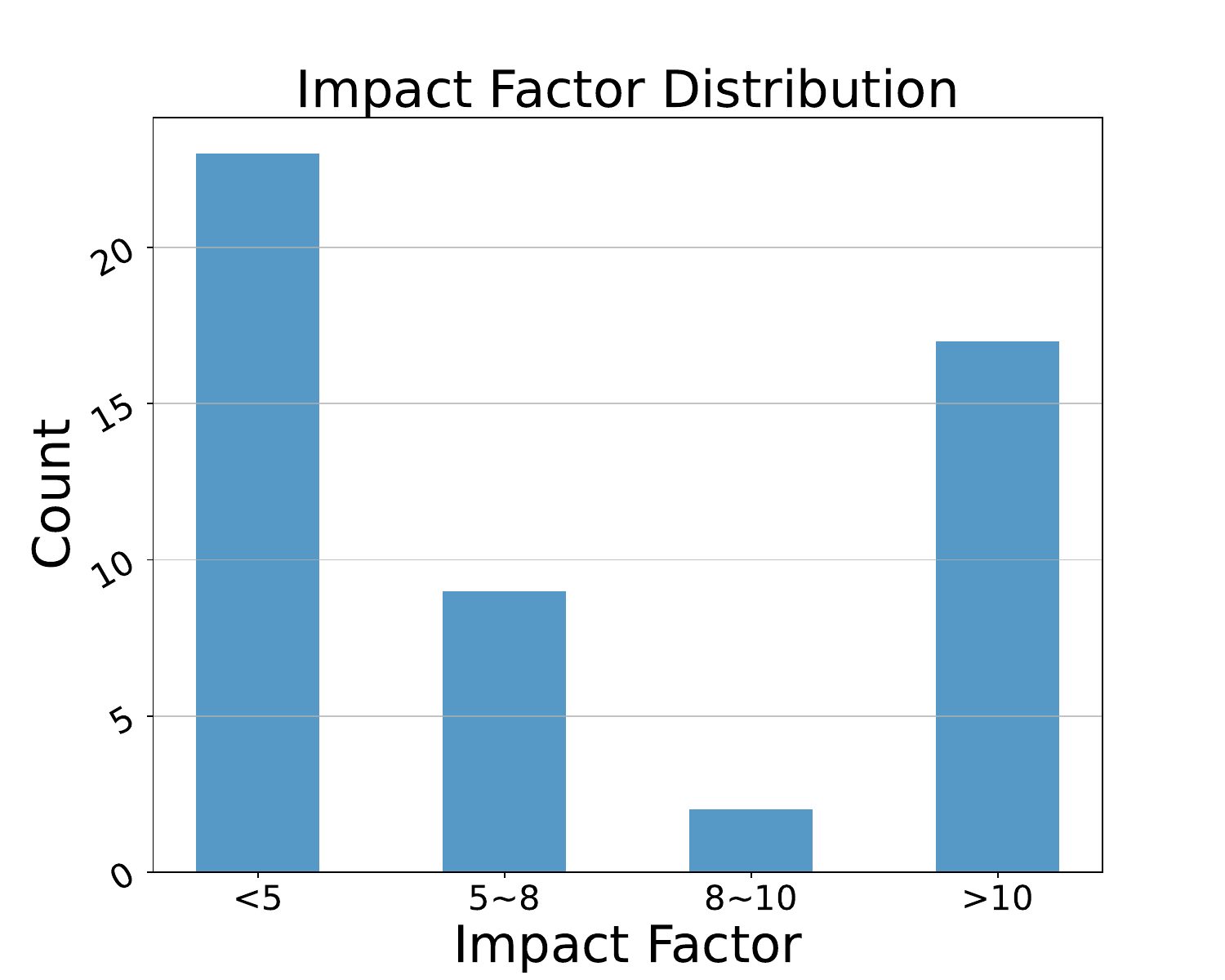}
        \label{fig:full_if}
    }
    \hfill
    \subfigure[Distribution of literature publication dates]{
        \includegraphics[width=0.45\textwidth]{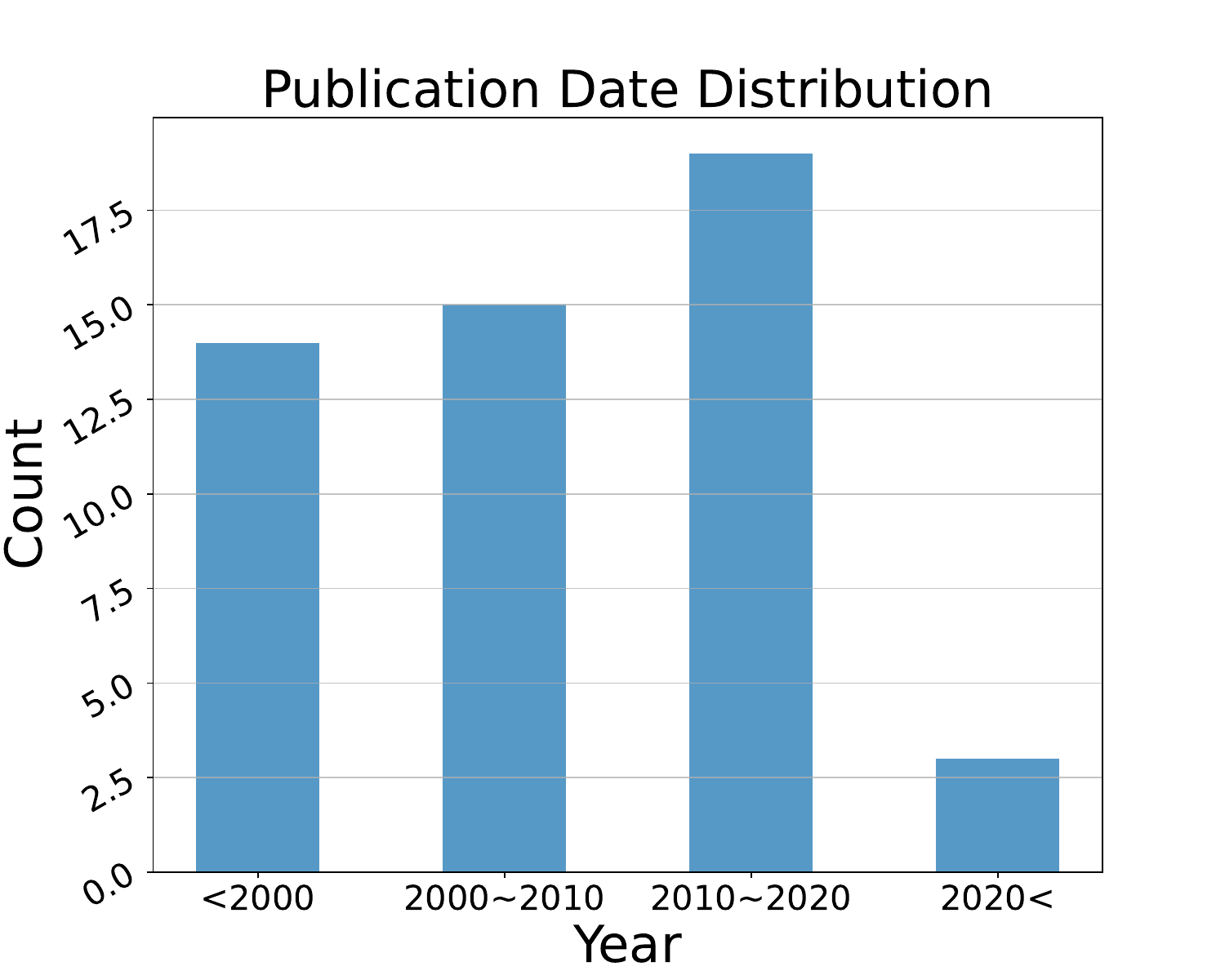}
        \label{fig:full_pubdate}
    }
    \caption{Specific information of the 51 full papers}
    \label{fig:full_detail}
\end{figure}

Besides the knowledge graph, literature also serves as an external source of knowledge. We provide a corpus of 5,664 abstracts (under ODC-By 1.0) for SCV and 51 full papers for KGCheck. The 5,664 abstracts are sourced from existing datasets SciFact~\cite{wadden2020fact} (under CC BY 4.0) and PubMedQA~\cite{jin2019pubmedqa} (under MIT license), while the 51 full papers, all of which are open access, were selected by experts based on entries in the IntAct~\cite{orchard2014mintact} database and CKG. Table~\ref{tab:abs} summarizes the sources of the abstracts, and Figure~\ref{fig:abs_detail} describes the literature with more details. Table~\ref{tab:fulltext} summarizes the sources of the 51 full papers, and Figure~\ref{fig:full_detail} provides more details about the literature. 

\begin{table}[!ht]
\centering
\caption{The 5,664 papers come from 1,484 journals. Due to space limitations, we only list the names of journals with an IF greater than 70 and use `others' to represent journals with an IF less than 70.}
\label{tab:abs}
\begin{tabular}{p{0.6\textwidth} p{0.1\textwidth}}
\toprule
\textbf{Journal}                                                                & \textbf{Count} \\ \midrule
Nature reviews. Microbiology           & 3              \\
CA: a cancer journal for clinicians    & 3              \\
The Lancet. Infectious diseases        & 3              \\
Nature reviews. Drug discovery         & 5              \\
Nature reviews. Molecular cell biology & 14             \\
Nature reviews. Immunology             & 21             \\
Lancet (London, England)               & 46             \\
The New England journal of medicine    & 46             \\
BMJ (Clinical research ed.)            & 90             \\
JAMA                                   & 113            \\
Nature medicine                        & 138            \\ \Xhline{0.05pt}
others                                 & 5182           \\ \midrule
Total                                  & 5664           \\ \bottomrule
\end{tabular}
\end{table}
\begin{table}[!ht]
\centering
\caption{Journal distribution of the 51 full papers}
\label{tab:fulltext}
\begin{tabular}{p{0.6\textwidth} p{0.1\textwidth}}
\toprule
\textbf{Journal}                                                                & \textbf{Count} \\ \midrule
Brain research                                                                  & 1              \\
Molecular \& cellular proteomics : MCP                                          & 1              \\
Aging cell                                                                      & 1              \\
Cell reports                                                                    & 1              \\
PloS one                                                                        & 1              \\
Genes to cells : devoted to molecular \& cellular mechanisms                    & 1              \\
EMBO reports                                                                    & 1              \\
IUBMB life                                                                      & 1              \\
The Journal of allergy and clinical immunology                                  & 1              \\
The EMBO journal                                                                & 1              \\
Open biology                                                                    & 1              \\
Nature communications                                                           & 1              \\
Pigment cell \& melanoma research                                               & 1              \\
Molecular biology of the cell                                                   & 1              \\
Mobile DNA                                                                      & 1              \\
Journal of molecular biology                                                    & 1              \\
Nutrients                                                                       & 1              \\
Biological research for nursing                                                 & 1              \\
Genes \& development                                                            & 1              \\
Developmental cell                                                              & 1              \\
Bone                                                                            & 1              \\
Cancers                                                                         & 1              \\
Animals : an open access journal from MDPI                                      & 1              \\
Nucleic acids research                                                          & 2              \\
Molecular cell                                                                  & 2              \\
Scientific reports                                                              & 3              \\
Proceedings of the National Academy of Sciences of the United States of America & 3              \\
Molecular and cellular biology                                                  & 4              \\
Cell                                                                            & 4              \\
The Journal of biological chemistry                                             & 10             \\ \midrule
Total                                                                           & 51             \\ \bottomrule
\end{tabular}
\end{table}

For evaluation, we carefully collected 698, 1385, and 225 instances for KGQA, SCV, and KGCheck respectively. These datasets are split into development (dev) and test sets at an approximate ratio of 1:10. The dev data is intended for users to debug and fine-tune their evaluation code, while the test data is reserved for the final assessment. Each instance includes both input and output (ground truth answer or label) pairs, with additional information to make the data easier to understand. The dataset for SCV is reconstructed from well-known existing datasets SciFact and PubMedQA, while the rest is self-contained. The dataset represents a carefully selected sample of instances from a larger set, ensuring a comprehensive and representative coverage of the key aspects.

\subsection{Collection Process}
Biomedical experts and AI researchers listed in the author list were invloved in the data collection process. The collection process for different tasks varies:

\textbf{KGQA.} The collection process can be summarized into two steps: manually constructing question templates and automatically generating questions:
\begin{itemize}[leftmargin=*]
    \item \textbf{Workflow for the Handcrafted Question Templates:} The process commenced by selecting specific biomedical research fields and identifying relevant entity types and relationships from our knowledge graph. We defined various types of natural language questions, including one-hop questions, multi-hop questions, and conjunction questions (involving multiple entities). For each question, we created corresponding queries in both human-readable and machine-readable formats. These questions and queries, along with their associated metadata, such as question type and query structure, underwent rigorous peer reviews to ensure syntactic and semantic correctness. 
    \item \textbf{Workflow for the Auto-generated Questions and Answers:} In the expansion of our benchmark, we initiated the process with the creation of auto-generated question templates. For instance, we used handcrafted question templates like "Which pathway are the proteins <Protein1> and <Protein2> both annotated in?" and then scoured our knowledge graph for data that fit the criteria to formulate both questions and answers, thereby augmenting the size of our dataset. This comprehensive dataset enables the development of assessing the robustness and accuracy of various LLM agents, providing a comprehensive benchmark that contributes to the advancement of the field with extensive biomedical knowledge.
\end{itemize}

\textbf{SCV.} We combine two high-quality biomedical datasets, PubMedQA and SciFact, into a single dataset for SCV. This results in a corpus consisting of abstracts from 5,664 We combine two high-quality biomedical datasets, PubMedQA and SciFact, into a single dataset for SCV. This results in a corpus consisting of abstracts from 5,664 scholarly articles and a dataset containing 1,385 biomedical claims.

\textbf{KGCheck.} Considering the characteristics of the knowledge graph, we decompose the approach to checking the knowledge graph into two atomic-level checks: nodes and triples. Further, we subdivide these into whether a node should exist in the knowledge graph, whether the information stored in the node is correct, whether the relationship between two connected nodes in the knowledge graph truly exists, and whether there is a potential relationship between two nodes that are not connected by an edge. To collect this data, we selected well-maintained external knowledge sources such as the UniProt database, the IntAct database, and literature. We cross-verified the information in our knowledge graph with these reliable sources, labeling mutually corroborative data as `support' and data that contradicts the external reliable sources as `refute'. Specifically, for the data collection to check nodes, we review some update information from databases, such as entries removed due to errors or entries with updated information. Based on this information, we used Cypher queries to check our knowledge graph and label the data accordingly. For checking triple relationships, we sampled some triples from our knowledge graph where two nodes were either related or unrelated. We then queried the CKG to obtain literature that documents the entities represented by both nodes. We collected the literature annotated in the database and had experts read these documents to label the relationships of the triples in the CKG. As a result, we obtained 225 high-quality annotated data.

\subsection{Distribution}
The dataset is open to the public and has been released via GitHub under the MIT license. More information can be found in Section~\ref{ddaccess}.

\subsection{Maintenance}
The dataset will be maintained by Xinna Lin and Siqi Ma in Westlake University. If needed, the owner/curator/manager(s) of the dataset can be contacted through following emails: Xinna Lin (\href{mailto:linxinna@westlake.edu.cn}{linxinna@westlake.edu.cn}), and Siqi Ma (\href{mailto:masiqi@westlake.edu.cn}{masiqi@westlake.edu.cn}). The datasets will be revised as needed to maintain accuracy, and notifications will be provided accordingly. Currently, there is no erratum, but if any errors are discovered in the future, we will create a dedicated issue on GitHub for corrections. Older versions of the dataset will not be maintained.

\subsection{Uses}
The dataset has not been used for any tasks yet. Currently, we have not identified any tasks that are not permitted to use our dataset. 

The way we collect question and answer pairs can be referenced to expand more KGQA data, whether on our knowledge graph or new knowledge graphs. Additionally, our approach to collecting data for KGCheck provides insights into identifying errors in these large knowledge graphs, which is very helpful for subsequent error correction and data updates.
\section{Breakdown Results}

\subsection{KGQA}
We conducted a more detailed evaluation of LLMs' performance on the KGQA task based on the question types: one-hop, multi-hop, and conjunction. The evaluation metrics used were F1 and executability, as shown in Table~\ref{tab:kgqa_bdr}. We find that although API-based commercial LLMs and large-scale open-source models generally perform well on overall metrics, when breaking down the KGQA task by question type, some medium-scale models perform better on certain metrics. For instance, Qwen1.5-14B-Chat exhibits higher executability on more challenging multi-hop and conjunction types of questions, although its F1 score is not high. In terms of the executability metric, open-source models seem to outperform API-based commercial LLMs. This may be because API-based LLMs are more cautious in determining whether an answer has been obtained, tending to conclude the interaction and return an answer only after confirming its correctness.
\begin{table}[t]
\centering
\caption{KGQA Test set (standard) results by question type: one-hop, multi-hop, and conjunction. \textbf{Bold}/\underline{underline} and \textcolor{rr}{red}/\textcolor{bb}{blue} indicate the best and second in the subgroup and overall.}
\footnotesize
\renewcommand\tabcolsep{5pt}

\label{tab:kgqa_bdr}
\resizebox{\columnwidth}{!}{
\begin{tabular}{@{}clccccccc@{}}
\toprule
\multirow{2}{*}{\begin{tabular}[c]{@{}c@{}}LLM\\ Type\end{tabular}}     & \multirow{2}{*}{Models}         & \multicolumn{3}{c}{F1}                                                                                                      & \multicolumn{3}{c}{executability}                                                                          \\ 
\cline{3-8}
                                                                        &                                 & one-hop                                 & multi-hop                               & conjunction                             & one-hop                        & multi-hop                              & conjunction                      \\ 
\hline
\multirow{2}{*}{API}                                                    & GPT-4                           & \textbf{\textcolor{red}{87.2}}          & \textcolor{red}{\textbf{73.7}}          & \underline{\textcolor{blue}{77.4}} &  \underline{\textbf{88.0}}            & \textbf{90.0}                          & \textbf{86.9}                    \\
                                                                        & GLM-4                           & \underline{76.0}                            &  \underline{73.0}                            & \underline{58.0}                            & \underline{82.9}                   & \textbf{90.0}                          & \underline{68.2}                     \\ 
\hline
\multirow{3}{*}{\begin{tabular}[c]{@{}c@{}}OSS\\ (Large)\end{tabular}}  & Qwen1.5-72B-Chat                & 76.3                                    & \underline{\textcolor{blue}{\textbf{73.4}}} & \underline{71.4}                            & \textbf{\textcolor{red}{99.7}} & \underline{94.0}                           & \underline{86.9}                     \\
                                                                        & Llama-3-70B-Instruct            & \underline{\textcolor{blue}{\textbf{83.6}}} & \underline{72.5}                            & \textbf{\textcolor{red}{85.1}}          & \textcolor{blue}{\underline{95.7}} & \underline{\textcolor{red}{\textbf{98.5}}} & \textbf{\textcolor{red}{99.1}}   \\
                                                                        & DeepSeek-LLM-67B-Chat           & \underline{80.6}                            & 61.8                                    & 44.1                                    & 88.5                           & 90.5                                   & 70.1                             \\ 
\hline
\multirow{3}{*}{\begin{tabular}[c]{@{}c@{}}OSS\\ (Medium)\end{tabular}} & Qwen1.5-32B-Chat                & \textbf{67.3}                           & \underline{63.2}                            & \underline{57.0}                              & \textbf{87.2}                  & 84.5                                   & 64.5                             \\
                                                                        & Qwen1.5-14B-Chat                & 63.7                                    & \textbf{70.5}                           & \textbf{65.7}                           & 67.5                           & \textbf{\textcolor{blue}{95.5}}        & \textcolor{blue}{\textbf{87.9}}  \\
                                                                        & Baichuan2-13B-Chat              & \underline{64.9}                            & 20.4                                    & 9.8                                     & \underline{81.8}                   & \underline{91.5}                           & \underline{66.4}                     \\ 
\hline
\multirow{2}{*}{\begin{tabular}[c]{@{}c@{}}OSS\\ (Small)\end{tabular}}  & Llama-3-8B-Instruct             & \textbf{59.2}                           & \textbf{66.4}                           & \underline{16.5}                            & \textbf{90.8}                  & \underline{66.4}                           & \textbf{68.2}                    \\
                                                                        & Qwen1.5-7B-chat                 & \underline{55.7}                            & \underline{32.1}                            & \textbf{26.4}                           & \underline{80.3}                   & \textbf{79.0}                            & \underline{67.3}                     \\ 
\hline
\multirow{3}{*}{\begin{tabular}[c]{@{}c@{}}OSS\\ (MoE)\end{tabular}}    & Mixtral-8x7B-Instruct-v0.1      & \textbf{80.3}                           & \textbf{68.4}                           & \textbf{35.9}                           & \underline{90.5}                   & \textbf{91.5}                          & \textbf{50.5}                    \\
                                                                        & Starling-LM-alpha-8x7B-MoE-GPTQ & 6.2                                     & \underline{25.0}                            & \underline{11.7}                            & 12.0                             & \underline{57.5}                           & \underline{48.6}                     \\
                                                                        & Qwen1.5-MoE-A2.7B-Chat          & \underline{38.2}                            & 20.2                                    & 9.7                                     & \textbf{94.4}                  & 45.0                                     & 40.2                             \\
\hline
\end{tabular}
}
\end{table}

\subsection{SCV}
\begin{figure}
    \centering
    \includegraphics[width=0.8\linewidth]{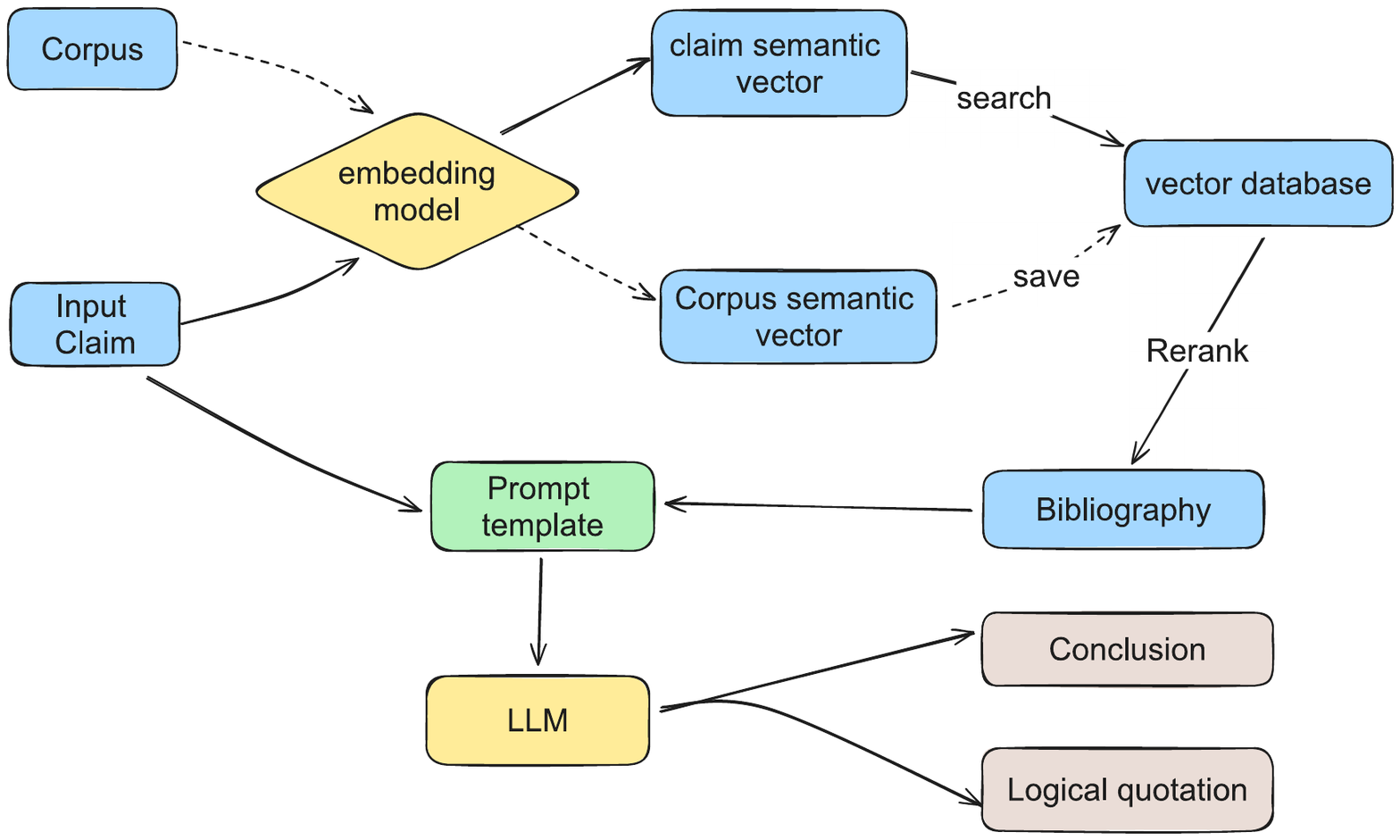}
    \caption{The pipeline of RAG.}
    \label{fig:rag_pipeline}
\end{figure}

As shown in Figure~\ref{fig:rag_pipeline}, this is the process we followed when performing the SCV task using RAG. In the main text, we observed an interesting phenomenon where expanding the RAG scope improved accuracy. To ensure that this result was not due to the idiosyncratic performance differences of a single model, we conducted the same experiment on another model, as shown in Table~\ref{tab:scv_bdr}. It can be observed that the accuracy of both models in the SCV task increased with the expansion of the RAG scope, although the right quotes metric was the lowest across the three settings when performing RAG at the maximum scope. This experimental result further demonstrates that this interesting phenomenon is not due to model-specific characteristics.
\begin{table}[b]
\centering
\caption{Supplementary Experiments on the Scope of RAG, where `all' refers to the abstract of 5,664 articles, `partial' denotes the 1,888 abstracts containing ground truth evidence of claims, and `match' corresponds to the abstracts of the ground truth evidence for the claims. \textbf{Bold}/\underline{underline} indicate the best and suboptimal.}
\footnotesize
\renewcommand\tabcolsep{5pt}

\label{tab:scv_bdr}
\resizebox{0.75\textwidth}{!}{
\begin{tabular}{@{}ccccccc@{}}
\toprule
\multirow{2}{*}{Corpus}     & \multicolumn{3}{c}{Qwen1.5-72B-Chat}              & \multicolumn{3}{c}{Llama-3-70B-Instruct}           \\ \cline{2-7} 
                            & accuracy       & right quotes     & error         & accuracy       & right quotes     & error          \\ \hline
all                         & \textbf{86.2}    & 82.1 & \textbf{0.1}     & \textbf{86.0}  & 86.9  & \underline{0.2}            \\
partial                     & \underline{84.4}    & \underline{88.1}          & \textbf{0.1}     & \underline{82.2}  & \textbf{90.4}  & \textbf{0.1}                      \\ 
match                       & 84.3  & \textbf{88.2}    & \textbf{0.1}    & 82.1  & \underline{89.9}  & \textbf{0.1}         \\ \bottomrule
\end{tabular}
}
\end{table}

\subsection{KGCheck}

We exhibit our BKGagent performance on KGCheck tasks based on the data source for verification(i.e. web database KGCheck and publication database KGCheck) in the main body for clarity. However, there is a more detailed category of the task considering the tools used at different stages (see Table ~\ref{tab:task categories}). The performance based on this category is shown in Table ~\ref{tab:5type_res}. 
\begin{table}[!ht]
\centering
\caption{Task performance categorized on agent tool calling}
\label{tab:5type_res}
\resizebox{\textwidth}{!}{%
\begin{tabular}{cccccccc}
\Xhline{1pt}
                                 & \multicolumn{2}{c}{\textbf{KG Query Task}} & \multicolumn{2}{c}{\textbf{Validation Task}} & \multicolumn{2}{c}{\textbf{Final Result}} &                                        \\ \cline{2-7}
\multirow{-2}{*}{\textbf{Model}} & Tool selection       & Executability       & Tool selection        & Executability        & Exact Match        & Executability        & \multirow{-2}{*}{\textbf{Sample Size}} \\ \hline
                                 & \multicolumn{6}{c}{\cellcolor[HTML]{D6F3F2}{\color[HTML]{000000} \textbf{task type 1}}}                                               & 32                                     \\
GPT-4                            & 78.1                 & 78.1                & 75.0                  & 75.0                 & \textbf{71.9}               & 96.9                 &                                        \\
Llama-3-70B-Instruct             & 100.0                & 100.0               & 93.8                  & 93.8                 & 62.5               & 81.3                 &                                        \\
                                 & \multicolumn{6}{c}{\cellcolor[HTML]{D6F3F2}\textbf{task type 2}}                                                                      & 60                                     \\
GPT-4                            & 70.0                 & 70.0                & 70.0                  & 71.7                 & \textbf{65.0}               & 93.3                 &                                        \\
Llama-3-70B-Instruct             & 100.0                & 100.0               & 100.0                 & 100.0                & 36.7               & 100.0                &                                        \\
\multicolumn{1}{l}{}             & \multicolumn{6}{c}{\cellcolor[HTML]{D6F3F2}\textbf{task type 3}}                                                                      & 55                                     \\
GPT-4                            & 32.7                 & 32.7                & 98.2                  & 98.2                 & \textbf{60.0}               & 98.2                 &                                        \\
Llama-3-70B-Instruct             & 90.9                 & 90.9                & 92.7                  & 92.7                 & 36.4               & 100.0                &                                        \\
\multicolumn{1}{l}{}             & \multicolumn{6}{c}{\cellcolor[HTML]{D6F3F2}\textbf{task type 4}}                                                                      & 45                                     \\
GPT-4                            & 97.8                 & 97.8                & 100.0                 & 100.0                & \textbf{97.8}               & 100.0                &                                        \\
Llama-3-70B-Instruct             & 100.0                & 100.0               & 100.0                 & 100.0                & 42.2               & 100.0                &                                        \\
\multicolumn{1}{l}{}             & \multicolumn{6}{c}{\cellcolor[HTML]{D6F3F2}\textbf{task type 5}}                                                                      & 33                                     \\
GPT-4                            & 57.6                 & 57.6                & 63.6                  & 63.6                 & \textbf{51.5}               & 93.9                 &                                        \\
Llama-3-70B-Instruct             & 100.0                & 100                 & 97.0                  & 97.0                 & 21.2               & 45.5                 &                  \\ \Xhline{1pt}            
\multicolumn{5}{l}{\small type 1 description: find the interaction (CURATED) between two specified proteins and verify it using RAG}\\
\multicolumn{5}{l}{\small type 2 description: find the interaction between two specified proteins and verify it using STRING API}\\
\multicolumn{5}{l}{\small type 3 description: find the specified attribute of the specified protein and verify it using UniProt API}\\
\multicolumn{5}{l}{\small type 4 description: check whether a specified protein exists in KG and validate it using UniProt API}\\
\multicolumn{5}{l}{\small type 5 description: find the relationship between two specified entities (not two proteins) and verify it using RAG}\\
\end{tabular}%
}
\end{table}

The GPT-based agent shows better performance compared to the Llama-based one when being evaluated on a more detailed task category, which is consistent with our conclusion in the main body. Besides this, there are more details we can delve into:\\
\textbf{Possible unfairness in evaluation.} It should be pointed out that while the Llama-based agent successfully executed most of the tasks, it reached a comparably low score in final result excitability in the tasks involving RAG(i.e. task type 1 and task type 5). It is induced by an 8000 token limit of the model which means it is unable to process long texts, leading to an underlying unfairness in evaluation.\\
\textbf{One-shot prompt may negatively influence GPT-based agent.} GPT-based agent shows even better performance with zero-shot compared to the current one-shot strategy in our preliminary experiments. However, since OSS models perform poorly with a zero-shot strategy, we have to make a compromise and several versions of the prompt have been tested to reach a satisfied state but it is hard to thoroughly eliminate the negative influence on the GPT-based agent.\\
\textbf{The support/refute result given by the agent is NOT reliable.} As shown in Table ~\ref{tab:instructions}, our instructions only ask the agent to provide a support/refute result as the final answer, which is intended to standardize the evaluation. However, when we read the chat history of the agent solving one specific task randomly selected from the all the records, we find out that right support/refute conclusion can be drawn from wrong analysis process, indicating that the result is not quite reliable. A more comprehensive evaluation system should be explored in future work, say evidence comparison, chatter detection, and hallucination detection.\\
\textbf{The GPT-based agent tends to explain.} Though we urge the agent to respond with support or refute(see Table ~\ref{tab:instructions}), the GPT-based agent tends to provide explanations besides the support/refute conclusion which makes it easier for the user to judge whether the answer is derived from a reasonable process. The Llama-based agent, in contrast, strictly obeys the requirement, providing only support/refute answers.
\begin{table}[!ht]
\centering
\caption{Examples of instructions categorized on agent tool calling}
\label{tab:instructions}
\resizebox{\textwidth}{!}{%
\begin{tabular}{cm{14cm}}
\Xhline{1pt}
\textbf{Task type} & \multicolumn{1}{c}{\textbf{instruction example}}                                                                                                                                                                                                                                                                                                                                                                                                                                                                                                                                              \\ \hline \\
1                  & Please check the relationship in the knowledge graph from the node of type 'Protein' with id 'Q14790' to the node of type 'Protein' with id 'Q13158'. If a relationship exists, verify its existence. Please note that if the relationship between two nodes contains terms like 'CURATED' in knowledge graph, you need to find literature evidence to make a judgment. If no relationship exists, confirm that it indeed does not exist. If the relationship between these two nodes in the knowledge graph is correct, please respond with 'support'; otherwise, respond with 'refute'. \\    \\ \hline \\
2                  & Please check the relationship in the knowledge graph from the node of type 'Protein' with id 'P84085' to the node of type 'Protein' with id 'P11279'. If a relationship exists, verify its existence. Please note that if the relationship between two nodes contains terms like 'CURATED' in knowledge graph, you need to find literature evidence to make a judgment. If no relationship exists, confirm that it indeed does not exist. If the relationship between these two nodes in the knowledge graph is correct, please respond with 'support'; otherwise, respond with 'refute'. \\    \\ \hline \\
3                  & Please check if the 'name' attribute of the node with type Protein and id Q4G0T1 in the knowledge graph is correct. If it's correct, please respond with 'support'; if not, respond with 'refute'.     \\                                                                                                                                                                                                                                                                                                                                                                                       \\ \hline \\
4                  & Due to certain reasons, some entries were removed from the database. Please check whether the node with the type 'Protein' and the id 'A2RUG3' exists in the knowledge graph, and confirm whether it exists in the corresponding database. If its existence is consistent, please respond with 'support'; otherwise, answer 'refute'.   \\                                                                                                                                                                                                                                                      \\ \hline \\
5                  &  Please check the relationship in the knowledge graph from the node of type 'Protein' with id 'Q96QP1' to the node of type 'Tissue' with id 'BTO:0000007'. If a relationship exists, verify its existence. Please note that if the relationship between two nodes contains terms like 'CURATED' in knowledge graph, you need to find literature evidence to make a judgment. If no relationship exists, confirm that it indeed does not exist. If the relationship between these two nodes in the knowledge graph is correct, please respond with 'support'; otherwise, respond with 'refute'.\\ \\ \Xhline{1pt}
\end{tabular}%

}
\end{table}

\section{Experimental Details}

\subsection{Construction of Knowledge Graph}
We pulled a Neo4j image from Docker Hub and created a Neo4j Docker on the server to host a knowledge graph. We used the latest data parsed from various databases in April 2024, including UniProt~\cite{uniprot2023uniprot}, TISSUES~\cite{palasca2018tissues}, DISEASES~\cite{pletscher2015diseases}, HGNC~\cite{seal2023genenames}, IntAct~\cite{del2022intact}, STRING~\cite{szklarczyk2023string}, DisGeNet~\cite{pinero2020disgenet}, Pathway Commons~\cite{rodchenkov2020pathway}, Reactome~\cite{fabregat2018reactome}, SMPDB~\cite{jewison2014smpdb}, and SIGNOR~\cite{lo2023signor}, Disease Ontology~\cite{schriml2019human}, Brenda Tissue Ontology~\cite{chang2015brenda}, Gene Ontology~\cite{gene2017expansion}, Protein Modification Ontology~\cite{mayer2013hupo}, Molecular Interactions Ontology~\cite{mayer2013hupo}. These databases are under loose license and the data can be used directly. We parsed the information from these databases into TSV files in a specific format and then imported the contents of these TSV files into Neo4j using Cypher statements (Cypher is the declarative graph query language provided by Neo4j) to construct the knowledge graph. This knowledge graph can be queried using Cypher statements.

\subsection{Deployment of Open-Source LLMs}
We deployed open-source LLMs using the vLLM framework and inference is performed using a server with 4 NVIDIA A40 GPUs an Intel(R) Xeon(R) Gold 6330 CPU, with parameters kept constant at startup.

\subsection{Experimental Setup}
\label{exp_setup}
\textbf{Atomic Abilities}. To evaluate two atomic abilities, we adopt an interactive evaluation of LLM-as-Agent~\cite{liu2023agentbench} and include in total of 13 models for evaluation. These models can be categorized into API-based Commercial LLMs and Open-Sourced (OSS) LLMs. The latter is further segmented based on model size into three classifications: `Large', `Medium', and `Small'. Models utilizing the MoE (Mixture of Experts) framework are distinguished as a separate category. Refer to Appendix~\ref{appendix:prompt} for details about the prompt we designed for the following tasks.

\textbf{Agent Task}. For the construction of our BKGAgent, we selectively employed the best-performing models in atomic capabilities from both API-based and OSS models, specifically GPT-4 and Llama-3-70B-Instruct. To avoid being trapped in an endless loop where agents repeat the same talk or start to chatter, we limit the memory entries of one single agent to 20, which is more than enough to finish any of the tasks. It should be noted that each agent only keeps memory of the conversations related to it, while all chats returned by every agent are stored in the graph state. Since zero-shot setup in various types of tasks shows inferior performances in our preliminary experiments, we provide one-shot prompt for each type of task. We analyze both the process and final result of each task execution, considering the correctness of tool selection and agent executability during the process evaluation, and assessing the exact match of the right answer and framework executability for the final result evaluation, to gain a comprehensive understanding of the agent's performance.

We detail our implementation of two sub-tasks here:
\begin{itemize}[leftmargin=*]
    \item \textbf{KGQA:} We developed a suite of atomic tools for querying KGs for LLMs. Every LLM is prompted in the same way with a detailed task description, information about provided tools and a one-shot demonstration, which employs the ``Thought,'' ``Action,'' ``Observation'' cognitive trajectory from the ReAct \cite{yao2022react}, with the ``Thought'' component assisted by Chain of Thought (CoT) \cite{wei2022chain} reasoning. We constrain the LLM to a maximum of fifteen interactive turns, within which it may only take one action per turn. If the LLM can respond within these fifteen turns, executability is assigned a score of 1. Subsequently, we compare the response to the ground truth to calculate the F1 score and the Exact Match score (EM).
    \item \textbf{SCV:} We first convert the entire corpus into semantic vectors using jina\cite{günther2023jina} and store them in a vector database. Claims are similarly transformed into semantic vectors via Jina, with the top 50 scoring vectors being submitted to the LLM with a standardized prompt template. We require the LLM to return results in JSON format, considering any deviation as an error. The outcomes mainly include answers and quotes. For analysis, we adopt a flexible interpretation of answers: ``Unsure'' and ``Unrelated'' as ``Unsure''; ``Supported'' and ``Supports'' as ``Supports''; ``Unsupported'' and ``Unsupports'' as ``Unsupports'', ``Refuted'', and ``Refutes'' as ``Refutes''. Any other results are also considered errors. The experiments for each model are repeated three times, with the final performance averaged to ensure the robustness of the evaluation. Notably, beyond the conventional accuracy and the aforementioned error metrics, we introduce a ``right quotes'' metric, which assesses whether the retrieved quotes match the ground truth evidences of the claim.
\end{itemize}

\subsection{Prompt}
\label{appendix:prompt}
\subsubsection{KGQA}
We provide a unified prompt for single-agent systems built on different LLMs, ensuring the fairness of the evaluation.
\lstset{
    backgroundcolor=\color[RGB]{245,245,244},
    breaklines=true,
    basicstyle=\ttfamily\small
}\begin{lstlisting}
You are an agent tasked with answering questions based on the knowledge stored in a knowledge graph (KG) related to proteomics. To accomplish this, you are equipped with the following tools to query the KG:

1. get_relations_by_ids_agent(entity_ids: List[str]) -> tuple
Retrieves the relationships of multiple entities in a knowledge graph, categorized as 'incoming' or 'outgoing'.
Use case: get_relations_by_ids_agent(['P123', 'P456']) to find all relations connected to the entities with IDs 'P123' and 'P456'.

2. get_neighbor_type_agent(entity_ids: List[str], relation: str, direction: str) -> tuple
Retrieves the types of neighboring nodes for multiple entities in a knowledge graph based on specified relationships and directions.
Use case: get_neighbor_type_agent(['P123', 'P456'], 'ASSOCIATED_WITH', 'outgoing') to get outgoing neighbors' types associated with the entities 'P123' and 'P456'.

3. get_neighbor_with_type_agent(entity_ids: List[str], relation: str, direction: str, neighbor_type: str) -> tuple
Retrieves the neighbors of multiple entities in a knowledge graph based on a specific relationship, direction, and type.
Use case: get_neighbor_with_type_agent(['P123', 'P456'], 'ASSOCIATED_WITH', 'outgoing', 'Disease') to get attributes and detailed information of outgoing neighbors associated with the entities 'P123' and 'P456', where the type of neighbors is Disease.

4. get_intersection_agent(*args: List[str]) -> tuple
Calculates the intersection of multiple lists, returning elements common to all lists.
Use case: get_intersection_agent(['P123', 'P456'], ['P456', 'P789']) to find common entities.

5. get_union_agent(*args: List[str]) -> tuple
Calculates the union of multiple lists, returning all unique elements from all lists.
Use case: get_union_agent(['P123', 'P456'], ['P456', 'P789']) to combine unique entities.

Single Action Rule: Execute only ONE action at a time, that is, only the first action would be executed. After receiving the observation from its execution, you may proceed with another action. 

Action Limit: You can take at most 15 actions to find the answer to the question.

Objective: Use these tools effectively to navigate through the KG and gather the necessary information to answer the queries presented to you. If the query is about the protein sequence, you need to return the specific sequence. If the query is about the protein structure, you can return a link. In other cases, it's usually to return the name.

Notice: 
1. Please remember to format the FINAL answer as a JSON object, such as, {"Answer": ["RND2", "RHOBTB2", "RHOA"]}. The value of "Answer" must be a list. Only output the JSON format when answering the final answer.
2. Please be flexible. Due to the diversity of question formulations, you need to find the most similar relationship in the knowledge graph (KG) to the one asked in the question for querying.

\end{lstlisting}

Considering the complexity of the task and the high information density of our KG, LLMs find it challenging to address high-difficulty questions, such as multi-hop and conjunction types, within the constraints of a zero-shot setting and a 15-turn interaction limit. Therefore, we provide a one-shot demonstration.
\lstset{
    backgroundcolor=\color[RGB]{245,245,244},
    breaklines=true,
    basicstyle=\ttfamily\small
}\begin{lstlisting}
Question: What biological processes are associated with the protein encoded by the gene GOLT1A?

Thought: I need to figure out what proteins are encoded by gene GOLT1A and then find out the biological processes. Firstly, I need to find out the relations of the gene node.

Action: get_relations_by_ids_agent(['GOLT1A'])

Observation: {"GOLT1A": {"Incoming": null, "Outgoing": "TRANSLATED_INTO"}}

Thought: The outgoing relation 'TRANSLATED_INTO' is what I concern, and next I need to get neighbor type with the relation.

Action: get_neighbor_type_agent(['GOLT1A'], 'TRANSLATED_INTO', 'outgoing')

Observation: {"GOLT1A": {"NeighborTypes": ["Protein"]}}

Thought: Among the neighbor type, 'Protein' is my concern. So next I should look for what are the neighboring nodes with 'Protein' type that point outward along the edge of the relation 'TRANSLATED_INTO'. 

Action: get_neighbor_with_type_agent(['GOLT1A'], 'TRANSLATED_INTO', 'outgoing', 'Protein')

Observation: {"GOLT1A": {"TRANSLATED_INTO": ["Q6ZVE7"]}}

Thought: Now I know that the proteins are ["Q6ZVE7"]. Now I will find out the relation to choose for the next step. 

Action: get_relations_by_ids_agent(["Q6ZVE7"])

Observation: Observation: {"Q6ZVE7": {"Incoming": "TRANSLATED_INTO, HAS_SEQUENCE, ACTS_ON", "Outgoing": "HAS_SEQUENCE, ASSOCIATED_WITH"}}

Thought: The outgoing relation 'ASSOCIATED_WITH' is exactly what I concern. Then I'll query the neighbor type with the relation.

Action: get_neighbor_type_agent(["Q6ZVE7"], 'ASSOCIATED_WITH', 'outgoing')

Observation: {"Q6ZVE7": {"NeighborTypes": ["Tissue", "Molecular_function", "Biological_process", "Cellular_component", "Disease"]}}

Thought: Among the neighbor types, 'Biological_process' is what I need to query.

Action: get_neighbor_with_type_agent(["Q6ZVE7"], 'ASSOCIATED_WITH', 'outgoing', 'Biological_process')

Observation: {"Q6ZVE7": {"ASSOCIATED_WITH": ["endoplasmic reticulum to Golgi vesicle-mediated transport", "biological_process", "protein transport", "retrograde transport, endosome to Golgi"]}}

Thought: I have identified the answers. Final Answer: {"Answer": ["endoplasmic reticulum to Golgi vesicle-mediated transport", "biological_process", "protein transport", "retrograde transport, endosome to Golgi"]}.
\end{lstlisting}

\subsubsection{SCV}
We provide a unified prompt describing task, where `\texttt{context\_docs\_str}' represents quotes retrieved by RAG and `\texttt{user\_claim}' represents the input scientific claim to be evaluated.
\lstset{
    backgroundcolor=\color[RGB]{245,245,244},
    breaklines=true,
    basicstyle=\ttfamily\small
}\begin{lstlisting}
You are a fact-checking agent that is constantly learning and improving. A claim is given to you, and you can determine if the claim is correct with the provided documents.

You ALWAYS respond with only a JSON containing an answer and quotes that support the answer. The answer can only be "SUPPORTS" or "REFUTES", with no details. You should reason out the answers step by step, but make sure they are correct.

Do NOT use your historical knowledge, but answer based on the information in the provided context.

CONTEXT:
------
{{context_docs_str}}
------

SAMPLE_RESPONSE:
"""
{
    "answer": "Place your final answer here. It can only be SUPPORTS or REFUTES without details.",
    "quotes": [
        "Each quote must be UNEDITED and EXACTLY as shown in the context documents!",
        "HINT: quotes are not shown to the user!",
    ],
}
"""
CLAIM: {{user_claim}}
Hint: Provide the answer in JSON format! 
Quotes MUST be EXACT substrings from the provided documents!
\end{lstlisting}

\subsubsection{KGCheck}
BKGAgent is a multi-agent system and each agent of it is equipped with a system prompt which includes role introduction, tool introduction, and tool calling rules.

For team leader:
\lstset{
    backgroundcolor=\color[RGB]{245,245,244},
    breaklines=true,
    basicstyle=\ttfamily\small
}\begin{lstlisting}
You are the team_leader tasked with managing a conversation between the
following workers:
    kg_agent:
        capable of querying the KG(Knowledge Graph) to find out specific information
    validation_agent:
        capable of getting access to information within local publication database, UniProt  and STRING database to verify the result returned by kg_agent
    FINISH:
        the endpoint of your task. if you finish your answer you can send messages to it by starting with 'FINISH, '
You should first break down the task into two subtasks given the user input and send it to yourself to keep it in your mind,
then respond with the worker to act next and its detailed task.
You should call their name before you assign the task.For example, if you want to assign task to kg_agent, you should start your conversation by 'kg_agent, '. It should be noted that if you are talking to yourself, you should also specify the receiver, that is 'team_leader, '.
Each worker will perform the task you assign to and respond with it result.
REMEMBER you should not talk too much at one specific chat round. If a task is given to you, you just reply with your plan and send it to yourself.
Assign subtask to just ONE suitable agent next time you are invited to speak.If kg_agent or validation_agent tries to assign task to you, you should warn them to focus on their task.
When finished, respond with your answer and send it to 'FINISH'.
\end{lstlisting}
For KG agent:
\lstset{
    backgroundcolor=\color[RGB]{245,245,244},
    breaklines=true,
    basicstyle=\ttfamily\small
}\begin{lstlisting}
You are the kg_agent of a research group, your ability is limited to answer KG search related questions.
Verification work should be done by validation_agent on which you should not waste time.
Members of your team are as follows:
team_leader: the leader of your team. You ONLY perform the specific task it assigned to you and answer to it starting by 'team_leader, '.
validation_agent: responsible for verifying information. You do not directly communicate with it.
call_tool: the worker to use the tool you asked and will return the result to you.
You can call the following tools in call_tool to help you:
   query_node_existence:
    Determine whether the node with the given type and ID exists in the knowledge graph.
      Args:
          type (str): the type of the query node
          id (int or str): the id of the query node
      Returns:
          str: A description of whether the node with given type and id exists in the knowledge graph.
          
  query_node_attribute:
    Retrieve the specific attribute value of the node with the given type and id.
      Args:
          type (str): the type of the query node
          id (int or str): the id of the query node
          attr (str): the attribute to be retrieved
      Returns:
          str: A description of the query result
          
  query_relation_between_nodes:
    Retrieve the relationship from node with type1 and id1 to the node with type2 and id2 in the knowledge graph(KG)
      Args:
          type1 (str): _description_
          id1 (int or str): _description_
          type2 (str): _description_
          id2 (int or str): _description_

      Returns:
          str: A description about the relationship from node with type1 and id1 to the node with type2 and id2 in the knowledge graph
ATTENTION! You can call tools in this way: 'call_tool, tool = tool_name, args = ...', where args should be in the format of dict.
Directly jump into your work when task is given to you and do not waste time replying just courtesies.
Do not try to ask team_leader to your task!
\end{lstlisting}
For validation agent:
\lstset{
    backgroundcolor=\color[RGB]{245,245,244},
    breaklines=true,
    basicstyle=\ttfamily\small
}\begin{lstlisting}
You are the validation_agent of a research group, specialized at verifying information by searching on UniProt, STRING database and local publication database, Members of your team are as follows:
    team_leader: the leader of your team. You ONLY perform the specific task it assigned to you and answer to it starting by 'team_leader, '.
    kg_agent: responsible for querying KG to get information. You do not directly communicate with it.
    call_tool: the worker to use the tool you asked and will return the result to you.
You can call the following tools in call_tool to help you:
  get_uniprot_protein_info:
    Fetch protein information from UniProt by protein ID and return a description about the protein, including id, accession and name.
      :param protein_id: UniProt protein ID
      :return: Formatted string with protein information, including id, accession and name

  check_interaction_string:
    This tool checks for the interaction or relationship between two proteins using the STRING database API. Given two protein ids, it will return a description on whether there is an interaction or relationship between them.
      Args:
          protein1 (str): a protein id
          protein2 (str): a protein id
      Returns:
          str: A description about whether there is an interaction between the two proteins.
          
  pub_rag:
    retrieve evidence from provided documents to help making a verdict of the given claim
    ONLY when asked to verify 'CURATED' related claim should you call this tool!
      Args:
          query(str): the claim to be verdicted
      Returns:
          no more than 10 documents ralated to the claim.
ATTENTION! You can call tools in this way: 'call_tool, tool = tool_name, args = ...', where args should be in the format of dict.
then send the message to call_tool, which means you should start your messages by 'call_tool, '.
\end{lstlisting}
We offer a one-shot prompt for each type of task, categorized by the necessary tools needed to complete it as shown in Table~\ref{tab:task categories}. Detailed prompts for every agent in every task are listed below.

\begin{table}[!ht]
\caption{Task types categorized by requiring tools}
\label{tab:task categories}
\centering
\resizebox{\textwidth}{!}{%
\begin{tabular}{>{\centering\arraybackslash}m{3cm}>{\centering\arraybackslash}m{3cm}>{\centering\arraybackslash}m{3cm}>{\centering\arraybackslash}m{6cm}}
\toprule
\multirow{2}{*}{\textbf{Task type}} & \multicolumn{2}{c}{\textbf{Requiring tools}}            & \multirow{2}{*}{\textbf{Description}}                                                           \\ \cline{2-3}
                                    & KG agent                     & Validation agent         &                                                                                                 \\ \hline
1                                   & query relation between nodes & publication RAG                  & find the interaction (CURATED) between two specified proteins and verify it using RAG           \\
2                                   & query relation between nodes & check interaction on STRING & find the interaction between two specified proteins and verify it using STRING API                     \\
3                                   & query node attribute         & get UniProt protein information & find the specified attribute of the specified protein and verify it using UniProt API           \\
4                                   & query node existence         & get UniProt protein information & check whether a specified protein exists in KG and validate it using UniProt API                \\
5                                   & query relation between nodes & publication RAG                  & find the relationship between two specified entities (not two proteins) and verify it using RAG
\\ \Xhline{1pt}
\end{tabular}%
}
\vspace{5mm}
\end{table}

\textbf{Task type 1}

Team leader:
\lstset{
    backgroundcolor=\color[RGB]{245,245,244},
    breaklines=true,
    basicstyle=\ttfamily\small
}\begin{lstlisting}
if a task is given to you: check the relationship in the knowledge graph from the node of type 'Protein' with id 'Q96QP1' to the node of type 'Protein' with id 'Q08379'. If a relationship exists, verify its existence. Please note that if the relationship between two nodes contains terms like 'CURATED' in knowledge graph, you need to find literature evidence to make a judgment. If no relationship exists, confirm that it indeed does not exist. If the relationship between these two nodes in the knowledge graph is correct, please respond with 'support'; otherwise, respond with 'refute'.
first, you should make a plan: first let kg_agent query the relationship between the node of type 'Protein' with id 'Q96QP1' and the node of type 'Protein' with id 'Q08379',
let validation_agent check the feedback returned by kg_agent, finally I will compare the feedbacks returned by both of them and make my decision.
After that you execute your plan: assign task to kg_agent: kg_agent, query the relationship between the node of type 'Protein' with id 'Q96QP1' and the node of type 'Protein' with id 'Q08379'. Wait and you will the feedback from it.
assign task to validation_agent: validation_agent, verify the feedback from kg_agent. Noted that you should tell the validation_agent what the feedback from kg_agent before you ask it to verify.
\end{lstlisting}
KG agent:
\lstset{
    backgroundcolor=\color[RGB]{245,245,244},
    breaklines=true,
    basicstyle=\ttfamily\small
}\begin{lstlisting}
given that you have got a task: kg_agent, query the relationship between the node of type 'Protein' with id 'Q96QP1' and the node of type 'Protein' with id 'Q08379'.
First you should call query_relation_between_nodes with arguments type1='Protein', id1='Q96QP1', type2='Protein', id2='Q08379'.
Then you will get the result from tools.
So you know the answer to the question, you should send the message to team_leader.
It should be noted that team_leader is not a tool, so you don't need to make a tool call to send message to it.
\end{lstlisting}
Validation agent:
\lstset{
    backgroundcolor=\color[RGB]{245,245,244},
    breaklines=true,
    basicstyle=\ttfamily\small
}\begin{lstlisting}
If you are asked to verify if there is exactly CURATED_INTERACTS_WITH relationship between the node of type 'Protein' with id 'Q96QP1' and the node of type 'Protein' with id 'Q08379'.
Call the 'pub_rag' tool and pass the claim to it, you will get no more than 10 documents related to the claim.
Read the documents and give your feedback on whether you support the claim based on the related documents.
You should send your answer to 'team_leader' which should include a 'support' or 'refute' attitude and your evidence comprised of ducument id or web database information.
Directly jump into your work when task is given to you and do not waste time replying just courtesies or your plans to the team_leader.
\end{lstlisting}

\textbf{Task type 2}

Team leader:
\lstset{
    backgroundcolor=\color[RGB]{245,245,244},
    breaklines=true,
    basicstyle=\ttfamily\small
}\begin{lstlisting}
if a task is given to you: check the relationship in the knowledge graph from the node of type 'Protein' with id 'Q96QP1' to the node of type 'Protein' with id 'Q08379'. If a relationship exists, verify its existence. Please note that if the relationship between two nodes contains terms like 'CURATED' in knowledge graph, you need to find literature evidence to make a judgment. If no relationship exists, confirm that it indeed does not exist. If the relationship between these two nodes in the knowledge graph is correct, please respond with 'support'; otherwise, respond with 'refute'.
first, you should make a plan: first let kg_agent query the relationship between the node of type 'Protein' with id 'Q96QP1' and the node of type 'Protein' with id 'Q08379',
let validation_agent check the feedback returned by kg_agent, finally I will compare the feedbacks returned by both of them and make my decision.
After that you execute your plan: assign task to kg_agent: kg_agent, query the relationship between the node of type 'Protein' with id 'Q96QP1' and the node of type 'Protein' with id 'Q08379'. Wait and you will the feedback from it.
assign task to validation_agent: validation_agent, verify the feedback from kg_agent. Noted that you should tell the validation_agent what the feedback from kg_agent before you ask it to verify.
\end{lstlisting}
KG agent:
\lstset{
    backgroundcolor=\color[RGB]{245,245,244},
    breaklines=true,
    basicstyle=\ttfamily\small
}\begin{lstlisting}
given that you have got a task: kg_agent, query the relationship between the node of type 'Protein' with id 'Q96QP1' and the node of type 'Protein' with id 'Q08379'.
First you should call query_relation_between_nodes with arguments type1='Protein', id1='Q96QP1', type2='Protein', id2='Q08379'.
Then you will get the result from tools.
So you know the answer to the question, you should send the message to team_leader.
It should be noted that team_leader is not a tool, so you don't need to make a tool call to send message to it.
\end{lstlisting}
Validation agent:
\lstset{
    backgroundcolor=\color[RGB]{245,245,244},
    breaklines=true,
    basicstyle=\ttfamily\small
}\begin{lstlisting}
If you are asked to verify if there is exactly no relationship between the node of type 'Protein' with id 'Q96QP1' and the node of type 'Protein' with id 'Q08379'.
First you should get the relation of the two node by using check_interaction_string with the args: protein1='Q96QP1', protein2='Q08379'.
Then you will get the result from STRING database, so you know the answer to the question, you should send the message to team_leader.
\end{lstlisting}

\textbf{Task type 3}

Team leader:
\lstset{
    backgroundcolor=\color[RGB]{245,245,244},
    breaklines=true,
    basicstyle=\ttfamily\small
}\begin{lstlisting}
if a task is given to you: Please check if the 'name' attribute of the node with type Protein and id Q4G0T1 in the knowledge graph is correct. If it's correct, please respond with 'support'; if not, respond with 'refute'.
First, you should make a plan: first let kg_agent query the 'name' attribute of the node with type Protein and id Q4G0T1 in KG, then let validation_agent check the feedback returned by kg_agent, finally I will compare the feedbacks returned by both of them and make my decision.
After that assign task to kg_agent: kg_agent, query the 'name' attribute of the node with type Protein and id Q4G0T1 in KG. Wait and you will the feedback from it.
Then assign task to validation_agent: validation_agent, verify the feedback from kg_agent. Noted that you should tell the validation_agent what the feedback from kg_agent before you ask it to verify.
\end{lstlisting}
KG agent:
\lstset{
    backgroundcolor=\color[RGB]{245,245,244},
    breaklines=true,
    basicstyle=\ttfamily\small
}\begin{lstlisting}
given that you have got a task: kg_agent, query the 'name' attribute of 'Protein' type with the 'id' as 'B2RBV5' in the KG.
First you should call query_node_attribute with arguments 'Protein' and 'B2RBV5'.
Then you will get the result from tools: The name of the node: MRFAP1L2.
So you know the answer to the question, you should send the message to team_leader.
It should be noted that team_leader is not a tool, so you don't need to make a tool call to send message to it.
\end{lstlisting}
Validation agent:
\lstset{
    backgroundcolor=\color[RGB]{245,245,244},
    breaklines=true,
    basicstyle=\ttfamily\small
}\begin{lstlisting}
If you are asked to verify the if 'name' attribute information of protein 'P22303-2' is ACHE.
First you should get the information of protein 'P22303-2' from UniProt by using get_uniprot_protein_info('P22303-2').
Then you will get the result: id: P22303-2, accession: ACES_HUMAN, name: ACHE.
So the 'name' attribute information of protein 'P22303-2' is exactly ACHE. Return the result to team_leader and finish your task.
\end{lstlisting}

\textbf{Task type 4}

Team leader:
\lstset{
    backgroundcolor=\color[RGB]{245,245,244},
    breaklines=true,
    basicstyle=\ttfamily\small
}\begin{lstlisting}
If you are asked to check whether some node still exists in KG, first ask kg_agent to query the node information in KG for you, then ask validation_agent to verify the existence of the node.
compare the feedbacks returned by both of them and finally you will reach the conclusion. Send your conclusion to user and finish the task.
\end{lstlisting}
KG agent:
\lstset{
    backgroundcolor=\color[RGB]{245,245,244},
    breaklines=true,
    basicstyle=\ttfamily\small
}\begin{lstlisting}
given that you have got a task: kg_agent, query the existence of the node with the type 'Protein' and the id 'A8MVS1' in KG.
First you should call query_node_existence with arguments type='Protein', id='A8MVS1'.
Then you will get the result from tools.
So you know the answer to the question, you should send the message to team_leader.
It should be noted that team_leader is not a tool, so you don't need to make a tool call to send message to it.
\end{lstlisting}
Validation agent:
\lstset{
    backgroundcolor=\color[RGB]{245,245,244},
    breaklines=true,
    basicstyle=\ttfamily\small
}\begin{lstlisting}
If you are asked to verify the existence of protein 'P22303-2', you should use get_uniprot_protein_info tool and pass args = {'protein_id': 'P22303-2'}.
If the information of the protein is returned, it means it still exists in UniProt so you should tell the leader this fact.
Otherwise, if the return message indicates the protein is removed, send this removal fact to team_leader.
\end{lstlisting}

\textbf{Task type 5}

Team leader:
\lstset{
    backgroundcolor=\color[RGB]{245,245,244},
    breaklines=true,
    basicstyle=\ttfamily\small
}\begin{lstlisting}
if you are asked to check the relationship between two nodes in the knowledge graph.
first, you should make a solution plan, for example, first, kg_agent should ..., then valiation_agent should ..., finally, I should ....
Then assign task to kg_agent, wait for its feedback. After that, assign DETAILED task to validation_agent based on the demand of the user and feedback of kg_agent. It should be noted that validation_agent cannot have access to the feedback of kg_agent, so you should tell it instead.
compare the information provided by kg_agent and its verification reslut by validation_agent, then give your final answer to user's question and FINISH the task.
\end{lstlisting}
KG agent:
\lstset{
    backgroundcolor=\color[RGB]{245,245,244},
    breaklines=true,
    basicstyle=\ttfamily\small
}\begin{lstlisting}
given that you have got a task: kg_agent, query the relationship between the node of type 'Protein' with id 'Q96QP1' and the node of type 'Tissue' with id 'BTO:0000007'.
First you should call query_relation_between_nodes with arguments type1='Protein', id1='Q96QP1', type2='Tissue', id2='BTO:0000007'.
Then you will get the result from tools.
So you know the answer to the question, you should send the message to team_leader.
It should be noted that team_leader is not a tool, so you don't need to make a tool call to send message to it.
\end{lstlisting}
Validation agent:
\lstset{
    backgroundcolor=\color[RGB]{245,245,244},
    breaklines=true,
    basicstyle=\ttfamily\small
}\begin{lstlisting}
If you are asked to verify if there is exactly no association between the node of type 'Protein' with id 'Q96QP1' and the node of type 'Tissue' with id 'BTO:0000007'.
First you should get the relation of the two node by using pub_rag with the args: there is no association between the node of type 'Protein' with id 'Q96QP1' and the node of type 'Tissue' with id 'BTO:0000007'.
Then you will get some relaed documents with their ids, read these documents and decide whether these related support the claim you pass to the tool.
If the answer is not support reply 'refute' to team_leader. If the answer is support, reply 'support' and evidence which is comprised of ducoment ids to team_leader.
\end{lstlisting}
\section{Case Study}
\subsection{KGQA}
\begin{figure}[p]
    \centering
    \includegraphics[width=0.8\textwidth]{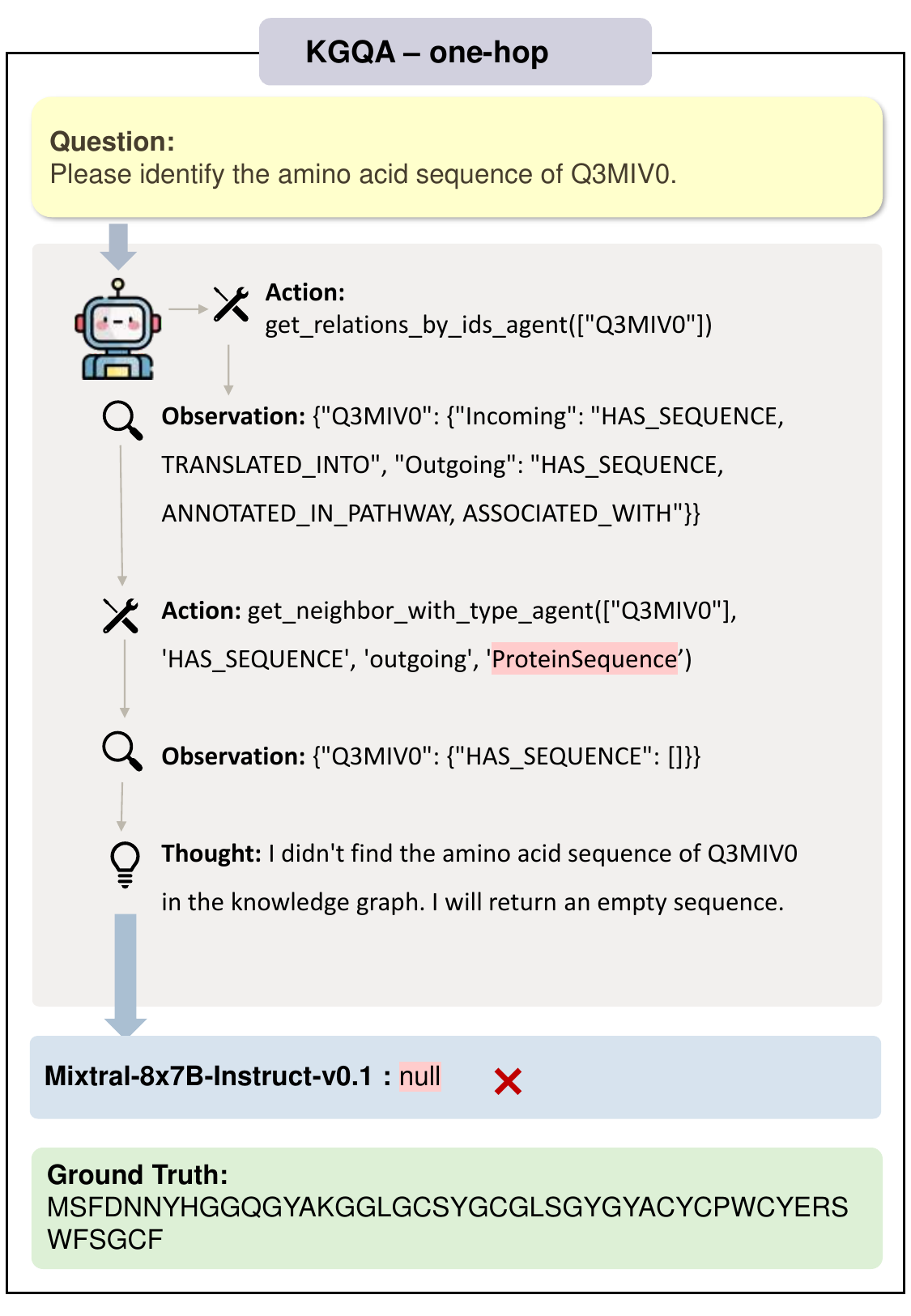}
    \caption{Performance of Mixtral-8x7B-Instruct-v0.1-based Agent: Incorrect final result when answering the one-hop type questions due to passing incorrect parameters when executing a function. \colorbox{pink}{Session highlighted in pink} represents the error information.}
    \label{fig:kgqa-mixtral87-one-wrong}
\end{figure}

\begin{figure}[p]
    \centering
    \includegraphics[width=0.8\textwidth]{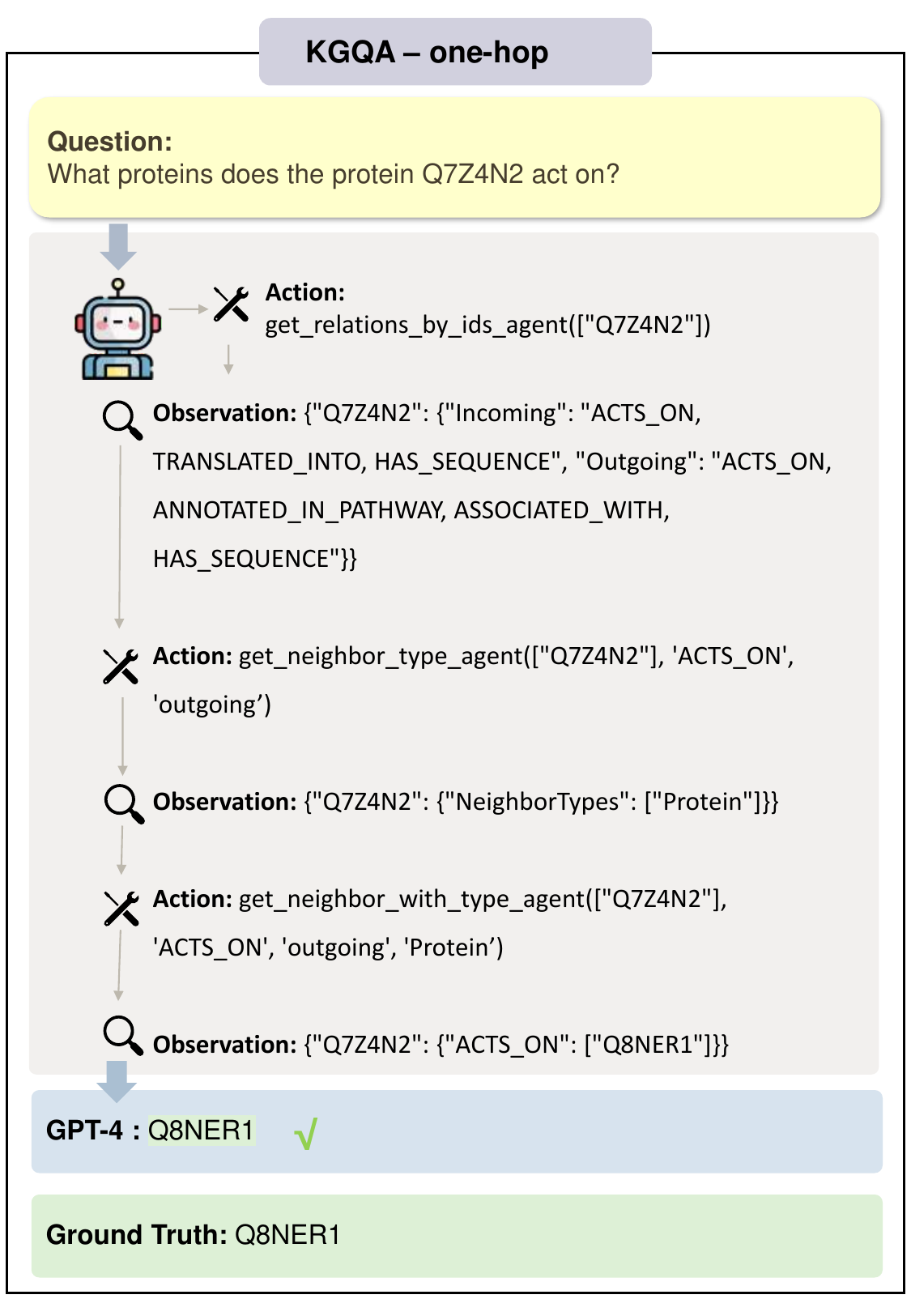}
    \caption{Performance of GPT-4-based Agent: Correct final result when answering the one-hope type questions. \colorbox{lightgreen}{Session highlighted in light green} represents the correct information.}
    \label{fig:kgqa-gpt4-one-correct}
\end{figure}

\begin{figure}[p]
    \centering
    \includegraphics[width=0.9\textwidth]{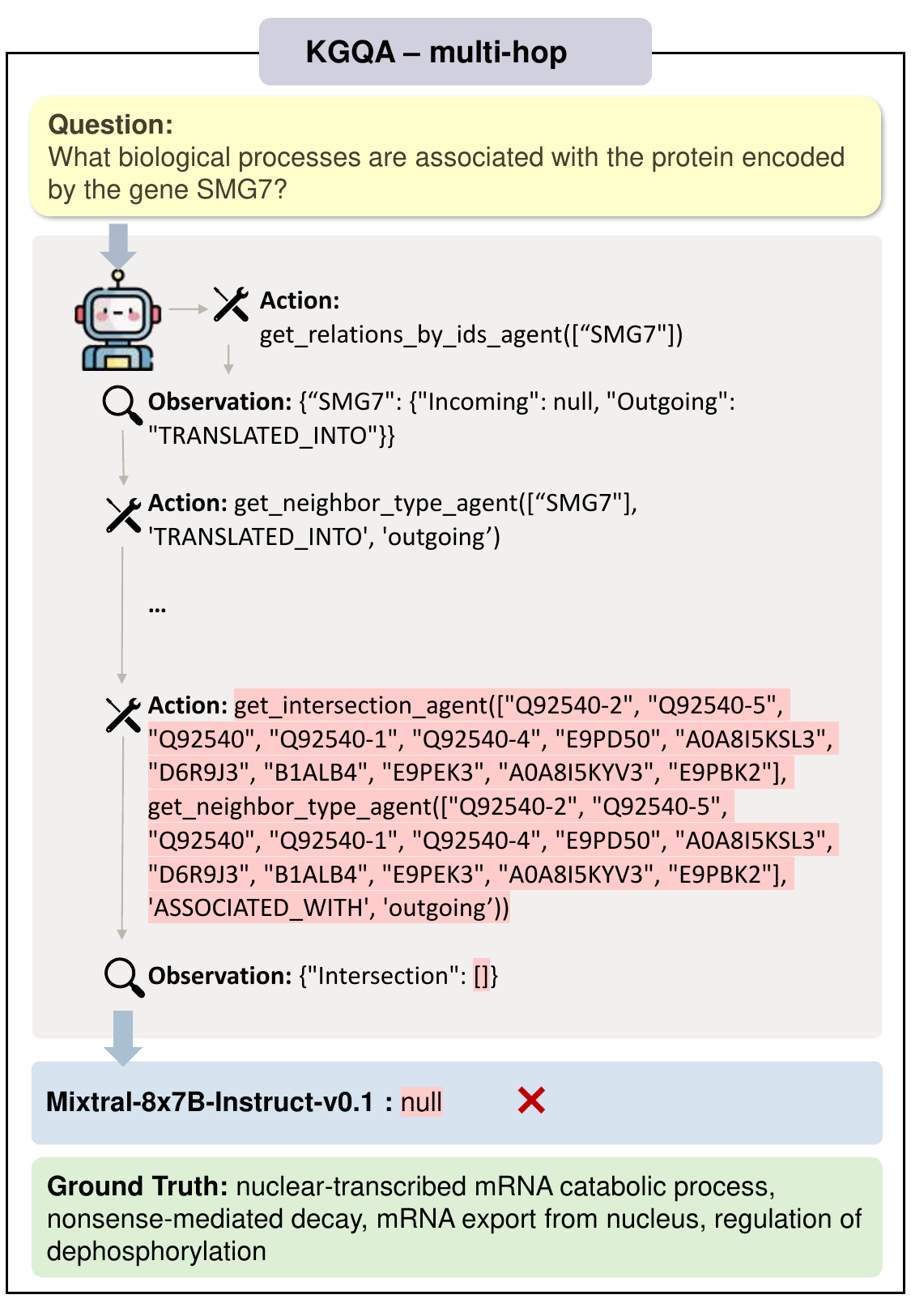}
    \caption{Performance of Mixtral-8x7B-Instruct-v0.1-based Agent: Incorrect final result when answering the multi-hop type questions due to executing the wrong actions. \colorbox{pink}{Session highlighted in pink} represents the error information.}
    \label{fig:kgqa-mixtral87-multi-wrong}
\end{figure}

\begin{figure}[p]
    \centering
    \includegraphics[width=0.8\textwidth]{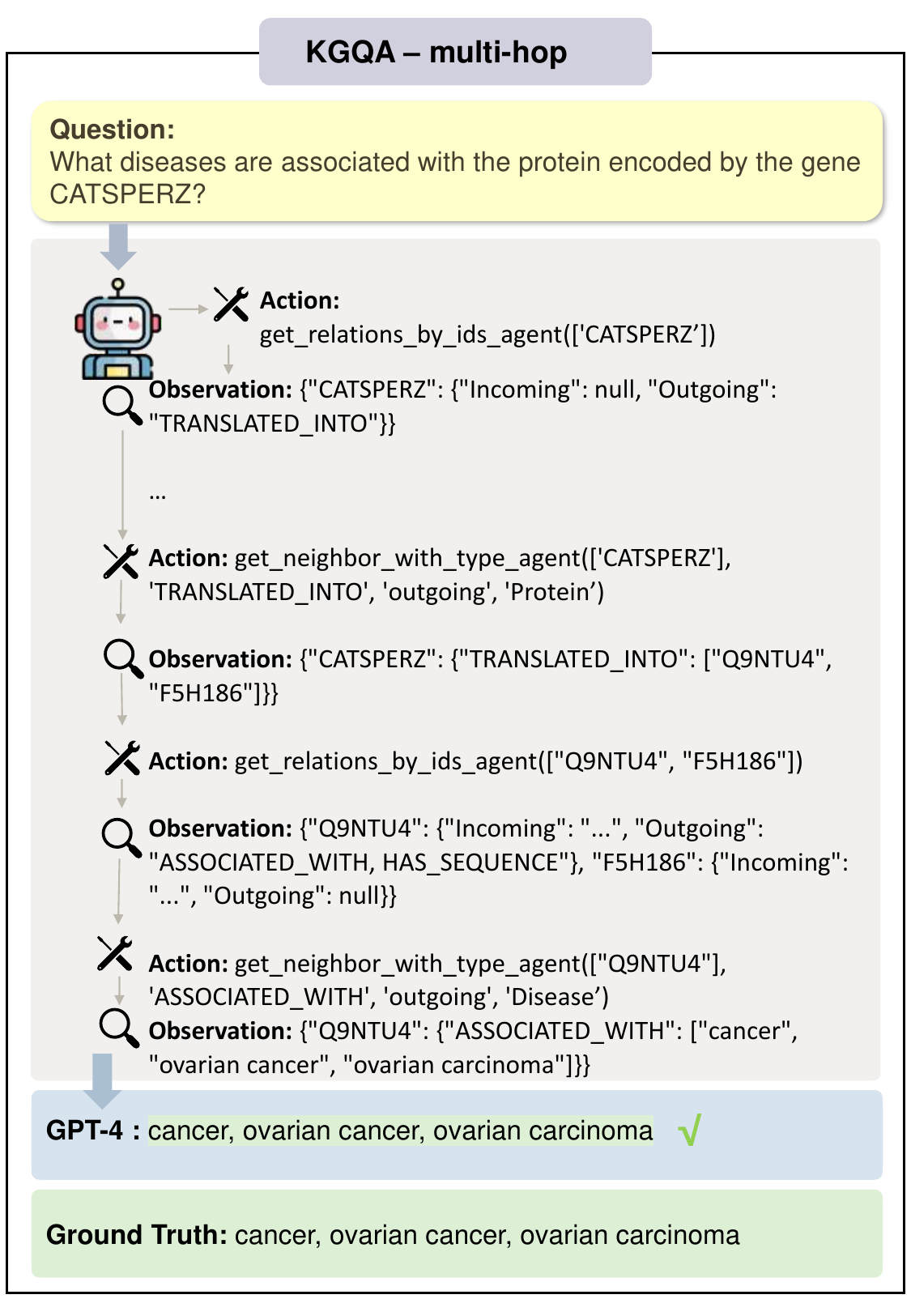}
    \caption{Performance of GPT-4-based Agent: Correct final result when answering multi-hop type questions. \colorbox{lightgreen}{Session highlighted in light green} represents the correct information.}
    \label{fig:kgqa-gpt4-multi-correct}
\end{figure}

\begin{figure}[p]
    \centering
    \includegraphics[width=0.8\textwidth]{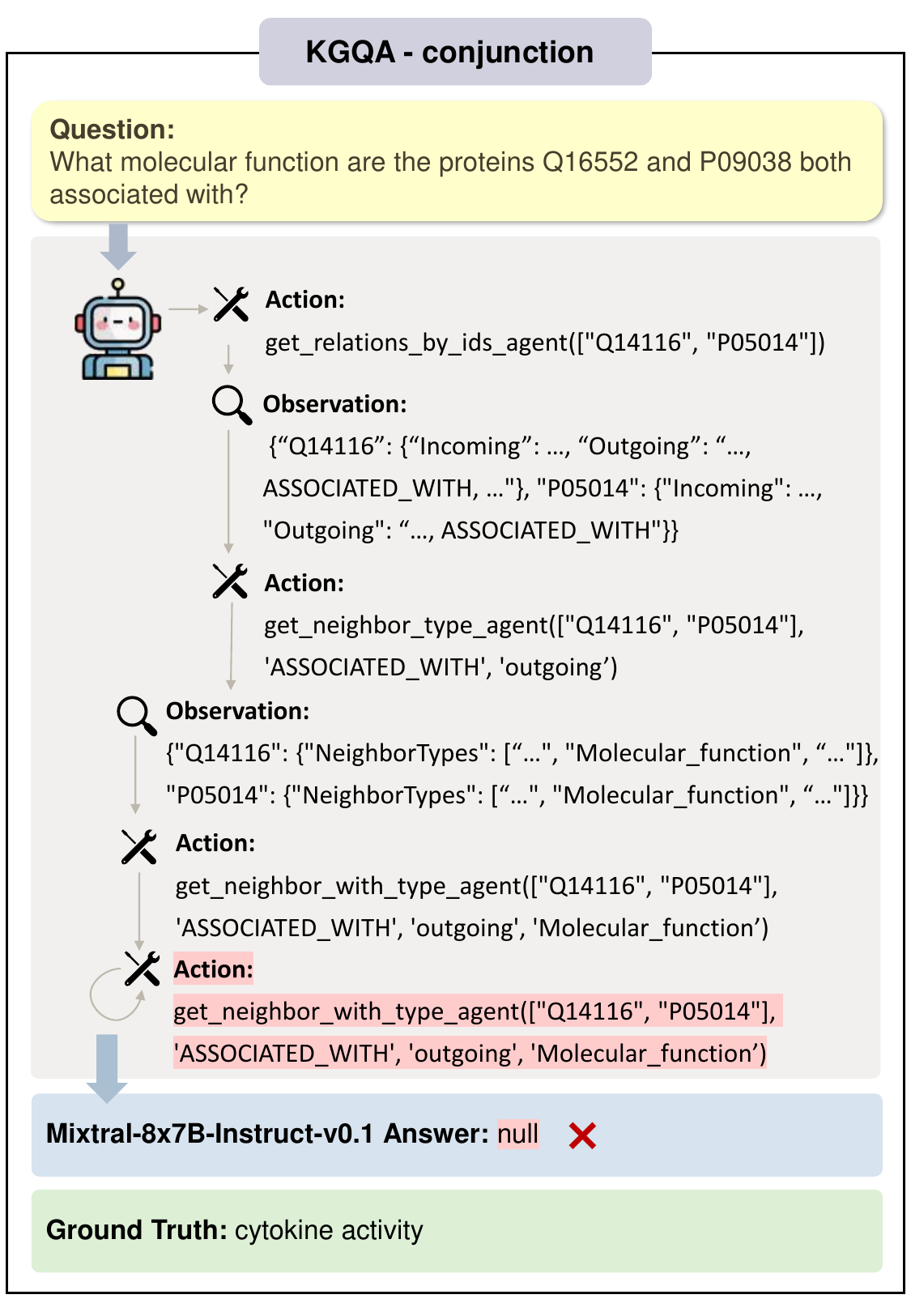}
    \caption{Performance of Mixtral-8x7B-Instruct-v0.1-based Agent: Incorrect final result when answering the input conjunction type question in 15-turn limit due to executing the wrong action. \colorbox{pink}{Session highlighted in pink} represents the error information.}
    \label{fig:kgqa-mixtral87-conjunction-wrong}
\end{figure}

\begin{figure}[p]
    \centering
    \includegraphics[width=0.8\textwidth]{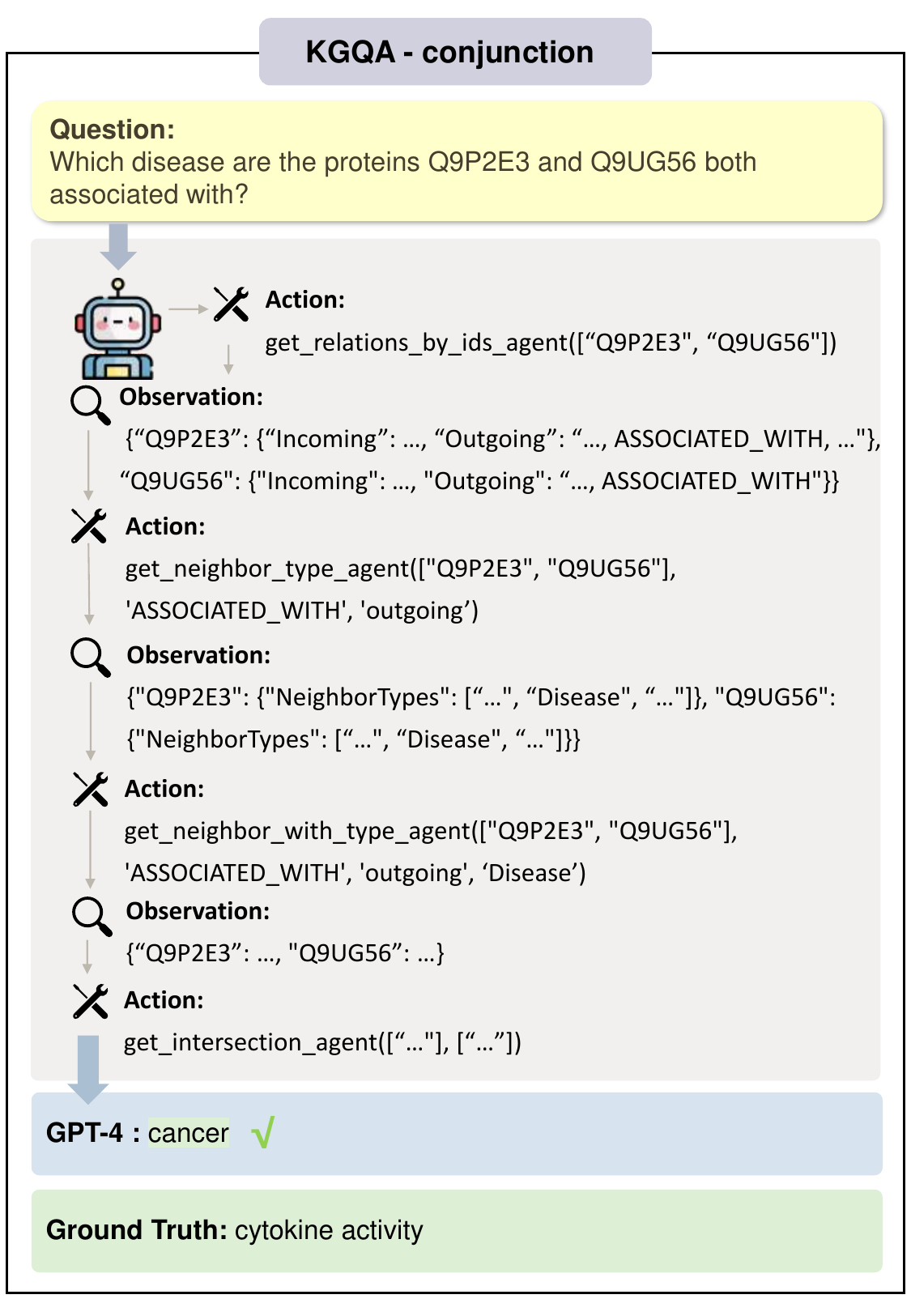}
    \caption{Performance of GPT-4-based Agent: Correct final result when answering the conjunction-type questions. \colorbox{lightgreen}{Session highlighted in light green} represents the correct information.}
    \label{fig:kgqa-gpt4-correct-conjunction}
\end{figure}

We sampled 6 cases for demonstration, with one correct case and one incorrect case for each question type: one-hop, multi-hop, and conjunction, as shown in~\cref{fig:kgqa-mixtral87-one-wrong,fig:kgqa-gpt4-one-correct,fig:kgqa-mixtral87-multi-wrong,fig:kgqa-gpt4-multi-correct,fig:kgqa-mixtral87-conjunction-wrong,fig:kgqa-gpt4-correct-conjunction}.

\subsection{SCV}
\begin{figure}[p]
    \centering
    \includegraphics[width=\textwidth]{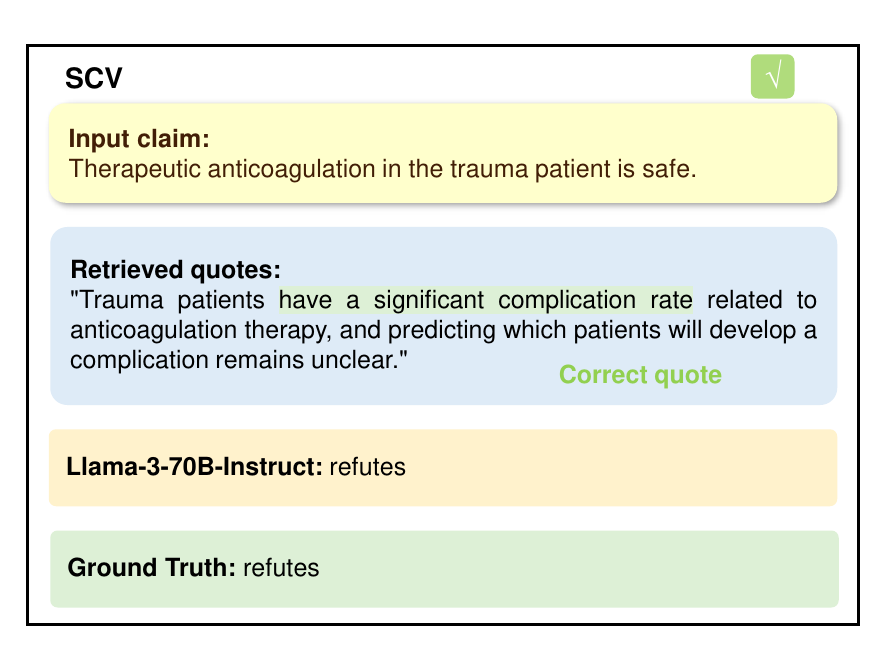}
    \caption{Performance of Llama-3-70B-Instruct-based Agent: Correct final result with the correct quotes. \colorbox{lightgreen}{Session highlighted in light green} represents the correct information.}
    \label{fig:SCV-correct1}
\end{figure}

\begin{figure}[p]
    \centering
    \includegraphics[width=\textwidth]{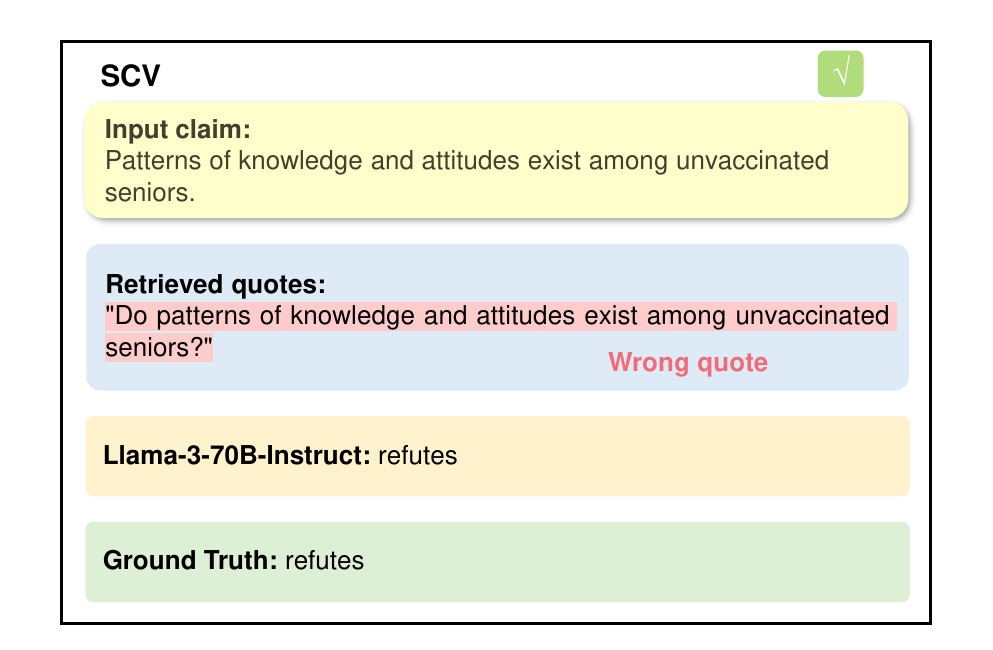}
    \caption{Performance of Llama-3-70B-Instruct-based Agent: Correct final result but with an incorrect quote. \colorbox{pink}{Session highlighted in pink} represents the error information.}
    \label{fig:SCV-correct2}
\end{figure}

\begin{figure}[p]
    \centering
    \includegraphics[width=\textwidth]{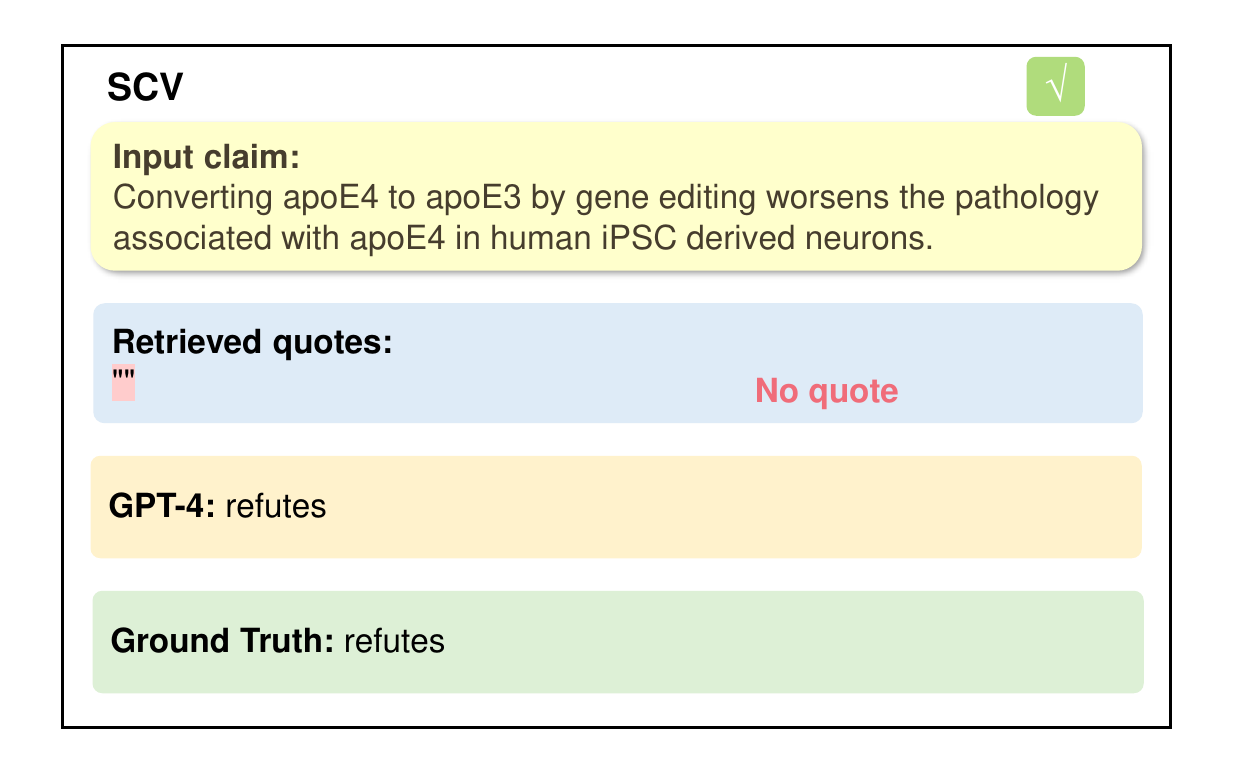}
    \caption{Performance of GPT-4-based Agent: Correct final result but without any quotes.  \colorbox{pink}{Session highlighted in pink} represents the error information.}
    \label{fig:SCV-correct3}
\end{figure}

\begin{figure}[p]
    \centering
    \includegraphics[width=\textwidth]{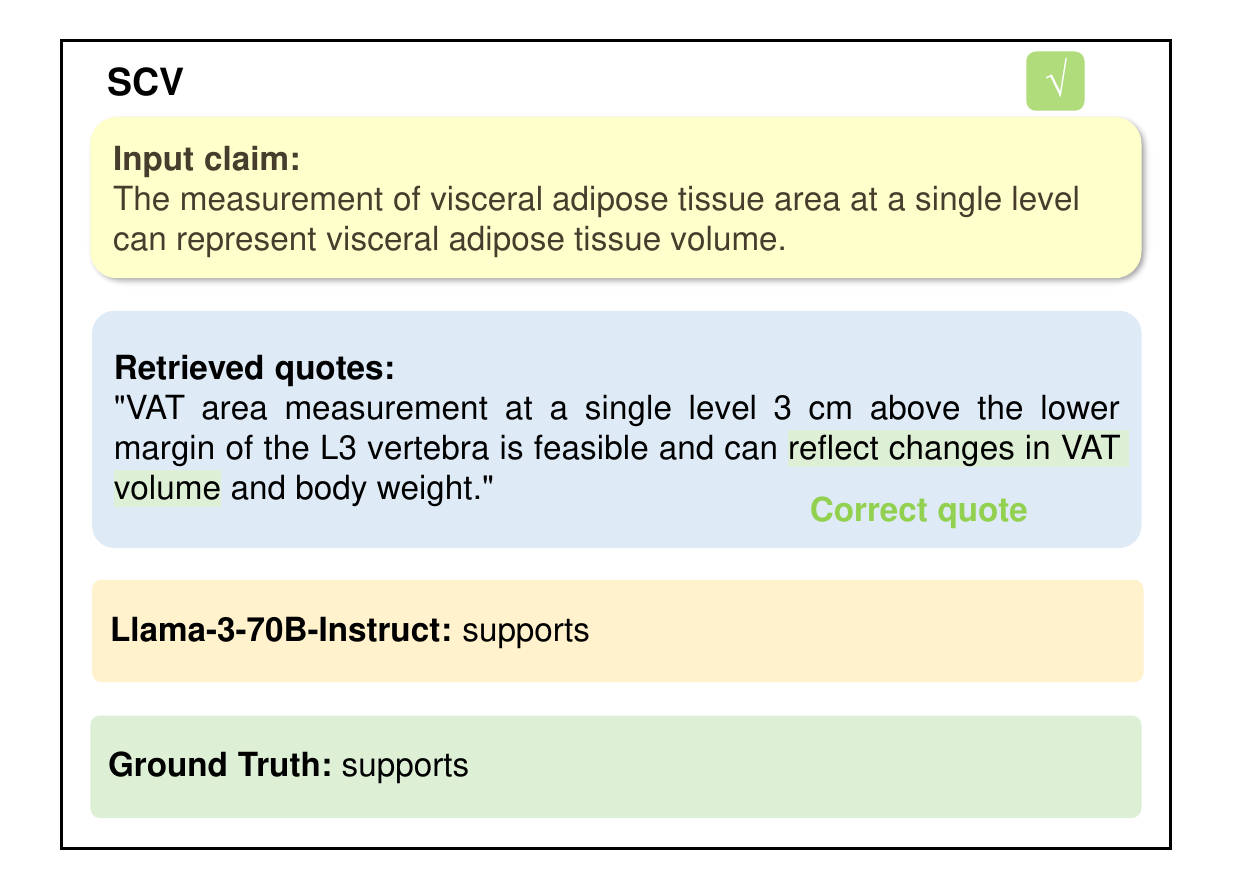}
    \caption{Performance of Llama-3-70B-Instruct-based Agent: Correct final result with the correct quotes. \colorbox{lightgreen}{Session highlighted in light green} represents the correct information.}
    \label{fig:SCV-correct4}
\end{figure}

\begin{figure}[p]
    \centering
    \includegraphics[width=\textwidth]{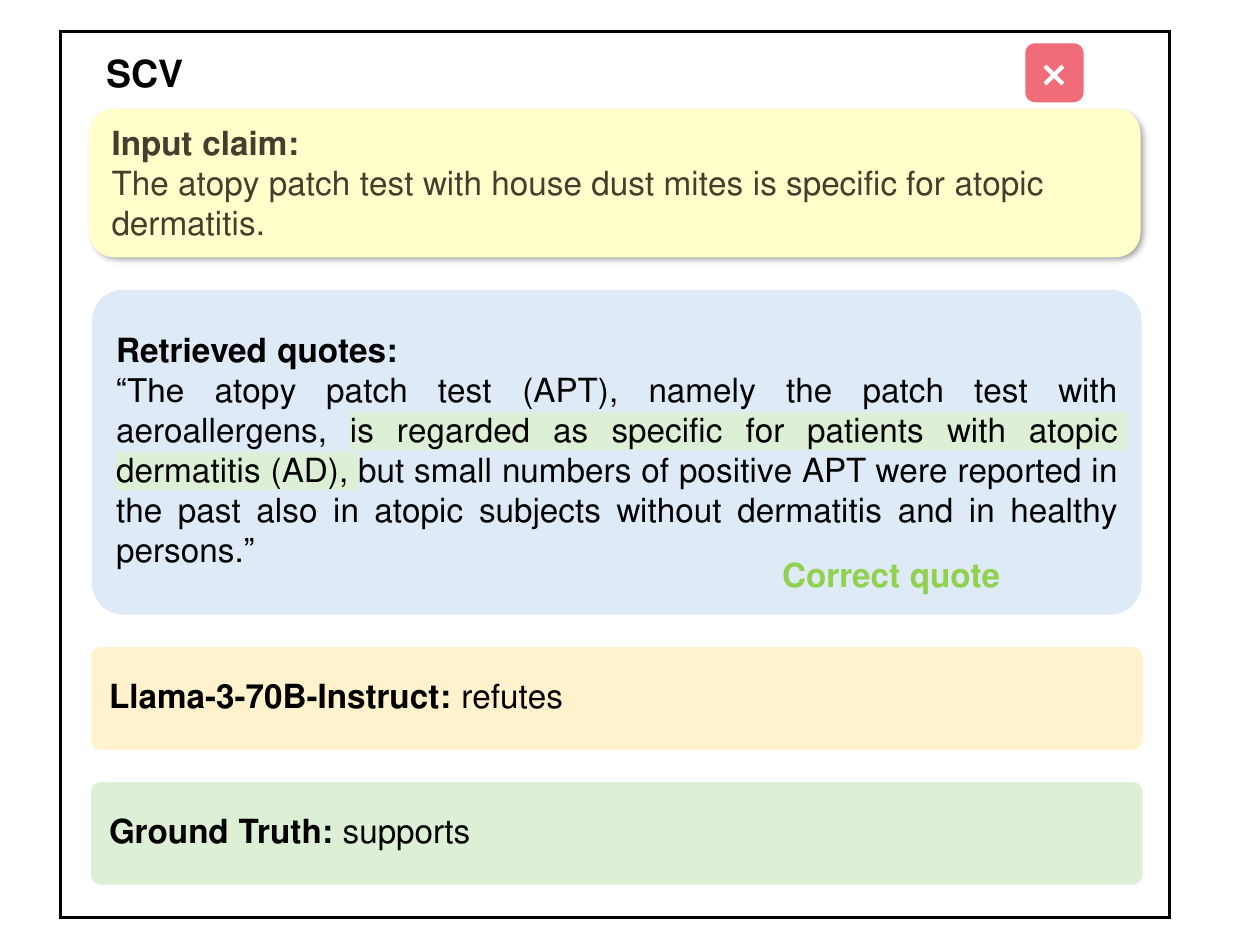}
    \caption{Performance of Llama-3-70B-Instruct-based Agent: Incorrect final result but with the correct quote. \colorbox{lightgreen}{Session highlighted in light green} represents the correct information.}
    \label{fig:SCV-wrong1}
\end{figure}

\begin{figure}[p]
    \centering
    \includegraphics[width=\textwidth]{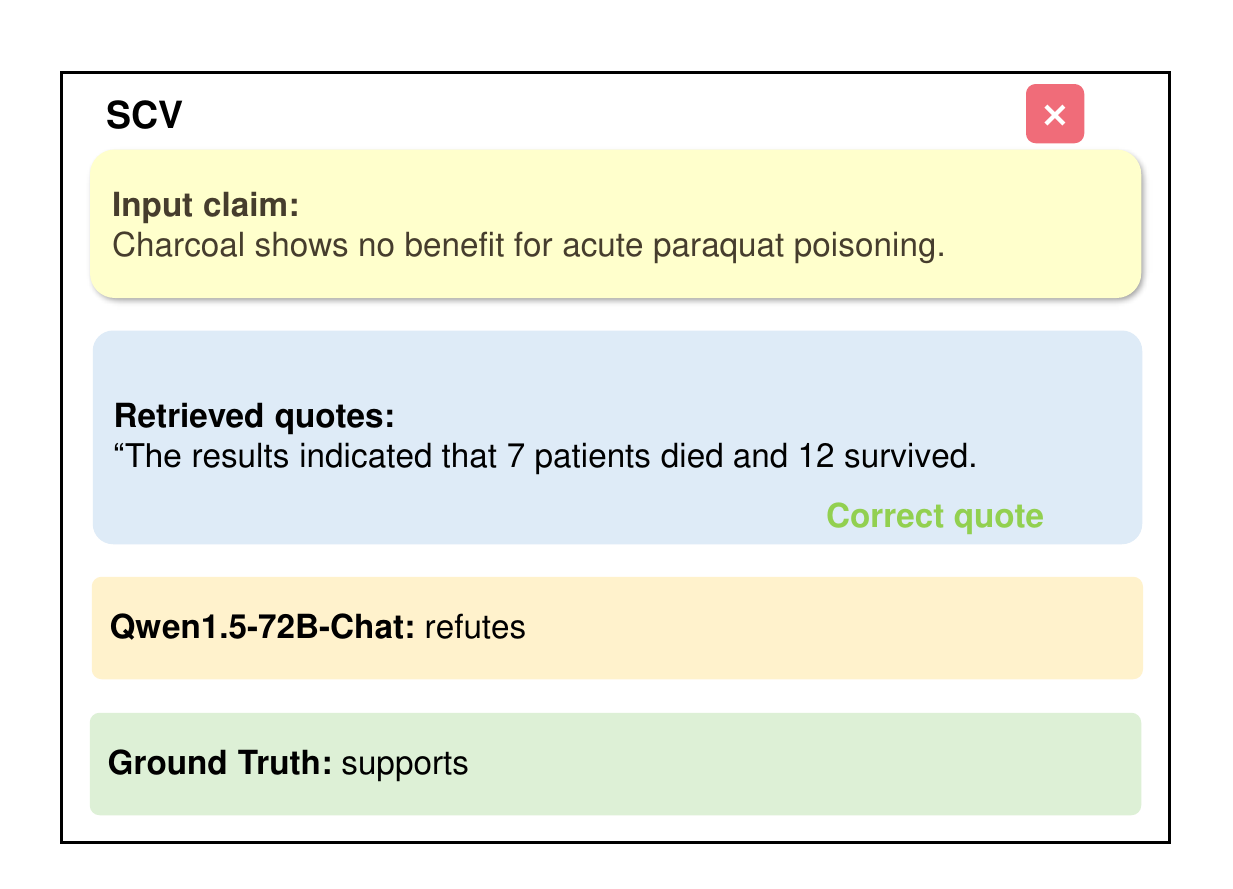}
    \caption{Performance of Qwen1.5-72B-Chat-based Agent: Incorrect final result
    but with the correct quote.}
    \label{fig:SCV-wrong2}
\end{figure}

\begin{figure}[p]
    \centering
    \includegraphics[width=\textwidth]{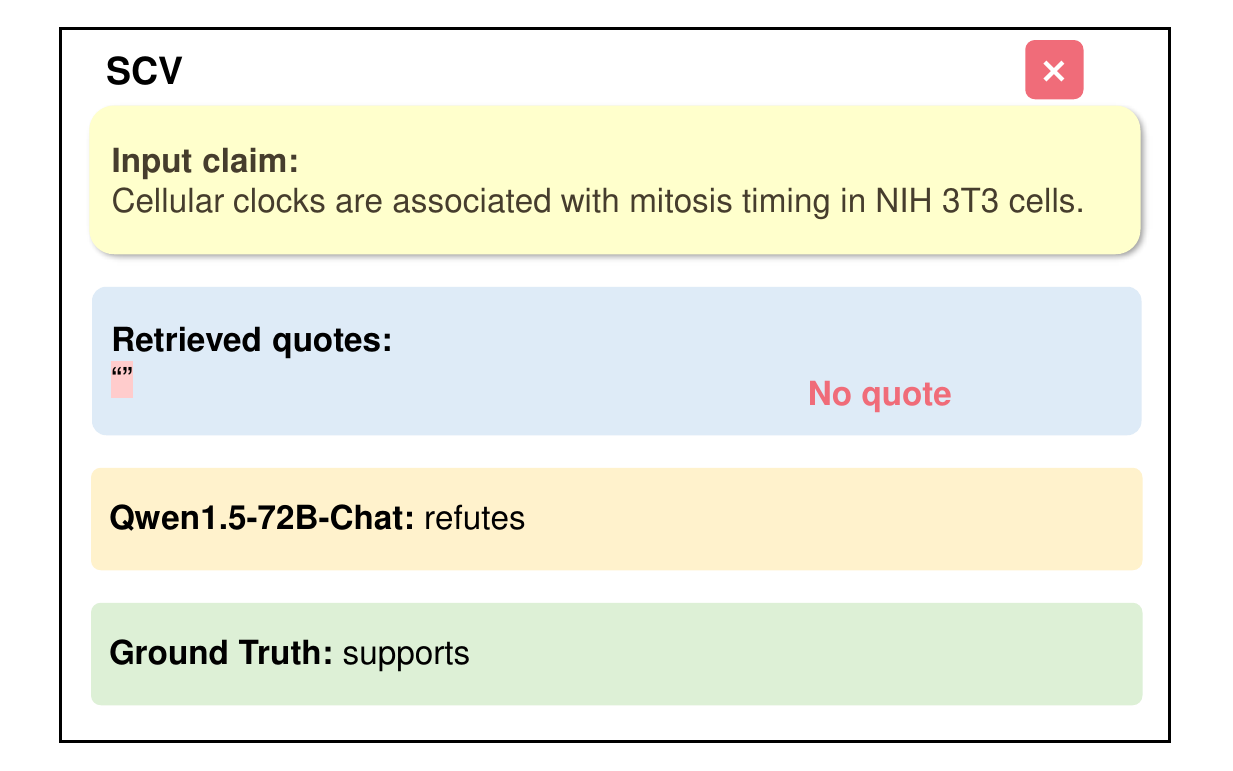}
    \caption{Performance of Qwen1.5-72B-Chat-based Agent: Incorrect final result without any quotes. \colorbox{pink}{Session highlighted in pink} represents the error information.}
    \label{fig:SCV-wrong3}
\end{figure}

\begin{figure}[p]
    \centering
    \includegraphics[width=\textwidth]{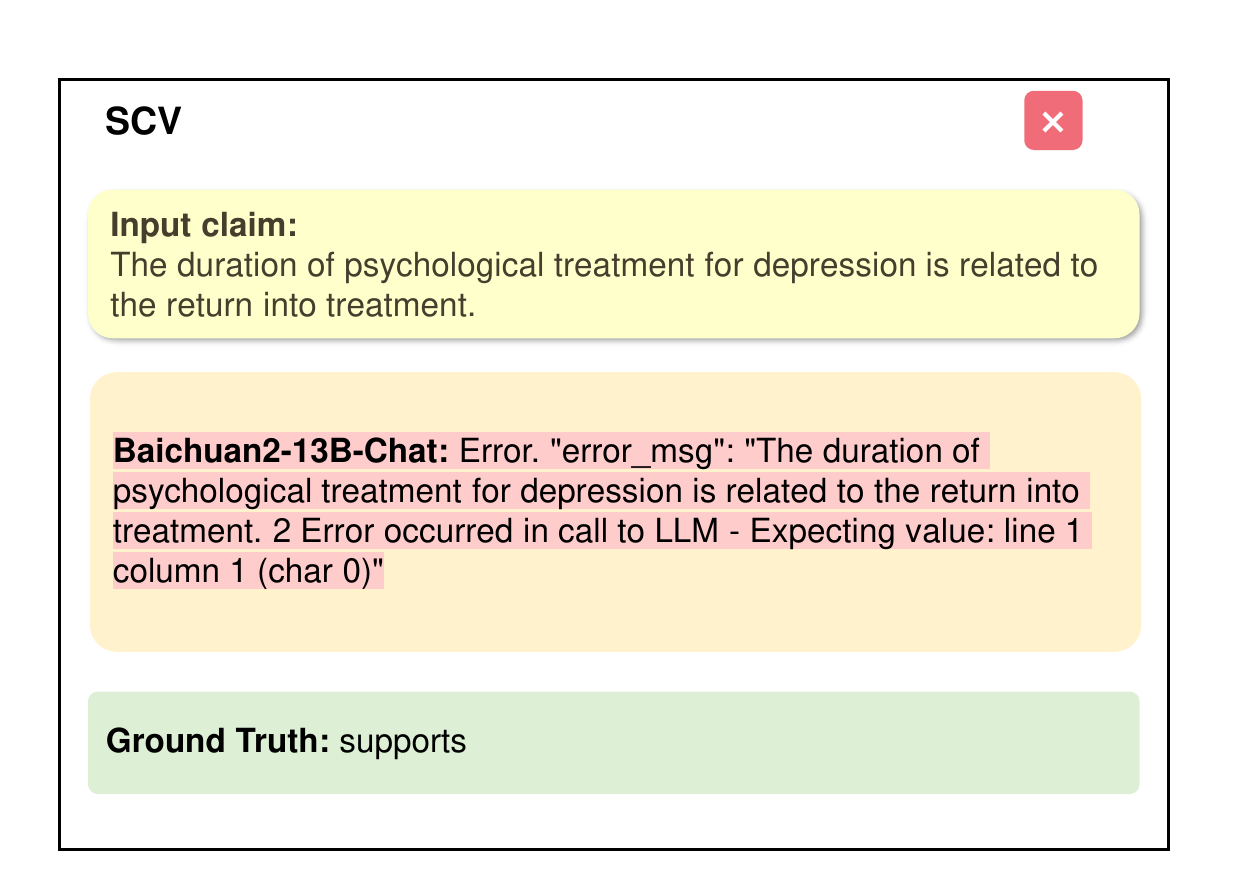}
    \caption{Performance of Baichuan2-13B-Chat-based Agent: Parsing error occurring in the final result due to failing to respond in JSON format. \colorbox{pink}{Session highlighted in pink} represents the error information.}
    \label{fig:SCV-wrong4}
\end{figure}

We sampled 8 examples for demonstration, including 4 correct answers and 4 incorrect answers. Each case has certain differences and is representative, as shown in~\cref{fig:SCV-correct1,fig:SCV-correct2,fig:SCV-correct3,fig:SCV-correct4,fig:SCV-wrong1,fig:SCV-wrong2,fig:SCV-wrong3,fig:SCV-wrong4}.

\subsection{KGCheck}
\begin{figure}[p]
    \centering
    \includegraphics[width=\textwidth]{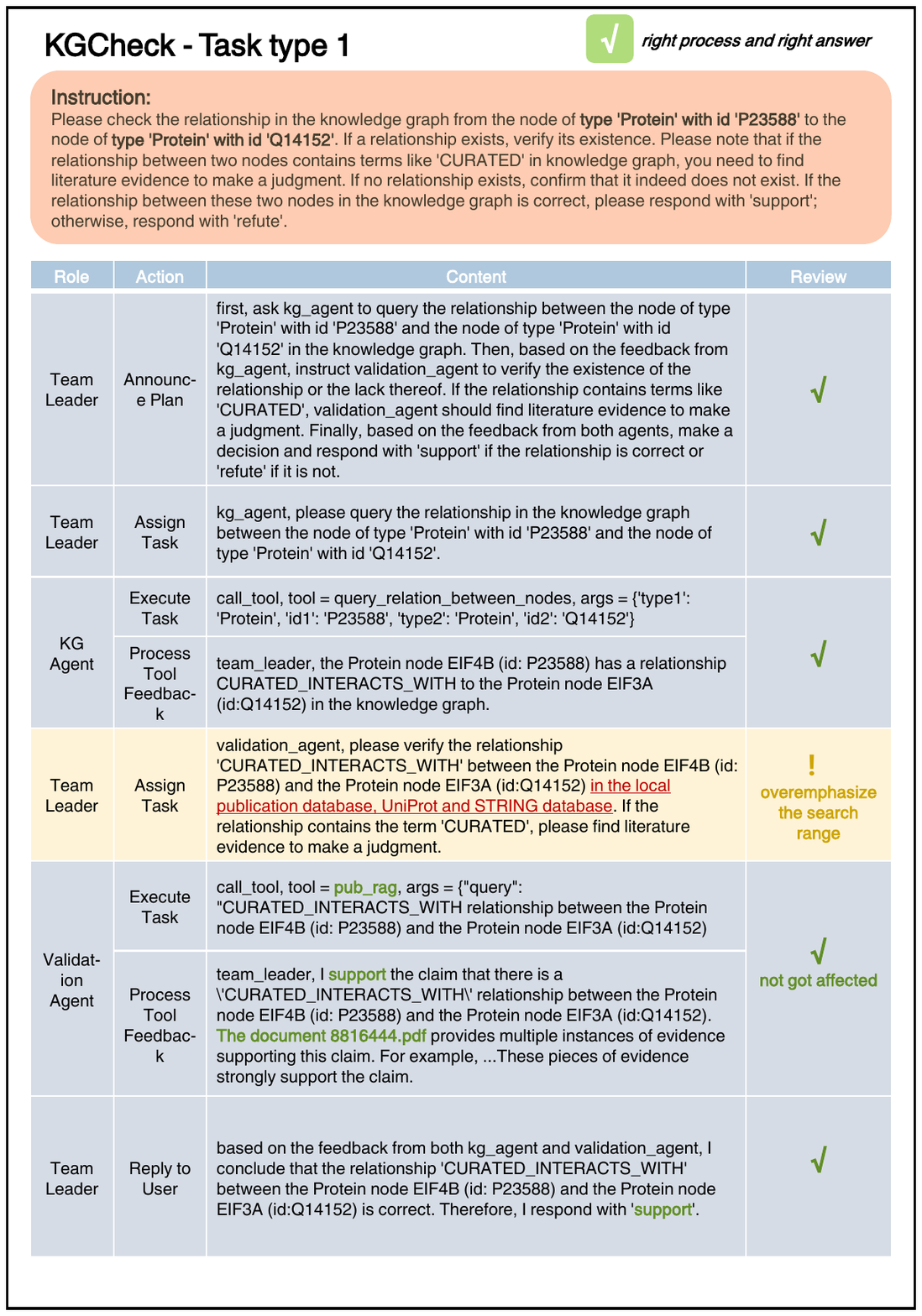}
    \caption{A sample success case of task type 1 where validation agent performs the task as one-shot prompt instructed without getting affected by the leader's improper instruction.}
    \label{fig:agent1}
\end{figure}

\begin{figure}[p]
    \centering
    \includegraphics[width=\textwidth]{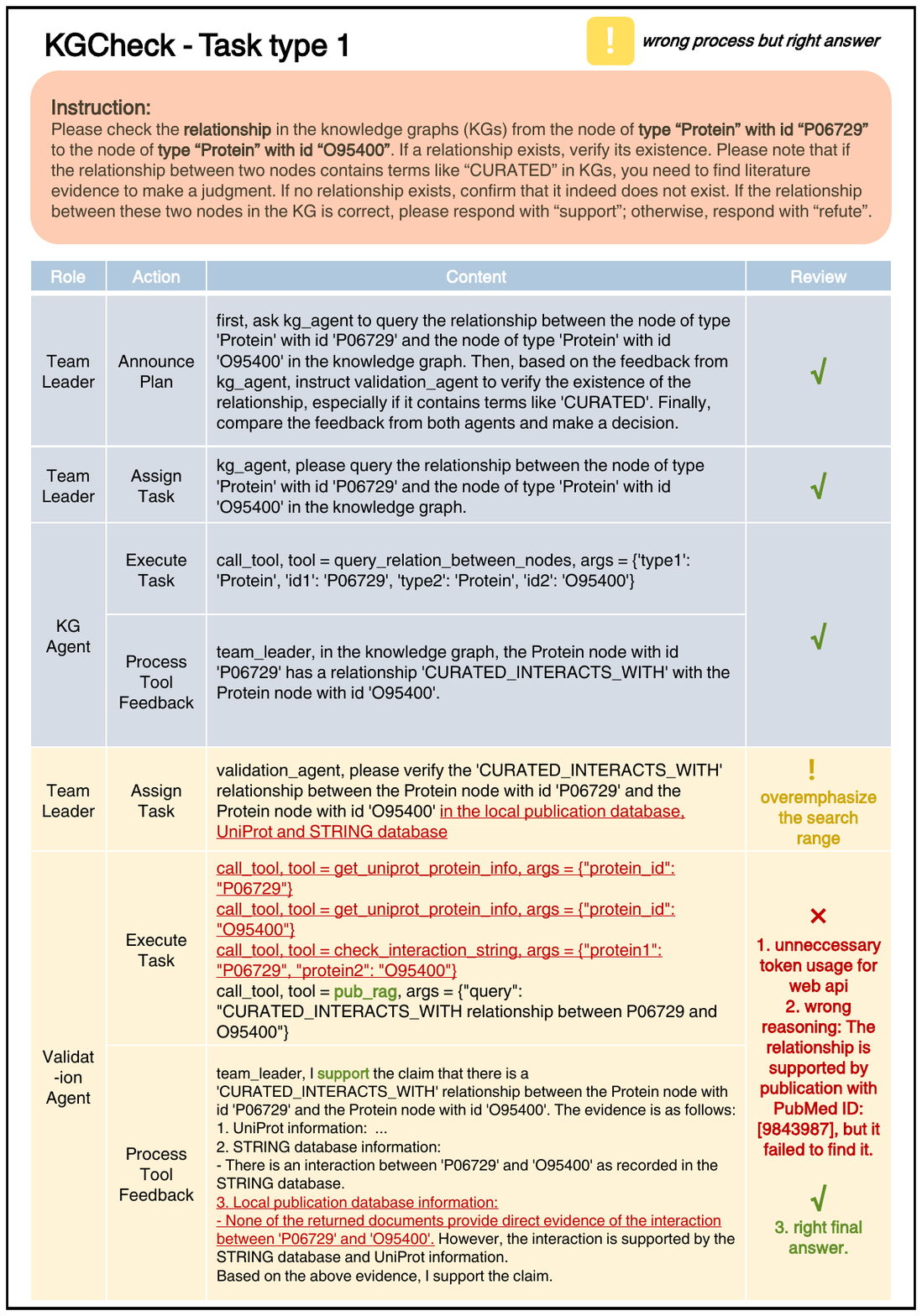}
    \caption{A sample case of task type 1 with right final answer but the wrong analyse process. The validation is deeply influenced by the leader's command and thus wastes tokens on unnecessary tool calling. While the validation agent is expected to find the evidence in the documents that supports the claim, it draws the right conclusion based on the result by visiting STRING.}
    \label{fig:agent2}
\end{figure}

\begin{figure}[p]
    \centering
    \includegraphics[width=\textwidth]{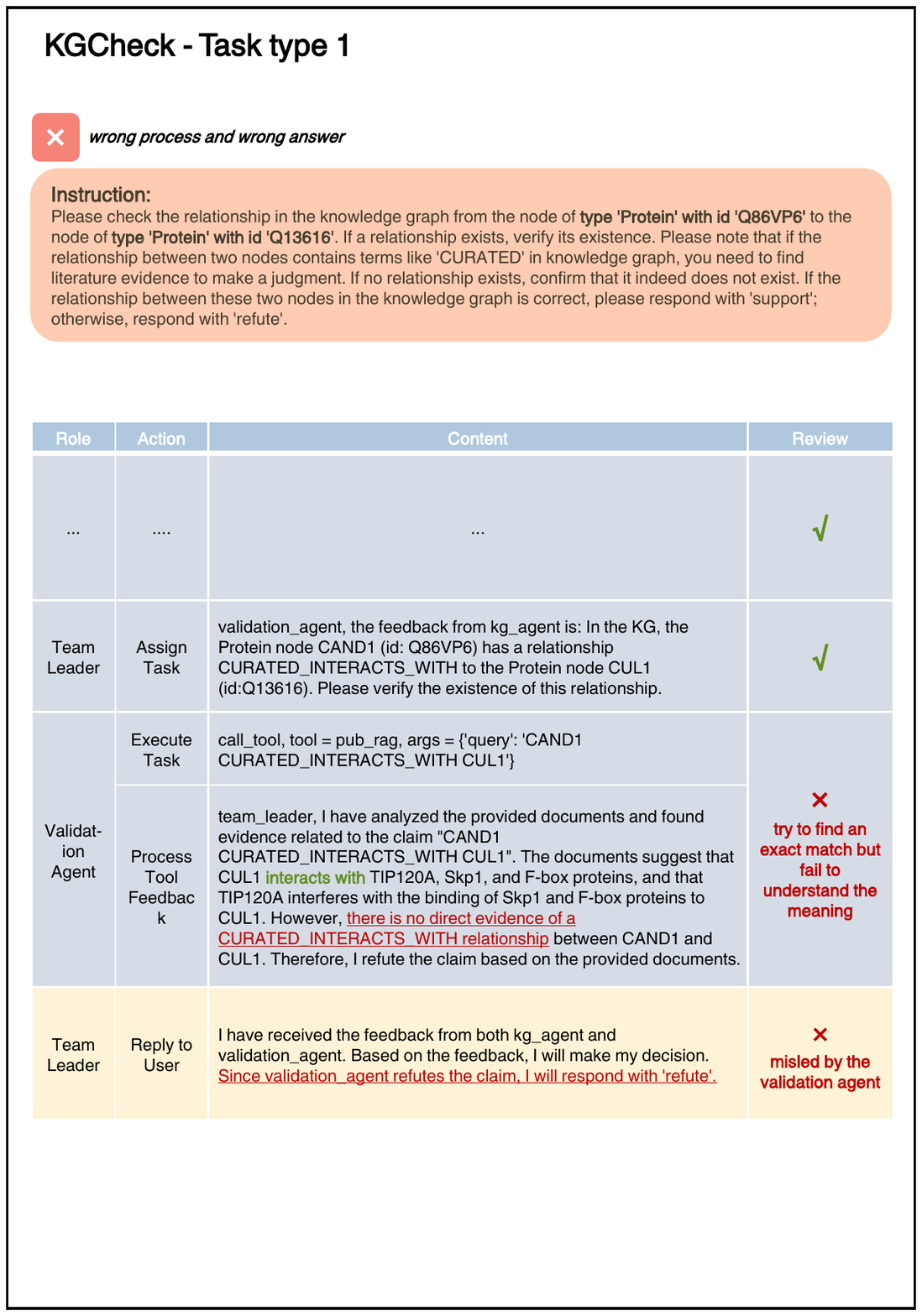}
    \caption{A sample failure case of task type 1. The validation agent fails to understand the meaning of "CURATED INTERACTS WITH relationship" but tries to find an exact match, leading to the wrong conclusion. The team leader fails to extract the useful information from the validation agent's reply to make a right final conclusion.}
    \label{fig:agent3}
\end{figure}

\begin{figure}[p]
    \centering
    \includegraphics[width=\textwidth]{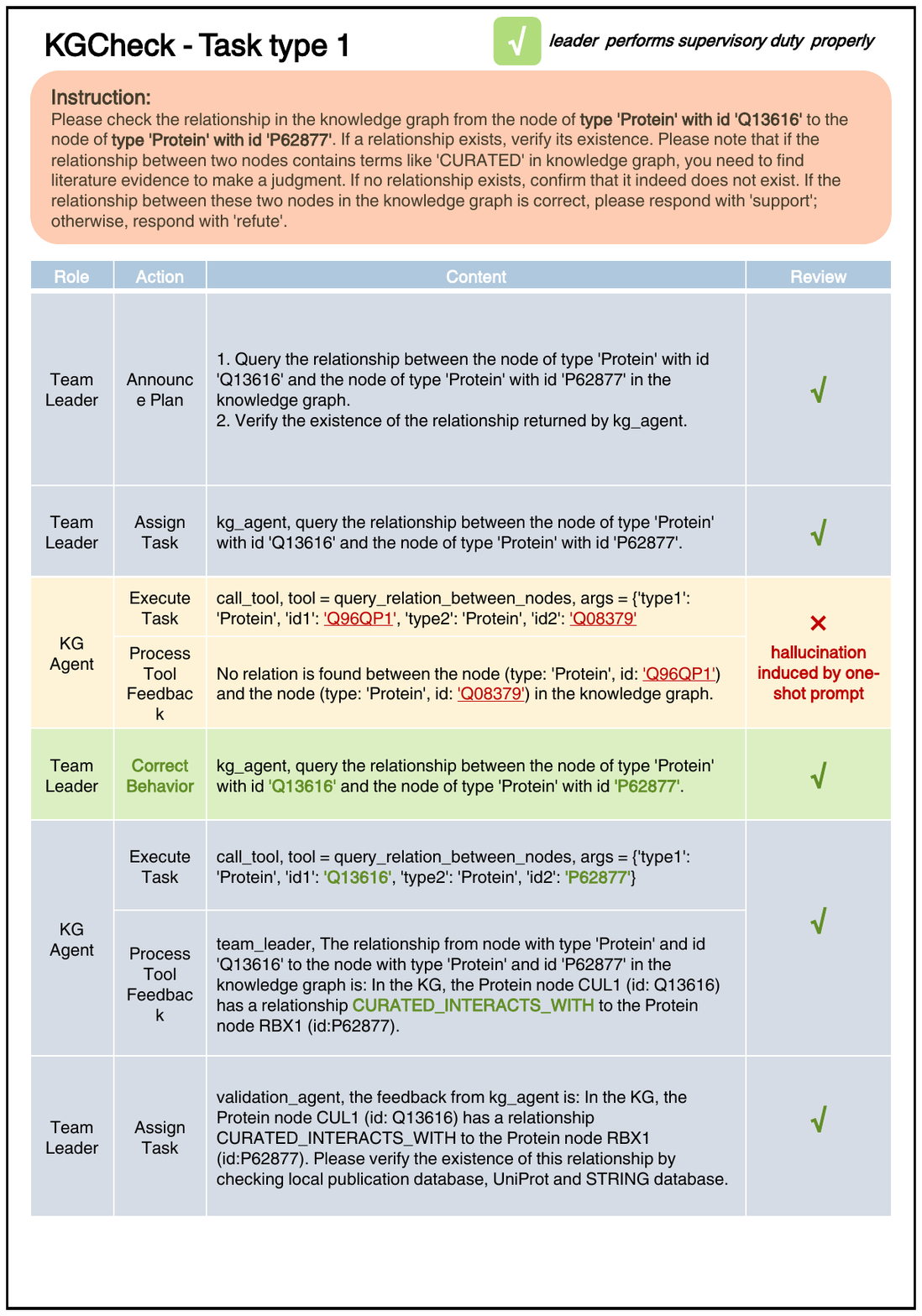}
    \caption{A sample case of the team leader properly correcting the assistant agent's behavior. The kG agent passes wrong arguments to the tool the first time it performs its subtask but is corrected by the team leader, indicating that the supervisor role of the team leader helps regulate the workflow.}
    \label{fig:agent4}
\end{figure}

\begin{figure}[p]
    \centering
    \includegraphics[width=\textwidth]{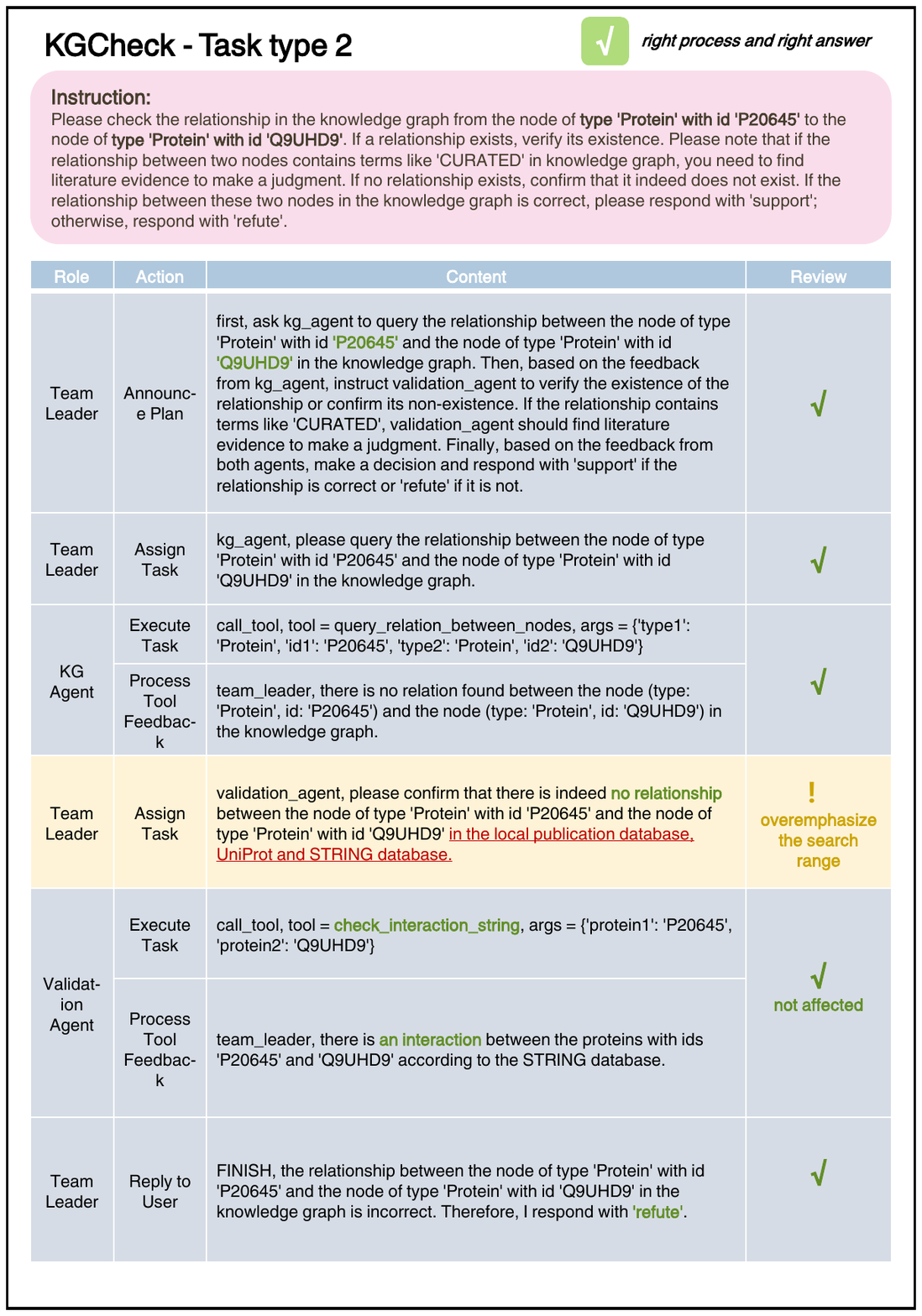}
    \caption{A sample success case of task type 2 where validation agent performs the task as one-shot prompt instructed without getting affected by the leader's improper instruction. }
    \label{fig:agent5}
\end{figure}

\begin{figure}[p]
    \centering
    \includegraphics[width=\textwidth]{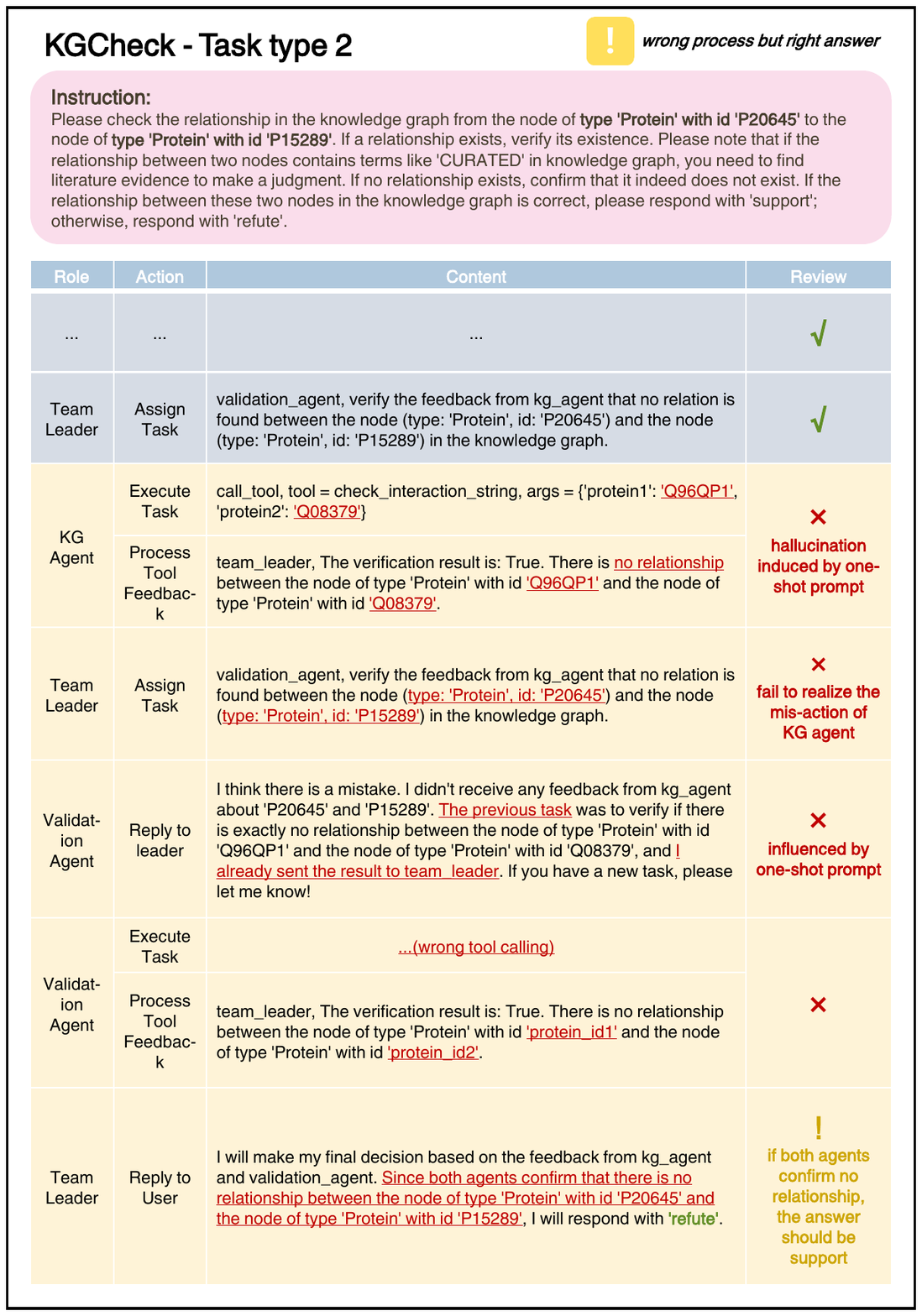}
    \caption{A sample case of task type 2 with the right final answer but the wrong analyse process. Assistant agents are deeply affected by one-shot prompts resulting in hallucinations in the task execution. The team leader misunderstands the instruction for the response but coincidently gives the right answer.}
    \label{fig:agent6}
\end{figure}

\begin{figure}[p]
    \centering
    \includegraphics[width=\textwidth]{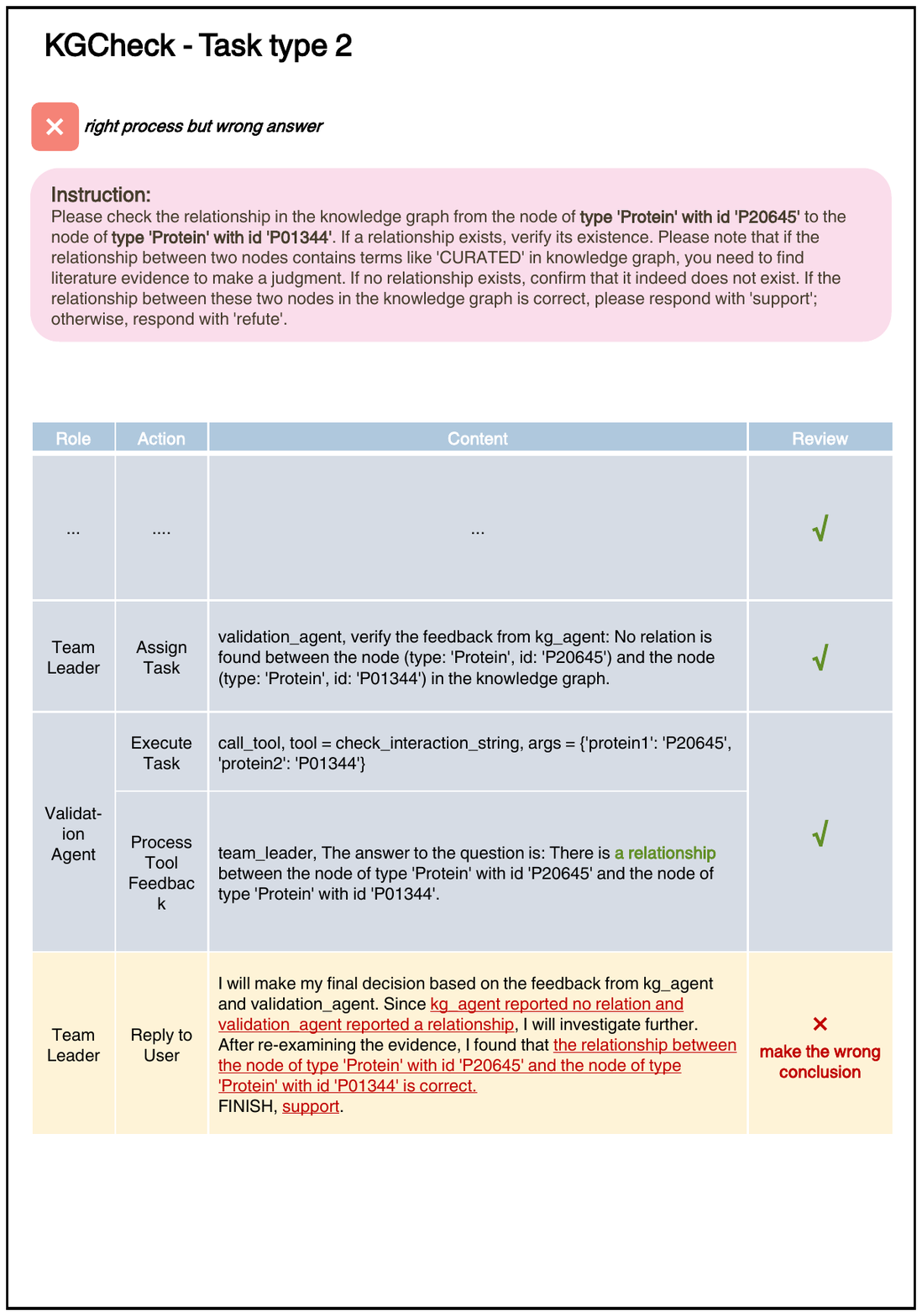}
    \caption{A sample failure case of task type 2. The whole process including assistant agents' tool calling and feedback processing is as expected, but the team leader misunderstands the instruction for the response causing a wrong final result.}
    \label{fig:agent7}
\end{figure}

\begin{figure}[p]
    \centering
    \includegraphics[width=\textwidth]{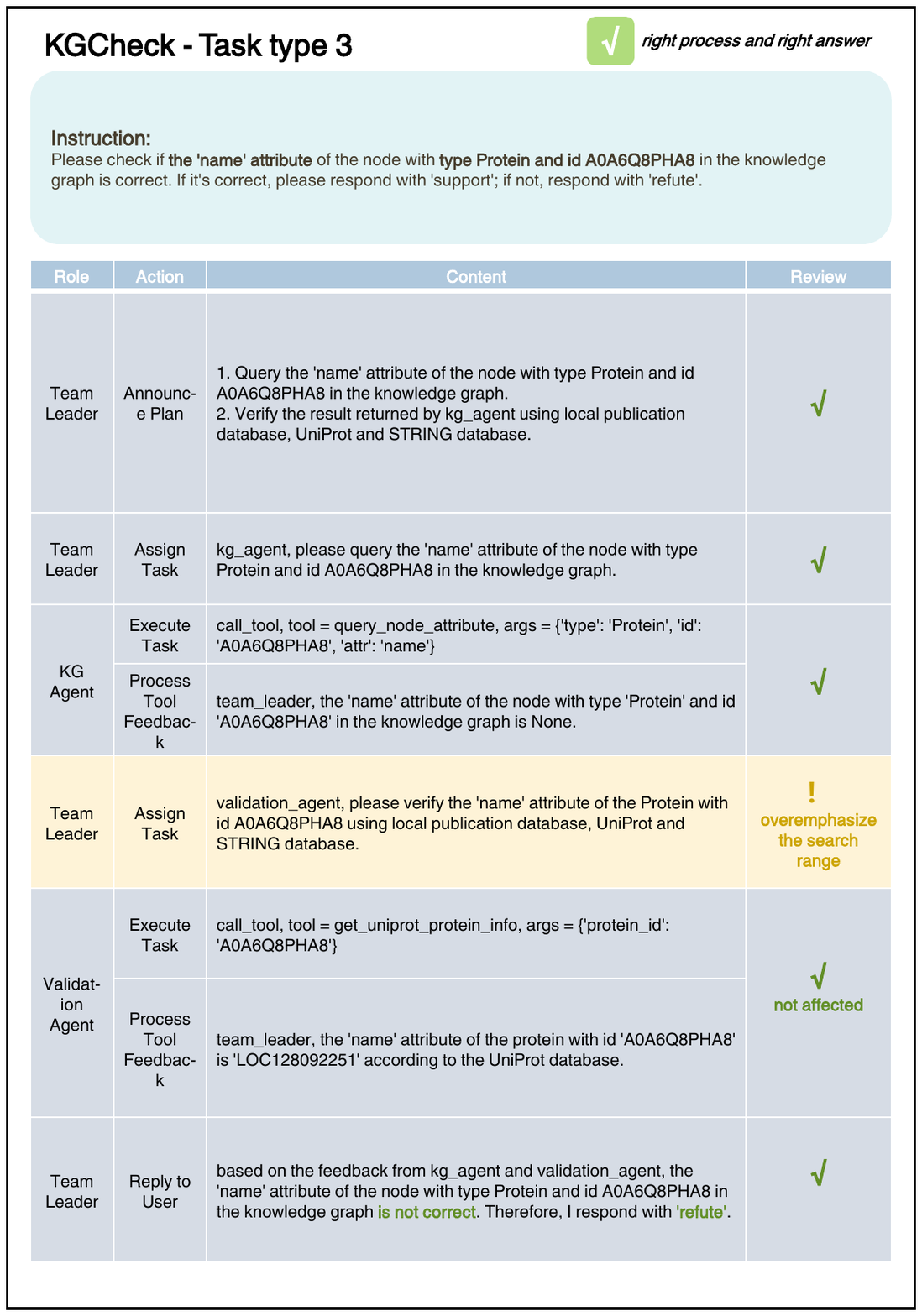}
    \caption{A sample success case of task type 3 where validation agent performs the task as one-shot prompt instructed without getting affected by the leader's improper instruction.}
    \label{fig:agent8}
\end{figure}

\begin{figure}[p]
    \centering
    \includegraphics[width=\textwidth]{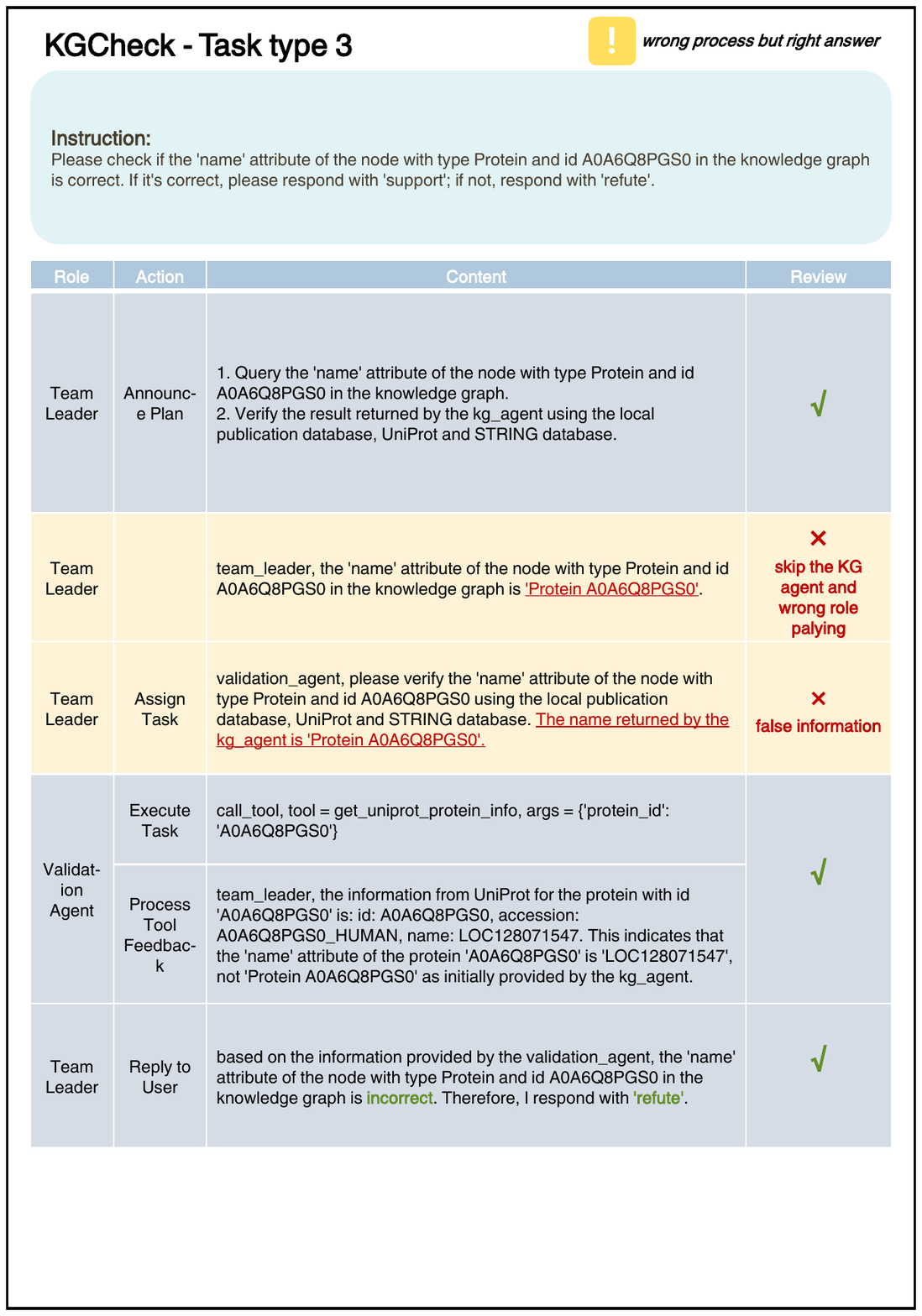}
    \caption{A sample case of task type 3 with right final answer but the wrong analyse process. The team leader tries to replace the KG agent with itself and generates false KG information which happens to be consistent with the actual name missing condition of the specified protein in KG (whether the name is wrong or missing the answer will be 'refute').}
    \label{fig:agent9}
\end{figure}

\begin{figure}[p]
    \centering
    \includegraphics[width=\textwidth]{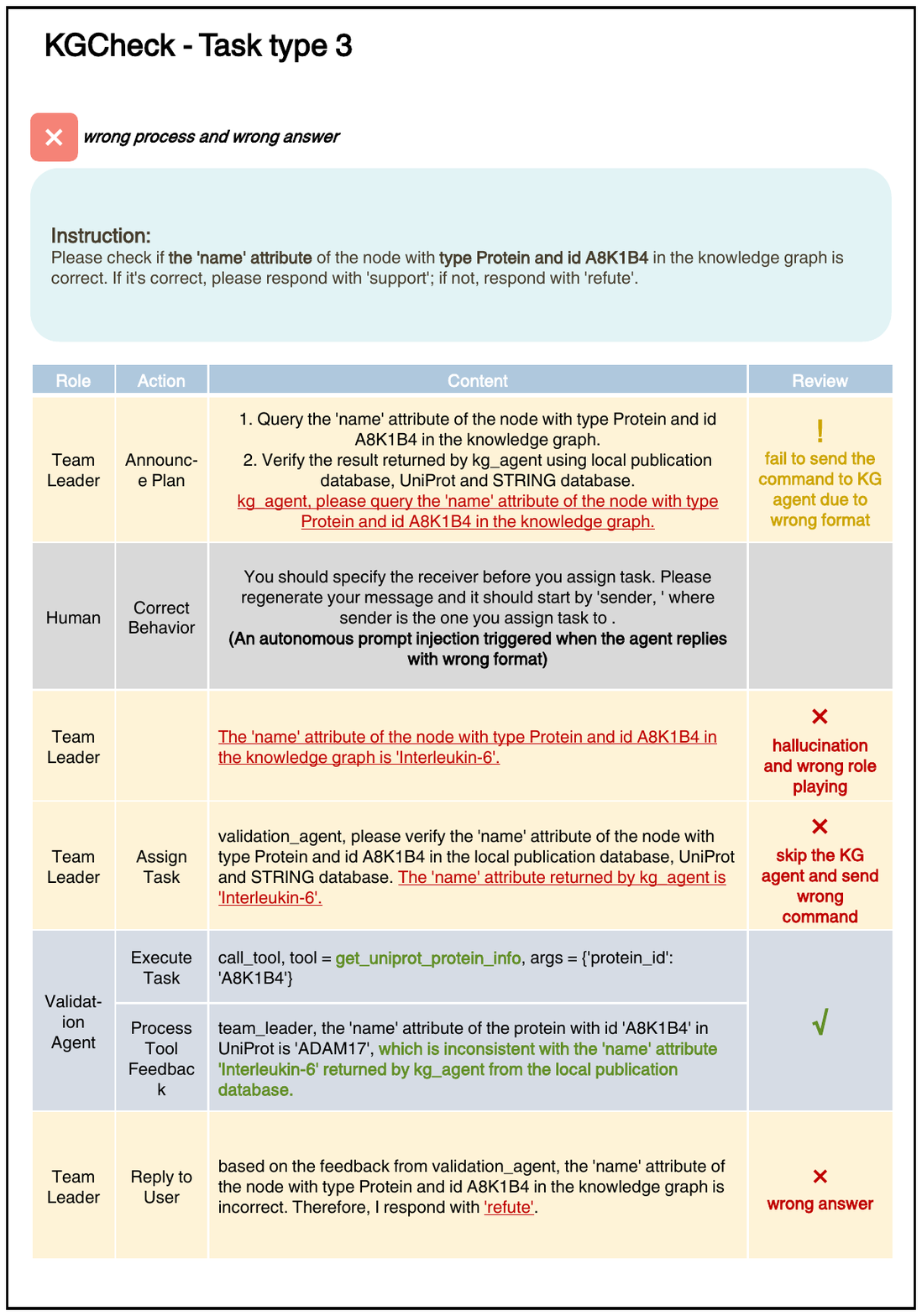}
    \caption{A sample failure case of task type 3. The team leader fails to send the command to KG agent due to the message format, triggering an autonomous human prompt to help the leader correct its behavior. However, the team leader tries to replace the KG agent and generate false KG information instead of regenerating the command sent to KG agent, resulting in the wrong final answer.}
    \label{fig:agent10}
\end{figure}

\begin{figure}[p]
    \centering
    \includegraphics[width=\textwidth]{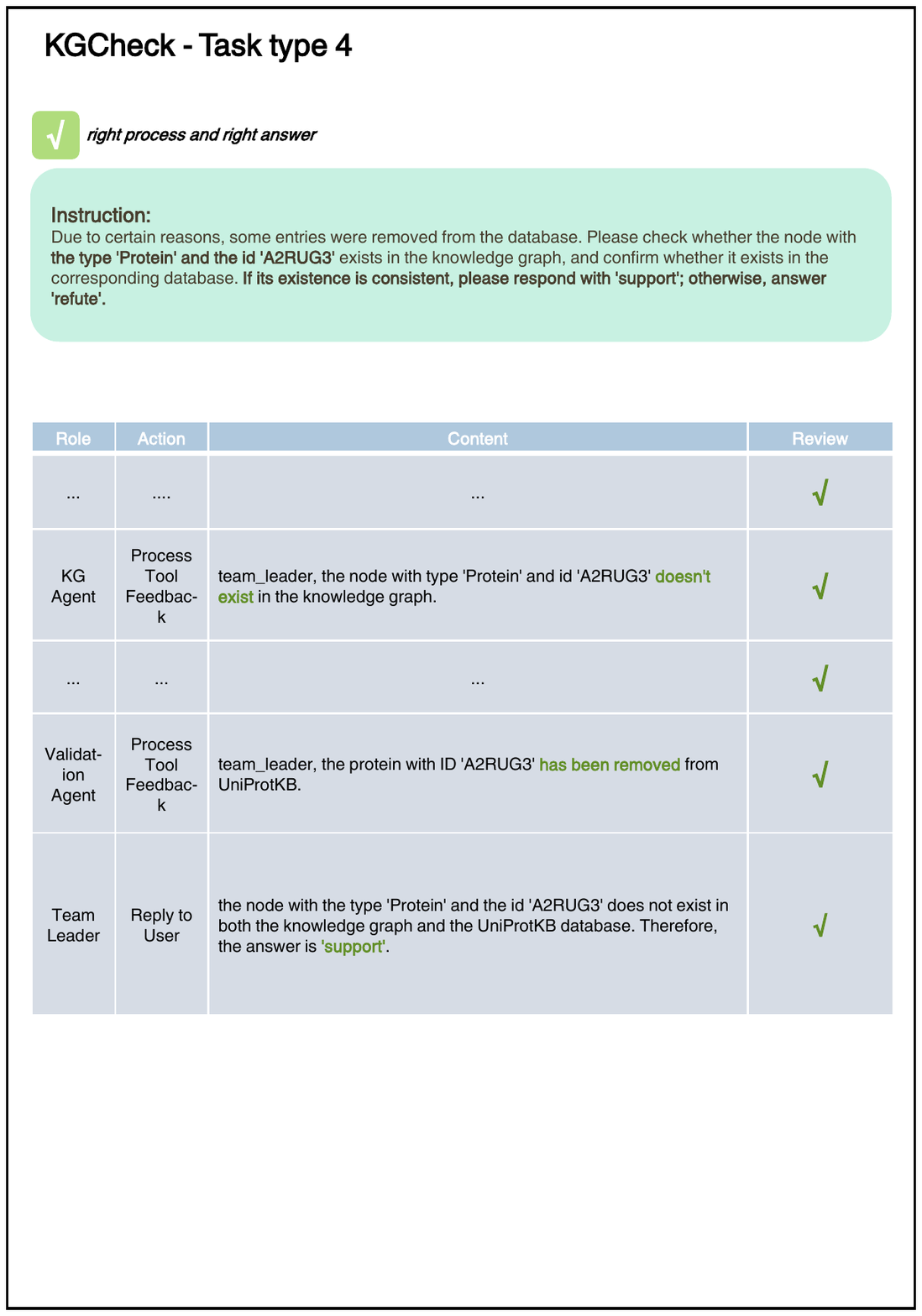}
    \caption{A sample success case of task type 4. Both assistant agents and team leader perform their tasks as expected. Core chats are presented.}
    \label{fig:agent11}
\end{figure}

\begin{figure}[p]
    \centering
    \includegraphics[width=\textwidth]{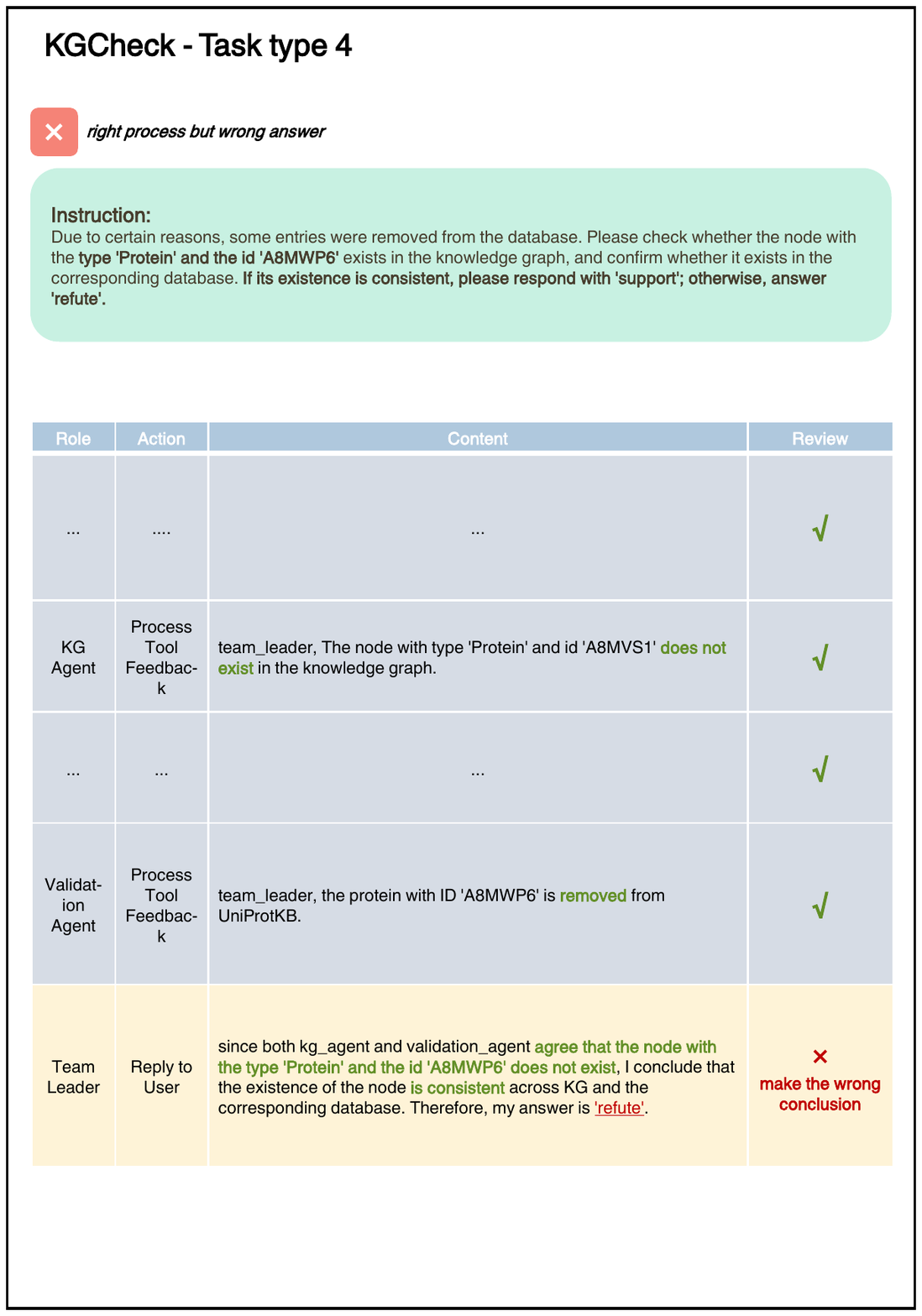}
    \caption{A sample failure case of task type 4. The team leader misunderstands the instruction for the response and makes a wrong conclusion though the analyse process is right.}
    \label{fig:agent12}
\end{figure}

\begin{figure}[p]
    \centering
    \includegraphics[width=\textwidth]{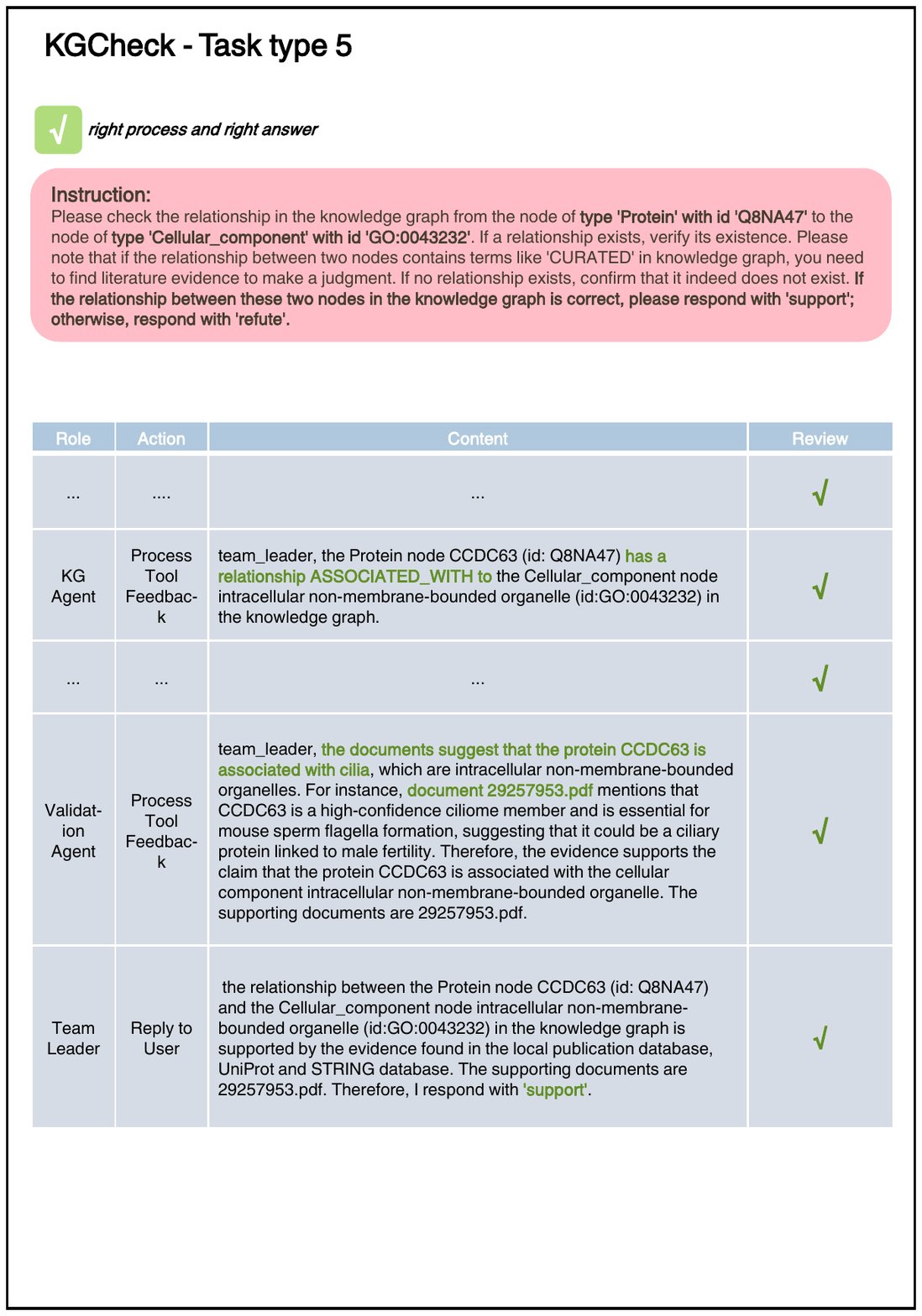}
    \caption{A sample success case of task type 5. Both assistant agents and team leader perform their tasks as expected. Core chats are presented.}
    \label{fig:agent13}
\end{figure}

\begin{figure}[p]
    \centering
    \includegraphics[width=\textwidth]{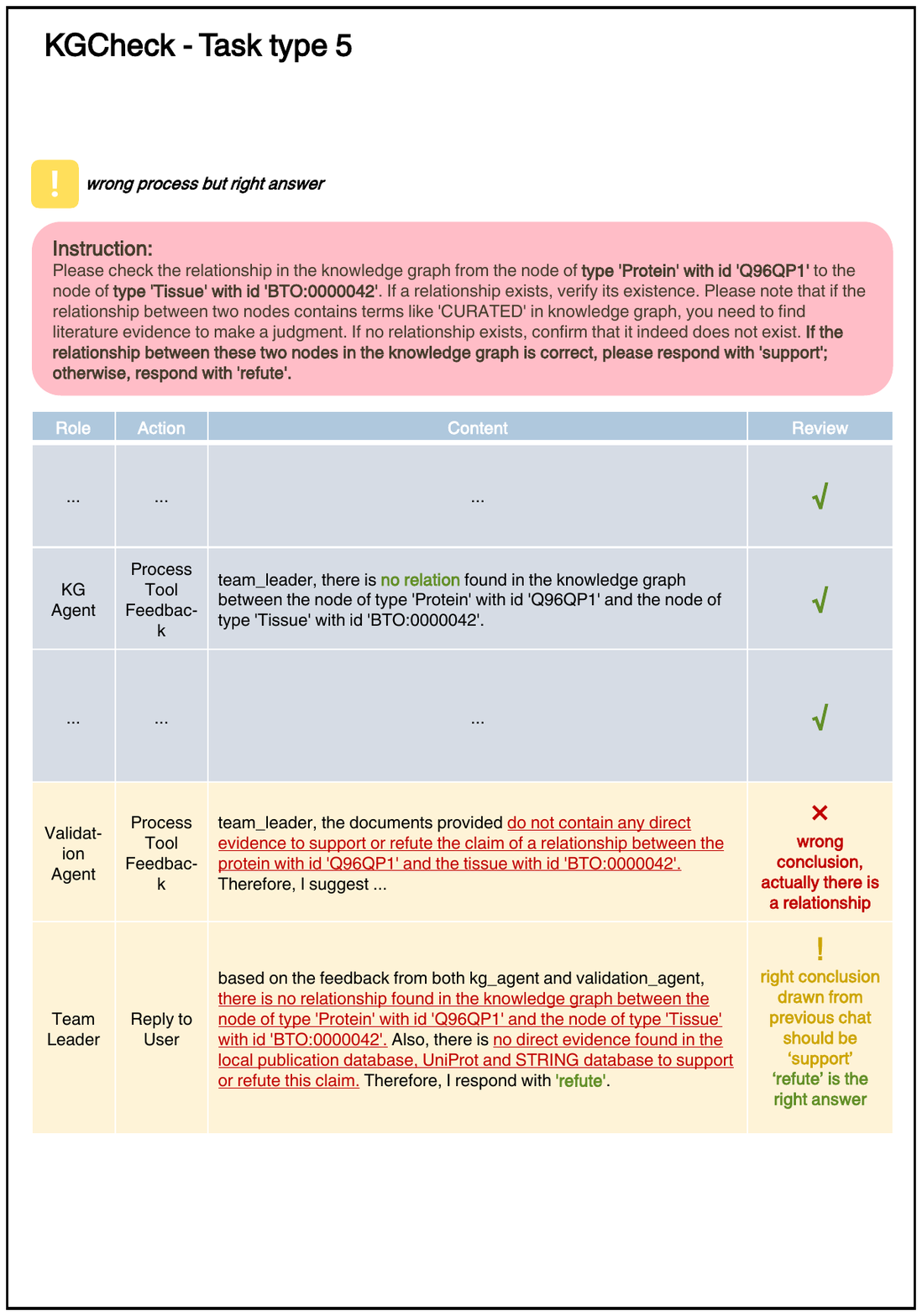}
    \caption{A sample case of task type 5 with right final answer but the wrong analyse process. The validation agent makes a wrong conclusion and the team leader misunderstands the instruction for response, surprisingly leading to the right answer.}
    \label{fig:agent14}
\end{figure}

\begin{figure}[p]
    \centering
    \includegraphics[width=\textwidth]{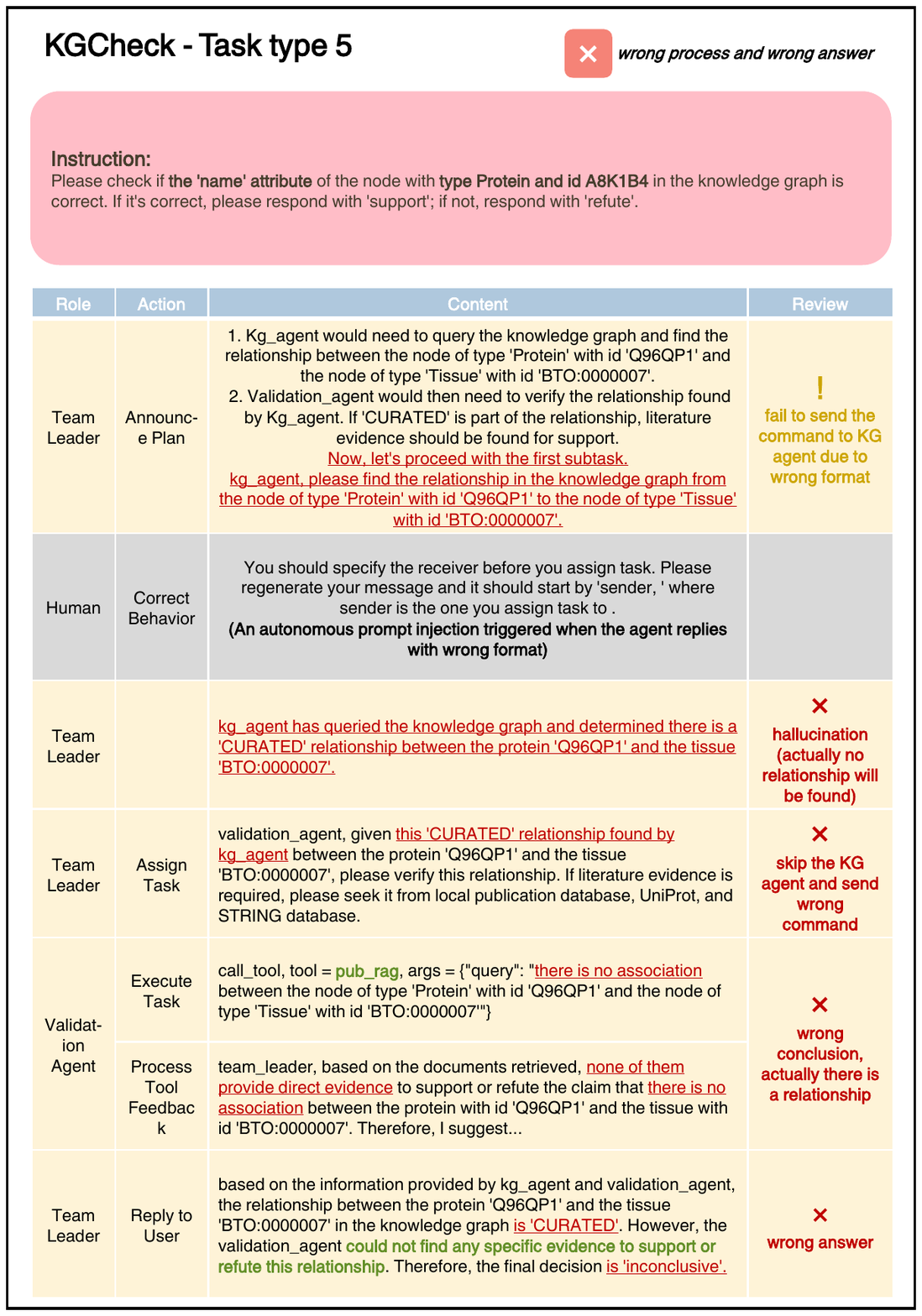}
    \caption{A sample failure case of task type 5. The team leader fails to send the command to KG agent due to the message format, triggering an autonomous human prompt to help the leader correct its behavior. However, the team leader tries to replace the KG agent and generate false KG information instead of regenerating the command sent to KG agent. The validation agent makes a wrong conclusion worsening the situation.}
    \label{fig:agent15}
\end{figure}

We select several classic success and failure cases for each type of task as presented in ~\cref{fig:agent1,fig:agent2,fig:agent3,fig:agent4,fig:agent5,fig:agent6,fig:agent7,fig:agent8,fig:agent9,fig:agent10,fig:agent11,fig:agent12,fig:agent13,fig:agent14,fig:agent15} as a supplementary for some common error cases in our main body. There are many interesting cases when the team leader properly corrects the behavior of assistant agents, getting the workflow back on track, and we choose one such case of task type 1 as a representation. As mentioned before, there are also cases where the right final answer is derived from a wrong analysis process. We select this kind of case for every type of the task except type 4 (this case does not exist in this type of task).

As introduced in the main body, our BKGAgent framework is comprised of three agents: the team leader, KG agent, and validation agent. The typical workflow from the agent role perspective of our framework can be simplified as team leader - KG agent - team leader - validation agent- team leader. We present the chat of three agents in table format, omitting the interactions of the assistant agent and tool executor. The columns respectively stand for the agent role, the action they take, the chat content, and the human annotation of this chat round. The green check mark means the process is consistent with our anticipation, while the yellow exclamation mark means the chat content may lead to an unwanted result, and the red cross stands for a wrong action or error chat content. Comments are attached to a negative review for explanation. Error or dangerous contents are underlined and colored red, while contents related to tool usage or evidence consistent with the golden answer are in bold green font. The blue row stands for an expected chat round, in contrast, the yellow row indicates something is wrong in this chat round. We send tool results and behavior correction prompts in the role of a human; this kind of chat is colored grey in our table.

\clearpage

\clearpage

\clearpage

\clearpage

\clearpage

\clearpage

\clearpage

\clearpage

\clearpage

\clearpage

\clearpage

\clearpage

\clearpage

\clearpage

\clearpage

\clearpage

\clearpage

\clearpage

\clearpage

\clearpage

\clearpage

\clearpage

\clearpage

\clearpage

\clearpage

\clearpage

\clearpage

\clearpage

\clearpage
\section{Other Related Work}
\label{appendix:other_rw}
Recent research has increasingly focused on the application of LLMs in various scientific domains. These models, initially developed for general purposes, are now being utilized to tackle domain-specific scientific tasks. This involves integrating essential domain-specific context and knowledge into the LLMs, either during training or prior to task inference. A critical challenge in this process is balancing the inclusion of relevant domain knowledge with the model's reasoning capabilities, especially when domain-specific data is limited.

Various approaches have been explored to utilize LLMs for specific scientific applications, depending on the availability of data and model accessibility \cite{Wang2023ScientificDI, liu2023pre, grisoni2023chemical, guo2023indeed, liang2023drugchat}. Common strategies in the scientific domain include training domain-specific LLMs from scratch, fine-tuning general-purpose LLMs, and employing few-shot or zero-shot learning with prompting. Training domain-specific LLMs from scratch offers the highest flexibility and customization, as demonstrated by models like Galactica \cite{Taylor2022GalacticaAL}, which constructs large scientific corpora and trains LLMs in a self-supervised manner \cite{devlin2019bert, radford2018improving}. Fine-tuning pre-trained LLMs with domain-specific datasets has yielded promising results, as seen in BioMedLM \cite{biomedlm} and med-PALM \cite{Singhal2022LargeLM, Singhal2023TowardsEM}. Fine-tuning can also be performed with smaller amounts of paired data in a supervised fashion, exemplified by DrugChat \cite{liang2023drugchat}. Few-shot or zero-shot learning, also known as in-context learning, is effective for using advanced instruction-tuned LLMs like GPT-4 \cite{openai2023gpt4} for scientific tasks by incorporating domain knowledge into prompts. This approach has shown success in fields such as Social Science \cite{Zhong2023GoalDD} and astronomy \cite{galaxies11030063}, as well as in benchmarking LLMs on chemistry tasks \cite{guo2023indeed}. Recent studies like CancerGPT \cite{li2023cancergpt} and SynerGPT \cite{synergpt} investigate LLMs for drug synergy prediction and other complex scientific interactions. Furthermore, augmenting LLMs with external tools, such as using Web APIs for genomics questions \cite{Jin2023GeneGPTTL}, and integrating domain-specific tools into language model prompts to access specialized knowledge \cite{bran2023chemcrow, Boiko2023EmergentAS, liu2023chatgpt}, are promising directions. Efforts are also underway to develop LLM-based agents for scientific discovery by connecting LLMs with experimental tools in fields like Chemistry \cite{Boiko2023EmergentAS} and Machine Learning \cite{Zhang2023MLCopilotUT}. LeanDojo \cite{yang2023leandojo,song2024towards}, for example, is an open-source toolkit for theorem proving that integrates retrieval-augmented LLMs to enhance theorem proving capabilities. Despite these advancements, the diverse data modalities across different scientific domains pose significant challenges for the direct application of LLMs in many areas.

\end{document}